\documentclass[journal]{IEEEtran}
\usepackage{tipa}
\usepackage{bbm}
\usepackage{mathrsfs}
\usepackage{cite,url,subfigure,epsfig,graphicx}
\usepackage{amssymb,amsmath}
\usepackage{indentfirst}
\usepackage{algorithmic}
\usepackage{algorithm}
\usepackage{pdfpages}
\usepackage{times,verbatim,amsfonts,amsmath,color}
\usepackage{pifont}
\usepackage{multirow}
\usepackage{booktabs}
\usepackage{adjustbox}
\usepackage{makecell}
\usepackage{tabularx}
\usepackage{caption}
\usepackage{tikz}
\usepackage[edges]{forest}
\usetikzlibrary{arrows.meta, shapes.geometric, calc, shadows}
\definecolor{hidden-draw}{RGB}{20,68,106}
\definecolor{hidden-pink}{RGB}{255,245,247}

\IEEEoverridecommandlockouts
\makeatletter
\def\ps@headings{\def\@oddhead{\mbox{}\scriptsize\rightmark \hfil \thepage}\def\@evenhead{\scriptsize\thepage \hfil \leftmark\mbox{}}\def\@oddfoot{}\def\@evenfoot{}}
\makeatother 

\newcommand{\tabincell}[2]{\begin{tabular}{@{}#1@{}}#2\end{tabular}}
\usepackage{geometry}
\begin{document}

\title{Large Model Based Agents: State-of-the-Art, Cooperation Paradigms, Security and Privacy, and Future Trends}
\author{Yuntao~Wang, Yanghe~Pan, Zhou~Su, Yi~Deng, Quan~Zhao, Linkang~Du, Tom~H.~Luan, Jiawen~Kang, and Dusit~Niyato
\thanks{Y.~Wang, Y.~Pan, Z.~Su, Y.~Deng, Q.~Zhao, L.~Du, and T.~H.~Luan are with the School of Cyber Science and Engineering, Xi'an Jiaotong University, Xi'an, China. \textit{(Corresponding author: Zhou~Su)}}
\thanks{J.~Kang is with the School of Automation, Guangdong University of Technology,
 Guangzhou, China.}
\thanks{D.~Niyato is with the College of Computing and Data Science, Nanyang Technological University, Singapore.}}

\maketitle

\begin{abstract}
With the rapid advancement of large models (LMs), the development of general-purpose intelligent agents powered by LMs has become a reality. It is foreseeable that in the near future, LM-driven general AI agents will serve as essential tools in production tasks, capable of autonomous communication and collaboration without human intervention. This paper investigates scenarios involving the autonomous collaboration of future LM agents. We review the current state of LM agents, the key technologies enabling LM agent collaboration, and the security and privacy challenges they face during cooperative operations. To this end, we first explore the foundational principles of LM agents, including their general architecture, key components, enabling technologies, and modern applications. We then discuss practical collaboration paradigms from data, computation, and knowledge perspectives to achieve connected intelligence among LM agents. After that, we analyze the security vulnerabilities and privacy risks associated with LM agents, particularly in multi-agent settings, examining underlying mechanisms and reviewing current and potential countermeasures. Lastly, we propose future research directions for building robust and secure LM agent ecosystems.
\end{abstract}


\IEEEpeerreviewmaketitle
\section{Introduction}
\subsection{Background of Large Model Based Agents}\label{sec:Background}
\begin{figure}[!t]
\centering 
\includegraphics[width=0.78\linewidth]{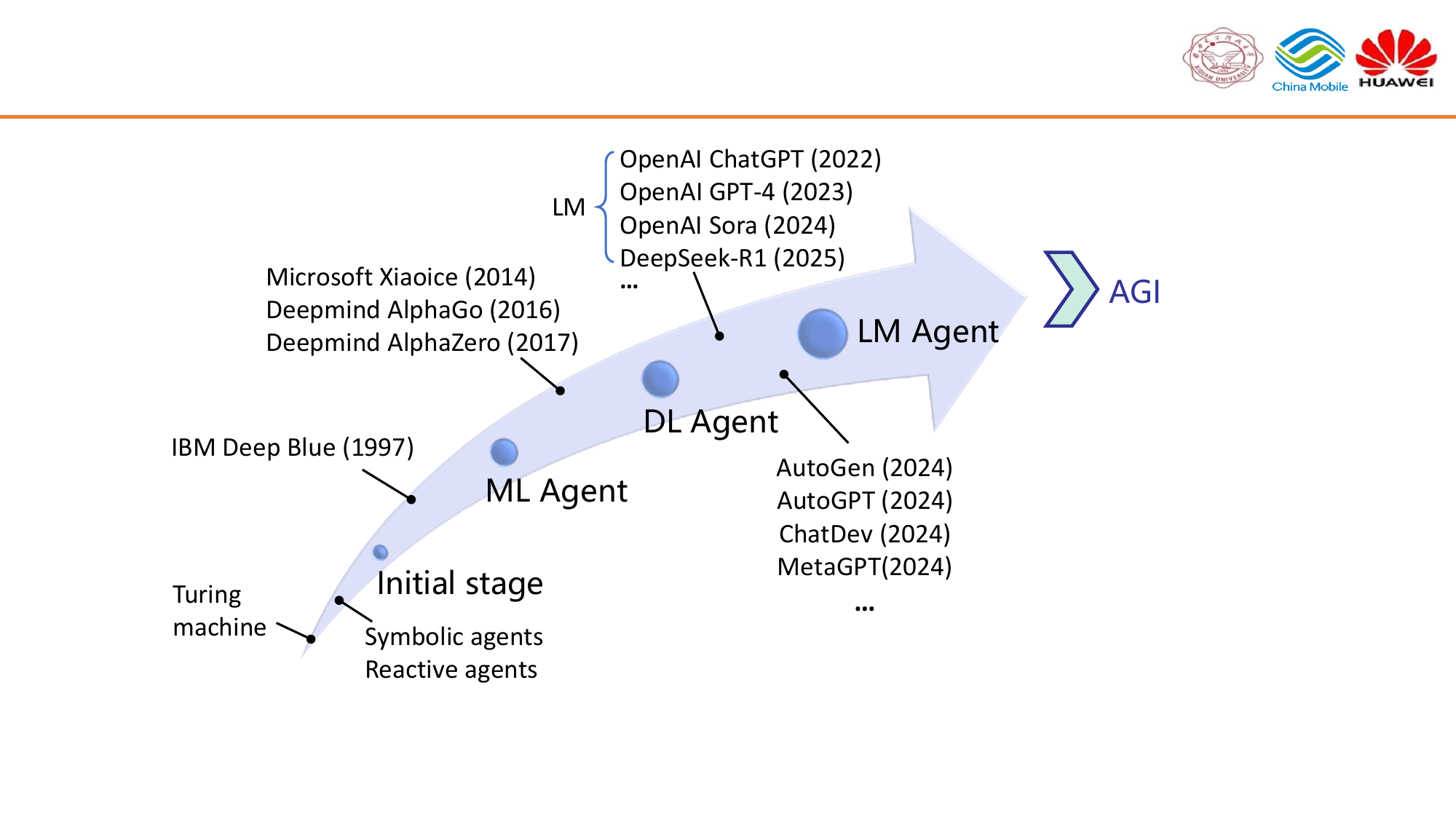}
\caption{Evolution history of AI agents.}\label{fig:history}\vspace{-4mm}
\end{figure}

In the 1950s, Alan Turing introduced the Turing Test to assess whether machines could exhibit intelligence comparable to humans. 
These artificial entities, commonly known as ``agents", serve as the core components of artificial intelligence (AI) systems.
AI agents {(also known as agentic AI \cite{shavit2023practices})} are autonomous entities capable of understanding and responding to human inputs, perceiving their environment, making decisions, and taking actions across physical, virtual, or mixed-reality settings to achieve specific goals \cite{xi2023rise}. 
They can be software-based or physical entities, functioning independently or in collaboration with humans or other agents.
Since the mid-20th century, significant progress has been made in the development of AI agents \cite{ribeiro2002reinforcement,silver2017mastering}, such as Deep Blue, AlphaGo, and AlphaZero, as shown in Fig.~\ref{fig:history}. 
Despite these advances, prior research primarily concentrated on refining specialized abilities such as symbolic reasoning or excelling in certain tasks such as Go or Chess, often neglecting the cultivation of general-purpose capabilities within AI models such as long-term planning, multi-task generalization, and knowledge retention.

\begin{figure}[!t]
\centering 
\includegraphics[width=1.0\linewidth]{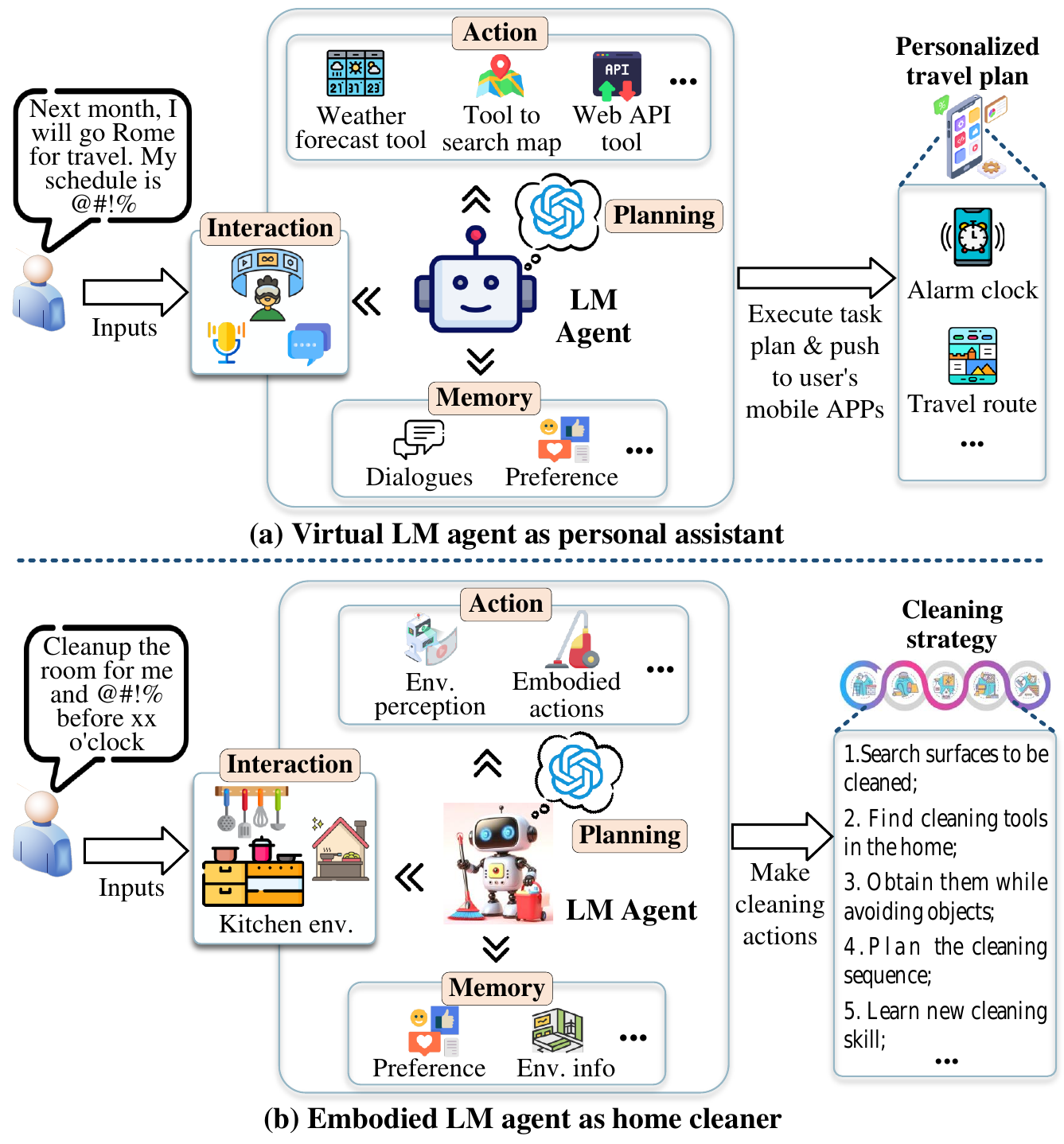}
  \caption{Use cases of LM agents: (a) a virtual LM agent acting as the personal assistant; (b) an embodied LM agent serving as the home cleaner. An LM agent, either in software or embodied form, generally consists of four key components: interaction, planning, action, and memory.}\label{fig:intro_soft_embo}\vspace{-2.5mm}
\end{figure}

With the rise of large models (LMs), including large language models (LLMs) such as OpenAI GPT-4, {DeepSeek-R1}, and Google PaLM 2, as well as multi-modal vision-language models (VLMs) such as Sora, LM-based agents (also known as LM agents or {agentic LMs \cite{paglieri2024balrog}}) are unlocking new possibilities. LM agents significantly enhance the inherent capabilities of AI systems, providing a versatile foundation for the next-generation AI agents \cite{song2023llm}.
Serving as the ``brain" of AI agents, LMs empower them with advanced capabilities in human-machine interaction (HMI), few/zero-shot planning, contextual understanding, knowledge retention, and general-purpose task solving across physical, virtual, or mixed-reality environments \cite{xi2023rise,cheng2024exploring}. {LM agents generally fall into two categories:
\begin{itemize}
    \item \textit{Virtual LM agents}, such as AutoGPT \cite{yang2023auto} and AutoGen \cite{wu2024autogen}, can autonomously interpret human instructions and use various tools (e.g., search engines and external APIs) to gather information and complete intricate tasks \cite{chen2024Re-Invoke}. For instance, as shown in Fig.~\ref{fig:intro_soft_embo}(a), an LM-powered personal assistant can generate personalized travel plans, set reminders, and manage tasks while continuously learning and adapting in dynamic environments.
    \item \textit{Embodied LM agents}, such as FigureAI's Figure 02 and Tesla's Optimus, engage directly with the physical world. These agents perceive and interact with their surroundings, allowing them to solve real-world problems \cite{ichter2022do}. For instance, as shown in Fig.~\ref{fig:intro_soft_embo}(b), an LM-powered household robot analyzes room layouts, surface types, and obstacles, to devise customized cleaning strategies rather than merely following generic instructions.
\end{itemize}}

LM agents are recognized as a significant step towards achieving artificial general intelligence (AGI) and have been widely applied across fields such as web search \cite{nakano2022webgpt}, virtual assistants \cite{wang2024mobileagentv2}, Metaverse gaming \cite{hu2024survey}, robotics \cite{ichter2022do}, autonomous vehicles \cite{jin2023surrealdriver}, and automated attack penetration \cite{deng2024pentestgpt}. 
As reported by MarketsandMarkets \cite{MarketsandMarkets}, the worldwide market for autonomous AI and autonomous agents was valued at USD 4.8 billion in 2023 and is projected to grow at a CAGR of 43\%, reaching USD 28.5 billion by 2028.
LM agents have attracted global attention, and leading technology giants including Google, OpenAI, Microsoft, IBM, AWS, Oracle, NVIDIA, and Baidu are venturing into the LM agent industry.

\begin{figure}[!t]
\centering 
  \includegraphics[width=8.8cm]{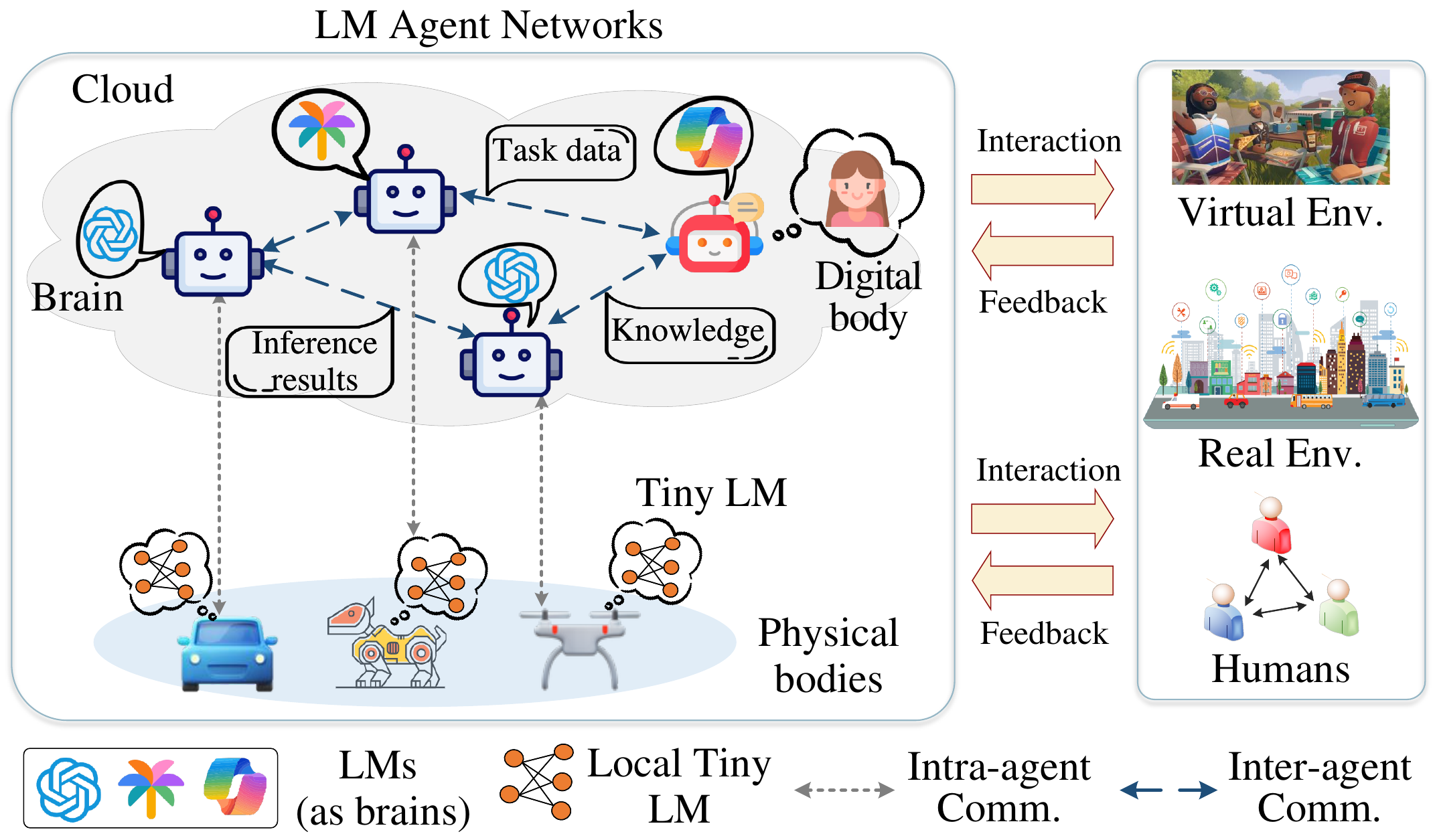}
  \caption{{Overview of LM agent networks. Each LM agent consists of an LM-powered \textit{brain} and a physical or digital \textit{body}. Within each LM agent, the brain synchronizes with its body via \textit{intra-agent} communications. LM agents communicate with one another in the cloud to share task-oriented, knowledge, and inference results via \textit{inter-agent} communications, establishing a network of interconnected intelligence.}}\label{fig:overview}\vspace{-3mm}
\end{figure}

{\begin{table*}[!t]
	\begin{center}\setlength{\abovecaptionskip}{0cm}
		\caption{{Summary of Key Abbreviations in Alphabetical Order}}\label{table-abbr}
		\begin{tabular}{ll|ll|ll}
			\toprule
			\textbf{Abbr.}  &\textbf{Definition}            & \textbf{Abbr.} &\textbf{Definition}           &\textbf{Abbr.} &\textbf{Definition} \\
			\midrule
			ACL   &Agent Communications Language       &AI   &Artificial Intelligence   &AIGC &AI-Generated Content  \\       
                AIS &Adaptive Instructional Systems   &AR/VR/MR  &Augmented/Virtual/Mixed Reality &ASR &Attack Success Rate  \\                 BKI  &Backdoor Keyword Identification &CoT  &Chain-of-Thought     &CPSS &Cyber-Physical-Social Systems \\
                CroPA &Cross-Prompt Attack &DEPS   &Describe-explain-plan-select         &DHTs &Distributed Hash Tables \\
                DoS &Denial-of-Service &DP   &Differential Privacy                        &FFN   &Feed-Forward Network    \\
                FPR &False Positive Rate &GoT &Graph-of-Thought                &HMI &Human-Machine Interaction  \\
                IP  &Intellectual Property &KG &Knowledge Graph          & LFU & Least Frequently Used \\
                LLMs  &Large Language Models  &LMs  &Large Models &LRU &Least Recently Used  \\
                LVMs  &Large Vision Models      &MAD  &Multi-Agent Debate &MARL &Multi-Agent Reinforcement Learning  \\
                MEI  &Mobile Edge Intelligence       &MFA  &Multi-Factor Authentication &MIAs &Membership Inference Attacks  \\
                MITM  &Man-In-The-Middle       &MLMs  &Masked Language Models &MoE &Mixture-of-Experts \\
                MSE  &Mean Squared Error       &NDN  &Named Data Networking &NLP &Natural Language Processing \\
                NMT  &Neural Machine Translation       &NSFW  &Not-Safe-For-Work &OS &Operating System  \\
                PEFT  &Parameter-Efficient Fine-Tuning       &PII  &Personally Identifiable Information &PoT &Plan-of-Thought \\
                PPI &Personal Preference Information   &QoS &Quality-of-Service &RAG &Retrieval-Augmented Generation  \\
                ROS &Robot Operating System &SKGR &Self-Knowledge-Guided Retrieval &SL &Split Learning \\
                SPP &Solo Performance Prompting &TMM &Transferable Multi-Modal &ToB &To-Business \\
                ToC &To-Customer  &ToT &Tree-of-Thought &TPSP &Third-Party Service Provider \\
                SHAP &Shapley Additive Explanations &VLMs &Vision-Language Models &XAI &Explainable Artificial Intelligence \\
			\bottomrule
		\end{tabular}
	\end{center}
\end{table*}}

\subsection{{Motivation of LM Agent Networks}}\label{sec:Roadmap}

Beyond the intelligence of single LM agents, connected LM agents can form LM agent networks to cooperatively address complex tasks beyond the capacity of individual agents \cite{10638533}.
As depicted in Fig.~\ref{fig:overview}, within LM agent networks, connected LM agents can freely share sensory data, task-oriented knowledge, and LM inference results. By leveraging collective computational power and shared expertise from specialized LM agents, it fosters cooperative decision-making and collective intelligence. For instance, in autonomous driving, connected autonomous vehicles, acting as LM agents, share real-time sensory data, coordinate movements, and negotiate passage at intersections to optimize traffic flow and enhance safety.
{Each LM agent includes two parts: (i) the \textit{brain} located in the cyberspace powered by LMs such as GPT-4o, PaLM 2, and DeepSeek-R1; and (ii) the physical or digital \textit{body} such as autonomous vehicle, robot dog, UAV, and digital human.} 
For embodied LM agents, a tiny local LM within the physical body handles local inference, while compute-intensive tasks are offloaded to the cloud LM, enabling collaborative computation between the cloud brain and the local brain.
Each LM agent can interact with other agents, virtual/real environments, and humans. {The brain of each LM agent can be deployed either as a \textit{standalone} entity or in a \textit{hierarchical} manner across various platforms such as cloud servers, edge servers, or end devices.} Communication within LM agent networks occurs via two primary modes: (i) \textit{intra-agent communications} for seamless synchronization of status, data, and knowledge between the cyber brain and physical/digital body to enable real-time functionality; and (ii) \textit{inter-agent communications} for efficient information exchange and computational coordination among multiple LM agents.

{LM agent networks represent a transformative leap in intelligent systems, which incorporate a range of cutting-edge technologies as its foundation. Particularly, foundation models enable cognitive capabilities through advanced reasoning and planning, knowledge-related technologies integrate external knowledge sources for context-aware actions, HMI technologies ensure seamless human-agent interactions via natural language processing (NLP) and augmented/virtual/mixed reality (AR/VR/MR), digital twin technologies synchronize physical and digital states through intra-agent communications, and multi-agent collaboration technologies foster efficient teamwork via inter-agent communications.
However, their full potential faces key challenges, including but not limited:
\begin{itemize}
    \item \textit{Dynamic Construction of LM Agent Networks.} Constructing dynamic LM agent networks requires addressing the versatility-efficiency-portability trilemma \cite{marro2024scalable}. The network should support diverse tasks and applications (versatility), achieve high resource utilization with low cost (efficiency), and operate across heterogeneous platforms and environments (portability).
    The inherent spatiotemporal dynamics introduce additional complexity. From temporal perspectives, agents' roles and behaviors evolve over time, necessitating adaptive reconfiguration mechanisms to ensure continuity in long-term operations. From spatial perspectives, the topologies of LM agent networks change due to mobility and environmental changes, requiring robust coordination protocols to maintain network coherence.
    \item \textit{Collaborative LM Service Provisioning in Heterogeneous Networks.} Once constructed, LM agent networks should enable heterogeneous agents to collaboratively provide LM services. Given the substantial computational resources required to run a complete LM, edge or terminal agents often lack the capacity to operate a full-scale LM. There exist two primary strategies: (i) \textit{LM lightweighting}, e.g., knowledge distillation \cite{zhang2024vpgtrans}, quantization \cite{zhu2024surveymodelcompress}, pruning \cite{ma2023llm}, and hardware acceleration \cite{shen2024agile}, for deploying compact LM variants tailored to individual agents' computational capacities. Collaboration is facilitated through methods such as role-playing for task-specific role assignment and distributed consensus for coordinated decision-making. (ii) \textit{LM sharding}, e.g., split learning (SL) \cite{10648594} and mixture-of-experts (MoE) \cite{fedus2022switch}, for distributed LM reconstruction, enabling agents to share workloads efficiently. The full-scale LM is segmented into smaller, manageable shards distributed across agents, allowing each edge or terminal agent to handle a portion of the model that aligns with its resource capabilities.
    \item \textit{Autonomous Optimization and Secure Collaboration.} Dynamic resources (e.g., data, knowledge, and computational power) allocation is vital to maximize throughput, minimize latency, and maintain seamless communication across cloud-edge-end layers.
    Security mechanisms in collaboration are also crucial to ensure trustworthiness (i.e., LM agents operate securely and transparently in collaborative tasks) and privacy preservation (i.e., safeguarding sensitive data during agent interactions).
\end{itemize}}

\subsection{Motivation of Securing LM Agent Networks}\label{sec:Challenges}

Despite the promising future of connected LM agents, security and privacy concerns pose significant barriers to their widespread adoption. Throughout the life-cycle of LM agents, numerous vulnerabilities can emerge, ranging from adversarial examples \cite{DBLP:conf/eacl/ZhuoLHSWHL23}, agent poisoning \cite{zou2024poisonedrag}, LM hallucination \cite{zhang2023siren}, to pervasive data collection and memorization risks \cite{DBLP:conf/iclr/CarliniIJLTZ23}.

\emph{1) Security/reliability vulnerabilities.}
LM agents are susceptible to ``hallucinations", where their LMs generate outputs that are plausible but incorrect and not grounded in reality, potentially spreading misinformation in multi-agent environments \cite{zhang2023siren}.
{Hallucinations can undermine the reliability of decision-making,} cause task failures, and pose risks to both physical systems and human safety. Ensuring the integrity of intermediate inference results in collaborative systems is crucial, {as biased or manipulated inputs can produce unfair or erroneous outcomes} \cite{liu2023prompt}.
{Additionally,} attacks such as adversarial manipulations \cite{DBLP:conf/eacl/ZhuoLHSWHL23}, poisoning \cite{fang2020local}, and backdoors \cite{rando2024universal} exacerbate these vulnerabilities by allowing malicious actors to manipulate inputs and {deceive models.
In collaborative settings, agent poisoning behaviors can undermine the collaborative systems} \cite{zou2024poisonedrag}, as malicious agents can disrupt the operation of others. {Furthermore,} integrating LM agents into cyber-physical-social systems (CPSS) expands the attack surface, {increasing opportunities for malicious exploitation} within interconnected systems.

\emph{2) Privacy breaches.}
The extensive data collection processes and data memorization capabilities of LM agents pose significant {privacy risks related to data misuse} and unauthorized access. {These agents often handle} large volumes of personal and {business-sensitive data} in both to-customer (ToC) and to-business (ToB) applications, {raising concerns about secure storage,} processing, and sharing \cite{DBLP:conf/uss/CarliniTWJHLRBS21}. Furthermore, LMs can inadvertently memorize sensitive information from training data or past interactions, {potentially exposing them in} subsequent interactions \cite{DBLP:conf/iclr/CarliniIJLTZ23}. These privacy risks are exacerbated in multi-agent collaborations, where LM agents may unintentionally leak {confidential user data, internal operational details, or proprietary business information} during communication and task execution.

\subsection{Related Surveys and Our Contributions}\label{subsec:Contributions}
Recently, LM agents have garnered significant interest across academia and industry, leading to a variety of research exploring their potential from multiple perspectives. Notable survey papers in this field are as below.
Andreas \textit{et al.} \cite{andreas2022language} present a toy experiment for AI agent construction and case studies on modeling communicative intentions, beliefs, and desires.
Wang \textit{et al.} \cite{wang2023survey} identify key components of LLM-based autonomous agents (i.e., profile, memory, planning, and action) and the subjective and objective evaluation metrics. Besides, they discuss the applications of LLM agents in engineering, natural science, and social science.
Xi \emph{et al.} \cite{xi2023rise} present a general framework for LLM agents consisting of brain, action, and perception. Besides, they explore applications in single-agent, multi-agent, and human-agent collaborations, as well as agent societies.
Xu \emph{et al.} \cite{10398474} provide a tutorial on key concepts, architecture, and metrics of edge-cloud AI-generated content (AIGC) services in mobile networks, and identify several use cases and implementation challenges.

Cheng \emph{et al.} \cite{cheng2024exploring} review key components of LLM agents (including planning, memory, action, environment, and rethinking) and their potential applications. Planning types, multi-role relationships, and communication methods in multi-agent systems are also reviewed.
Guo \emph{et al.} \cite{guo2024large} discuss the four components (i.e., interface, profiling, communication, and capabilities acquisition) of LLM-based multi-agent systems and present two lines of applications in terms of problem solving and world simulation.
Durante \emph{et al.} \cite{durante2024agent} introduce multimodal LM agents and a training framework including learning, action, congnition, memory, action, and perception. They also discuss the different roles of agents (e.g., embodied, simulation, and knowledge inference), as well as the potentials and experimental results in different applications including gaming, robotics, healthcare, multimodal tasks, and NLP.
Hu \emph{et al.} \cite{hu2024survey} outline six key components (i.e., perception, thinking, memory, learning, action, and role-playing) of LLM-based game agents and review existing LLM-based game agents in six types of games.
Qu \emph{et al.} \cite{qu2024mobileedgeintelligencelarge} provide a comprehensive survey on integrating mobile edge intelligence (MEI) with LLMs, emphasizing key applications of deploying LLMs at the network edge along with state-of-the-art techniques in edge LLM caching, delivery, training, and inference.

{For the security and privacy of LLMs and agents, Yao \emph{et al.} \cite{YAO2024100211} provide a comprehensive review of LLMs' dual role in cybersecurity, examining their potential to enhance security and privacy (``the good"), the risks and threats they introduce (``the bad"), and their inherent vulnerabilities (``the ugly"). 
Friha \emph{et al.} \cite{friha2024llm} comparatively review recent optimization and autonomy methods for resource-constrained edge environments, while reviewing security concerns, trust issues, and mitigation strategies for LLM-powered MEI.
Das \emph{et al.} \cite{Das2025Security} explore the security and privacy challenges of LLMs, analyzing application-specific risks across various domains, including transportation, education, and healthcare. They also evaluate the scope of LLM vulnerabilities, emerging attack vectors, and potential defense strategies.} 

Existing surveys on LM agents primarily focus on the general framework design of LLM agents and multi-agent systems, as well as their potential in specific applications. 
Distinguished from the above-mentioned existing surveys, this survey focuses on the networking aspect of LM agents, including the general architecture, enabling technologies, key characteristics, and collaboration paradigms to construct networked systems of LM agents within physical, virtual, or mixed-reality environments. Additionally, with the advances of LM agents, it is urgent to examine their security and privacy in future AI agent systems. This work comprehensively reviews the security and privacy issues of LM agents and discusses both existing and potential defense mechanisms, which are overlooked in previous surveys.
Table~\ref{contribution} compares the contributions of our survey with previous related surveys in the field of LM agents.

\begin{table}[!t]
   \centering 
    \caption{A Comparison of Our Survey with Relevant Surveys}\label{contribution}
    \resizebox{1.01\linewidth}{!}{
        \begin{tabular}{|c|c|l|}
        \hline
        \textbf{Year.} &\textbf{Refs.} &\textbf{Contribution} \\ \hline 
        {2022} &\cite{andreas2022language} &\tabincell{l}{A toy experiment for AI agent construction and case studies \\in modeling communicative intentions, beliefs, and desires.} \\ \hline

        {2023} &\cite{wang2023survey} &\tabincell{l}{Survey on key compoments and evaluation policies of LLM \\agents, and applications in engineering, natural science, and \\social science.} \\ \hline

        {2023} &\cite{xi2023rise} &\tabincell{l}{Discussions on general framework for LLM agents and \\applications of single-agent, multi-agent, and human-agent \\collaborations, as well as agent societies.} \\ \hline


{2024} &\cite{10398474} &\tabincell{l}{Tutorial on key concepts, architecture, and metrics of \\edge-cloud AIGC services in mobile networks, and \\identify use cases and key implementation challenges.} \\ \hline



        {2024} &\cite{cheng2024exploring} &\tabincell{l}{Discuss key components and applications of LLM agents, \\and review planning types, multi-role relationships, and \\communication modes in multi-agent systems.} \\ \hline


        {2024} &\cite{guo2024large} &\tabincell{l}{Discuss key components of LLM-based multi-agent systems \\and applications in problem solving and world simulation.} \\ \hline

        {2024} &\cite{durante2024agent} &\tabincell{l}{Discuss key concepts and the framework of multimodal LM \\agents, the different roles of agents, and the potentials and \\experimental results in gaming, robotics, healthcare, NLP, \\and multimodality.} \\ \hline

         {2024} &\cite{hu2024survey} &\tabincell{l}{Discuss key components of LLM-based game agents and \\review existing approaches in six types of games.} \\ \hline


        {2024} &\cite{qu2024mobileedgeintelligencelarge} &\tabincell{l}{Survey on key applications of deploying LLMs at network \\edges and state-of-the-art techniques in edge LLM caching, \\delivery, training, and inference.} \\ \hline

        {{2024}} &{\cite{YAO2024100211}} &{\tabincell{l}{Survey on LLMs' potential to enhance security \& privacy,\\risks they introduce, and their inherent vulnerabilities.}} \\ \hline

        {{2024}} & {\cite{friha2024llm}} &{\tabincell{l}{Survey on optimization and autonomy techniques for edge\\environments, review security concerns, trust issues, and\\mitigation strategies for LLM-powered MEI applications.}} \\ \hline

        {{2025}} &{\cite{Das2025Security}} &{\tabincell{l}{Explore security and privacy challenges of LLMs, analyze\\application-specific LLM risks, and evaluate the scope of\\LLM vulnerabilities, emerging attack vectors, and potential\\ defense strategies.}} \\ \hline


        {Now} &\textbf{Ours} &\tabincell{l}{Comprehensive survey of the fundamentals, security, and \\privacy of connected LM agents, discussions on the general \\architecture, enabling technologies, key characteristics, and \\cooperation paradigms of connected LM agents, discussions \\on security/privacy threats, state-of-the-art countermeasures, \\and open research issues in future LM agent systems.} \\ \hline
        \end{tabular}}
\end{table}

\begin{figure}[!t]
\centering \setlength{\abovecaptionskip}{0.cm}
  \includegraphics[height=24.cm,width=9.2cm]{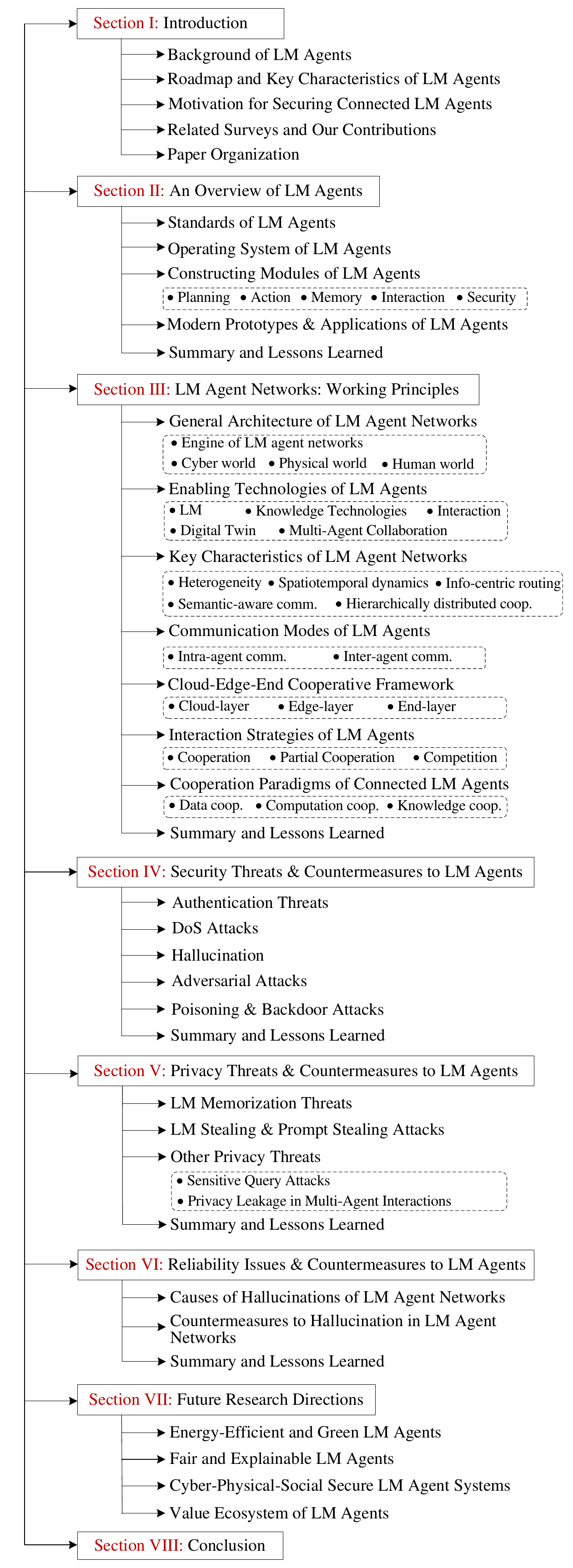}
  \caption{{Organization structure of this paper.}}\label{fig:organization}\vspace{-5mm}
\end{figure}

In this paper, we present a systematic review of the state-of-the-arts in both single and connected LM agents, focusing on cooperation paradigms, security and privacy threats, existing and potential countermeasures, and future trends.
Our survey aims to 1) provide a broader understanding of how LM agents work and how they cooperate in multi-agent scenarios, 2) examine the scope and impact of security, privacy, and reliability challenges associated with LM agents and their interactions, and 3) highlight effective strategies and solutions for defending against these threats to safeguard LM agents in various intelligent applications.
The main contributions of this work are four-fold.
\begin{itemize}
  \item We comprehensively review recent advances in LM agent construction across academia and industry. We investigate the fundamentals of LM agents, including the general architecture, and key components (i.e., planning, memory, action, interaction, and security modules). Industrial prototypes and potential applications of LM agents are also discussed.
  \item We discuss key characteristics and enabling technologies of LM agent networks and explore practical cloud-edge-end collaboration paradigms from the aspects of data cooperation, computation cooperation, and knowledge cooperation. We systematically discuss interaction strategies of LM agents (i.e., cooperation, partial cooperation, and competition). 
  \item We comprehensively analyze existing and potential security, privacy, and reliability threats, their underlying mechanisms, categorization, and challenges for both single and connected LM agents. We also review state-of-the-art countermeasures and examine their feasibility in securing LM agent networks.
  \item Lastly, we discuss open research opportunities and point out future research directions of LM agent networks, aiming to inspire more ongoing research and innovations in this field. 
\end{itemize}

\subsection{Paper Organization}\label{subsec:organization}
The remainder of this paper is organized as below. Section~\ref{sec:SINGLEAGENT} gives an overview of single LM agents, while Section~\ref{sec:Fundamentals} presents the fundamentals for networking LM agents.
Section~\ref{sec:Security} and Section~\ref{sec:Privacy} introduce the taxonomy of security and privacy threats to LM agents, respectively, along with state-of-the-art countermeasures. Section~\ref{sec:FUTUREWORK} outlines open research issues and future directions in the field of LM agents. Finally, conclusions are drawn in Section~\ref{sec:CONSLUSION}. Fig.~\ref{fig:organization} depicts the organization structure of this survey, {and Table~\ref{table-abbr} summarizes the key acronyms}.

\begin{table*}[!t]
\centering \setlength{\abovecaptionskip}{0cm}
    \caption{Progress of Standards for LM Agents}\label{Roadmap}
\begin{tabular}{|c|c|c|}
\hline
\textbf{Standard} & \textbf{Publication Date} & \textbf{Main Content}\\ \hline
IEEE SA - P3394 & 2023-09-21 & \begin{tabular}[c]{@{}c@{}}The natural language interface is defined, including various protocols and guidelines to facilitate seamless \\and efficient interaction between APPs, agents, and LLM-powered systems.\end{tabular} \\ \hline
IEEE SA - P3428 & 2023-12-06 & \begin{tabular}[c]{@{}c@{}}The integration of LLMs with existing education systems to ensure that LLMs can seamlessly interact with \\AIS while addressing issues of bias, transparency, and accountability in educational environments.\end{tabular} \\ \hline
\end{tabular}
\end{table*}

\section{An Overview of Large Model Agents}\label{sec:SINGLEAGENT}
\begin{figure}[!t]
\centering \setlength{\abovecaptionskip}{-0.cm}
  \includegraphics[width=9.8cm]{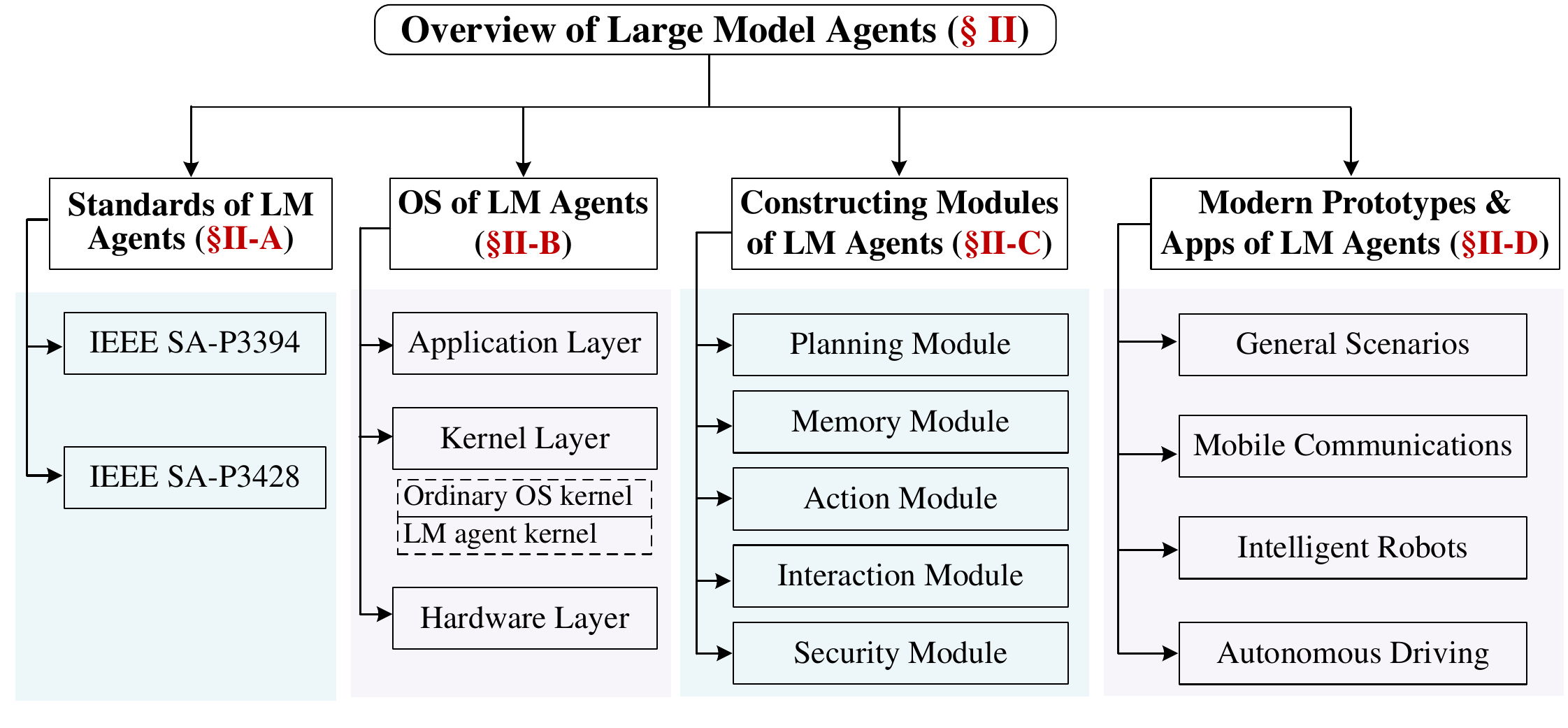}
  \caption{{Organization structure of the Section~\ref{sec:SINGLEAGENT}.}}\label{fig:S2architecture}\vspace{-3mm}
\end{figure}

In this section, we first introduce existing standards of LM agents. Then, we discuss the operating system (OS) and constructing modules of LM agents.
Next, we introduce typical prototypes and discuss modern applications of LM agents. {Fig.~\ref{fig:S2architecture} shows the organization structure of this section.}

\subsection{Standards of LM Agents}\label{subsec:Standards}
We briefly introduce two existing standards on LM agents: IEEE SA-P3394 and IEEE SA-P3428.

\textit{1) The IEEE SA - P3394 standard\footnote{https://standards.ieee.org/ieee/3394/11377/}}, launched in 2023, defines natural language interfaces to enhance communication between LLM applications, agents, and human users. It establishes a series of protocols and guidelines to facilitate seamless and efficient interaction between applications, agents, and LLM-powered systems. These protocols and guidelines include, but are not limited to, API syntax and semantics, voice and text formats, conversation flow, prompt engineering integration, LLM thought chain integration, as well as API endpoint configuration, authentication, and authorization for LLM plugins. The standard is expected to advance technological interoperability, promote AI industry development, enhance the practicality and efficiency of LMs, and improve AI agents' functionality and user experience.

\textit{2) The IEEE SA - P3428 standard\footnote{https://standards.ieee.org/ieee/3428/11489/}}, launched in 2023, aims to develop standards for LLM agents in educational applications. The primary goal is to ensure the interoperability of LLM agents across both open-source and proprietary systems. Key areas of focus include the integration of LLMs with existing educational systems and addressing technical and ethical challenges. This includes ensuring that LLMs can seamlessly interact with other AI components, such as adaptive instructional systems (AIS), while addressing issues of bias, transparency, and accountability within educational contexts. The standard is intended to support the effective LLMs-empowered educational applications, thereby enabling more personalized, efficient, and ethically sound AI-driven educational experiences.

\begin{figure}[!t]
\centering 
  \includegraphics[width=9.5cm]{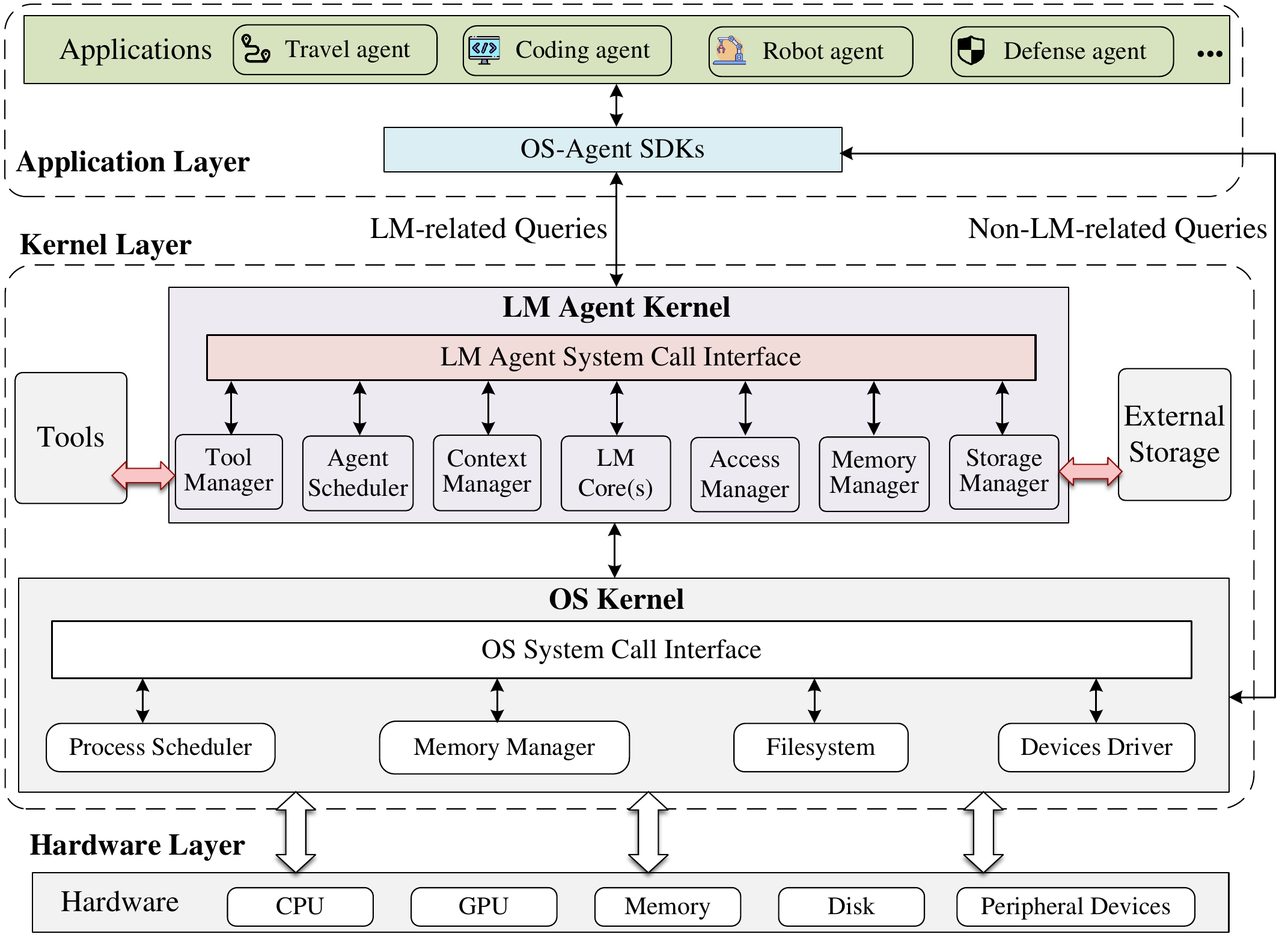}
  \caption{{Illustration of the OS architecture of LM agents \cite{mei2024aios}.}}\label{fig:OSarchitecture}\vspace{-3mm}
\end{figure}

\subsection{Operating System (OS) of LM Agents}
According to \cite{mei2024aios}, the OS architecture of LM agents consists of three layers: application, kernel, and hardware, as shown in Fig.~\ref{fig:OSarchitecture}.
\begin{itemize}
    \item The \textit{application layer} hosts agent applications (e.g., travel, coding, and robot agents) and offers an SDK that abstracts system calls, simplifying agent development. Typical agent developing frameworks include LangChain, AutoGen, Dify, and LlamaIndex.
    \item The \textit{kernel layer} includes the ordinary \textit{OS kernel} (to process LM-related queries such as file operation and network request) and an additional \textit{LM agent kernel} (to process LM-related queries such as reasoning, planning, and tool parsing), with a focus on without altering the original OS structure. Key modules in the LM agent kernel \cite{mei2024aios} include the agent scheduler for task planning and prioritization, context manager for LM status management, memory manager for short-term data, storage manager for long-term data retention, tool manager for external API interactions, and access manager for privacy controls.
    \item The \textit{hardware layer} comprises physical resources (CPU, GPU, memory, etc.), which are managed indirectly through OS system calls, as LM kernels do not interact directly with the hardware.
\end{itemize}

\begin{figure}[!t]
\centering 
  \includegraphics[width=7.7cm]{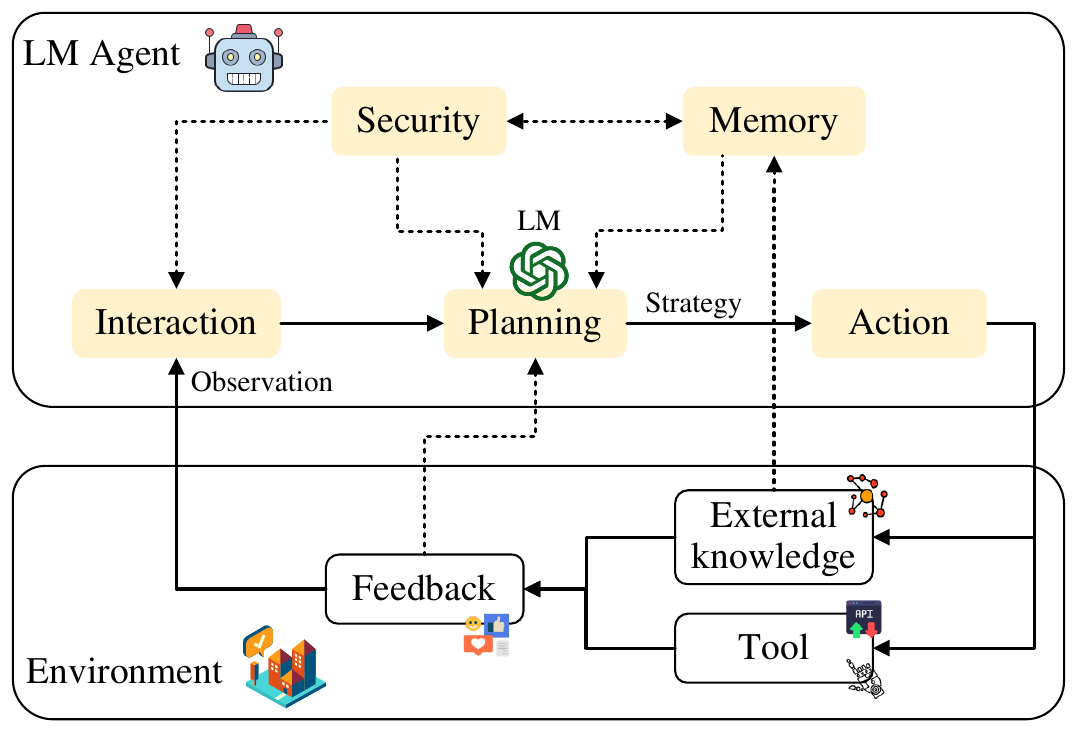}
  \caption{Workflow of the five constructing modules of LM agents in interacting with the real/virtual environment.}\label{fig:AgentWorkflow}\vspace{-3mm}
\end{figure}

\subsection{Constructing Modules of LM Agents}\label{ConstructingModules}
According to \cite{xi2023rise,cheng2024exploring}, there are generally five constructing modules of LM agents: planning, action, memory, interaction, and security. 
As depicted in Fig.~\ref{fig:AgentWorkflow}, these modules together enable LM agents to perceive, plan, act, learn, and interact efficiently and securely in complex and dynamic environments.
Particularly, empowered by LMs, the \textit{planning module} produces strategies and action plans with the help of the memory module, enabling informed decision-making.
The \textit{action module} executes these embodied actions, adapting actions based on real-time environmental feedback to ensure contextually appropriate responses.
The \textit{memory module} serves as a repository of accumulated knowledge (e.g., past experiences and external knowledge), facilitating continuous learning and improvement.
The \textit{interaction module} enables effective communication and collaboration with humans, other agents, and environment.
The \textit{security module} is integrated throughout LM agents' operations, ensuring active protection against threats and maintaining integrity and confidentiality of data and processes.

\begin{figure}[!t]
\centering 
  \includegraphics[width=9.4cm]{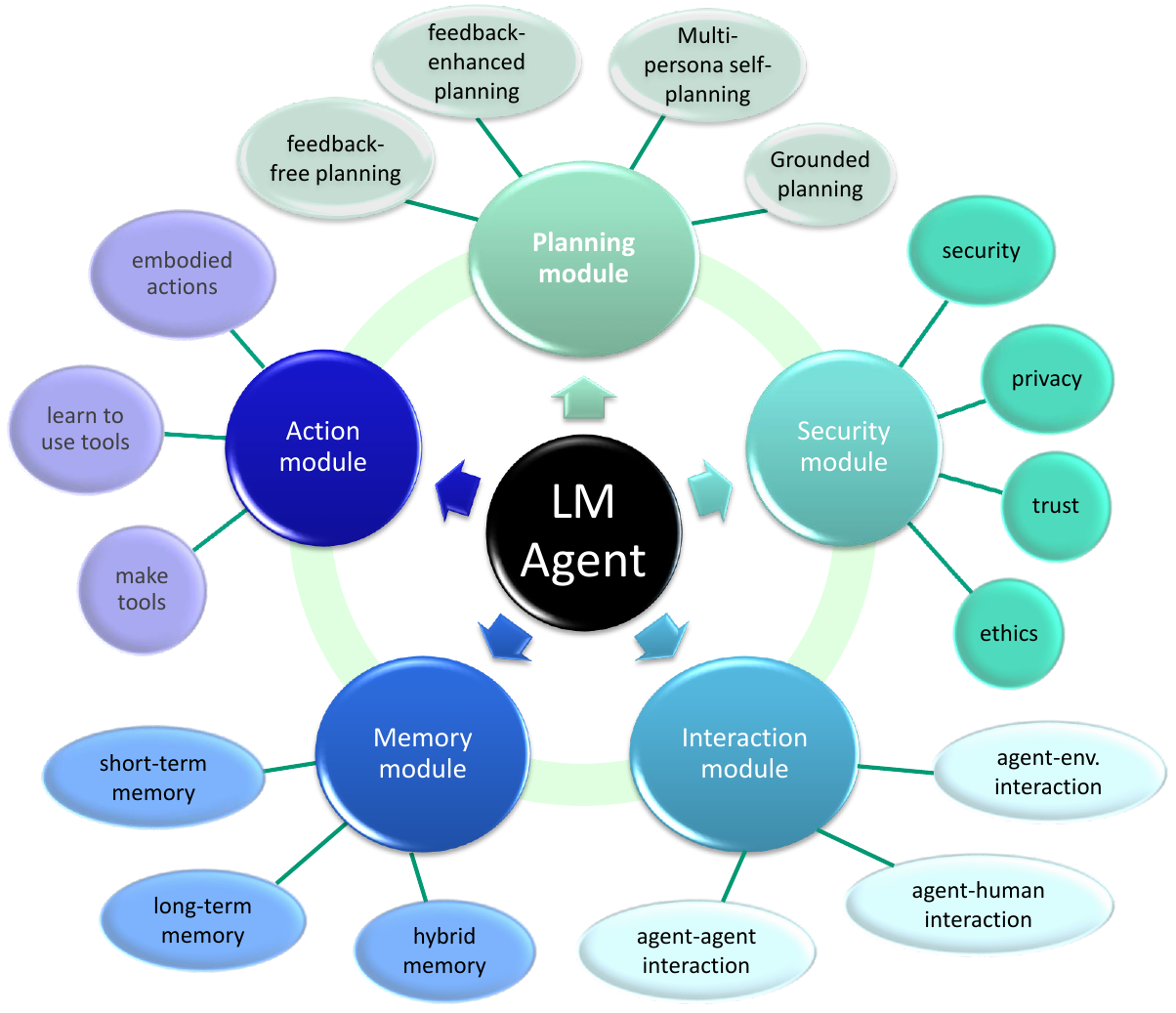}
  \caption{Illustration of five constructing modules (i.e., planning, action, memory, interaction, and security) of connected LM agents including their key components.}\label{fig:modules}\vspace{-2mm}
\end{figure}

\begin{figure*}[!t]
\centering \setlength{\abovecaptionskip}{-0.cm}
  \includegraphics[width=13cm]{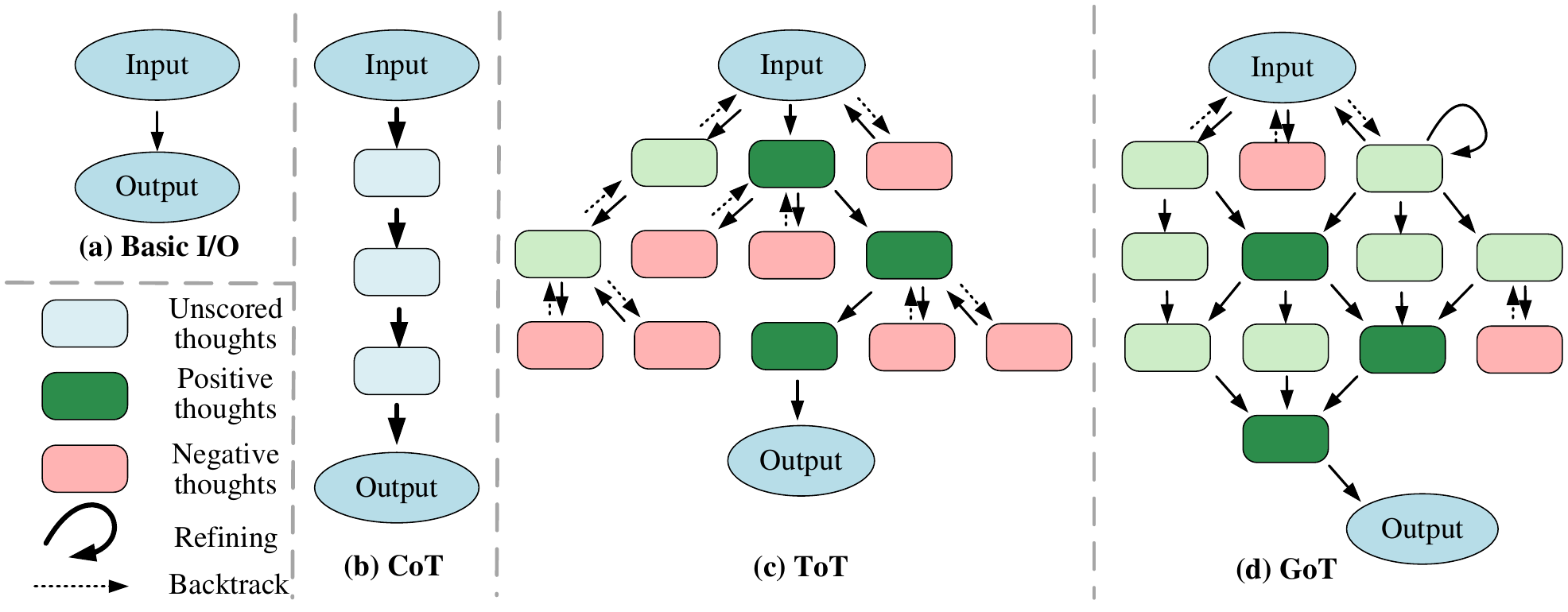}
  \caption{{Illustration of typical feedback-free planning methods of LMs: (a) basic I/O; (b) chain-of-thought (CoT) \cite{wei2023cot}; (c) tree-of-thought (ToT) \cite{yao2023tot}; (d) graph-of-thought (GoT) \cite{besta2024graph}.}}\label{fig:feedback-freeplanning}\vspace{-2mm}
\end{figure*}

\begin{figure}[!t]
\centering 
  \includegraphics[width=5.6cm]{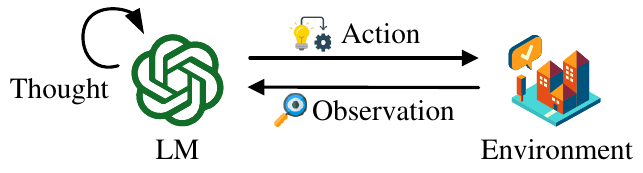}
  \caption{{Illustration of ReAct \cite{yao2023ReAct}: a typical feedback-enhanced planning methods of LMs.}}\label{fig:ReAct}\vspace{-2mm}
\end{figure}

\begin{figure}[!t]
\centering 
  \includegraphics[width=1.02\linewidth]{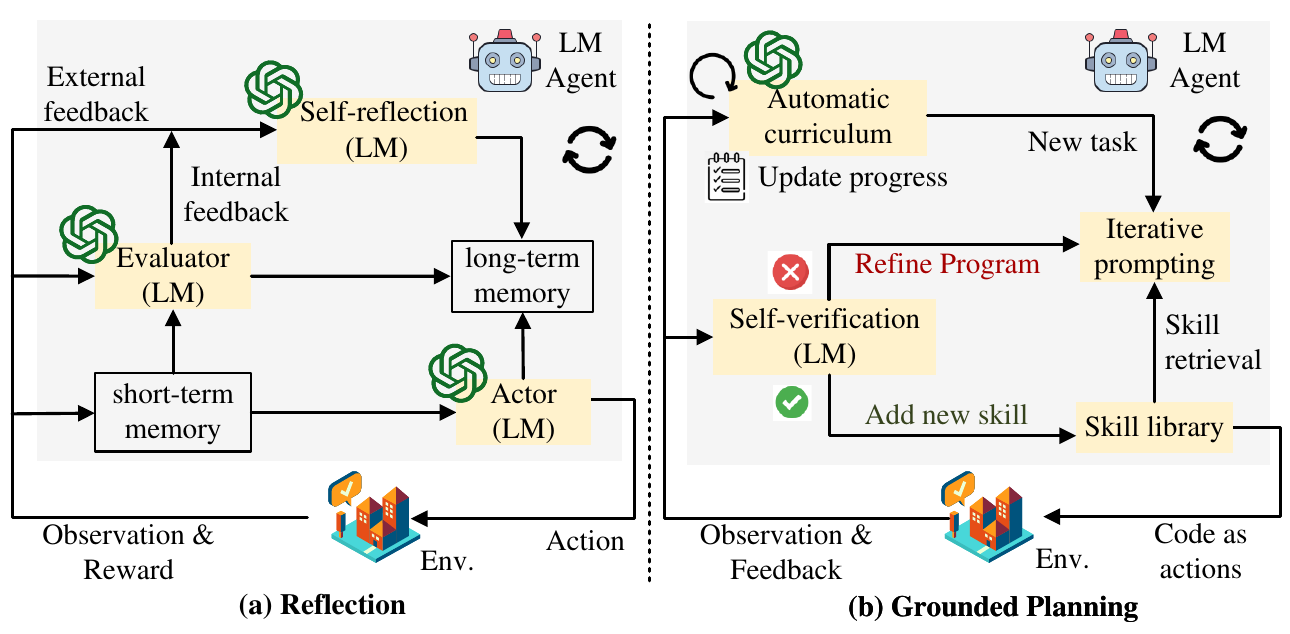}
  \caption{{Illustration of typical feedback-enhanced planning methods of LMs: (a) Reflexion \cite{shinn2023Reflexion}; (b) ground planning \cite{wang2023voyager}.}}\label{fig:Reflection+GroundedPlan}\vspace{-2mm}
\end{figure}

\subsubsection{Planning Module}
The planning module serves as the core of an LM agent \cite{song2023llm,cheng2024exploring}. It utilizes advanced reasoning techniques to devise effective solutions {for} agents to complex problems.
It includes the following working modes.
\begin{itemize}
    \item \textit{Feedback-free planning:}
    The planning module enables LM agents to understand the complex problems and find reliable solutions by breaking them down into necessary steps or manageable sub-tasks \cite{song2023llm,yang2023auto}, as shown in Fig.~\ref{fig:feedback-freeplanning}.
    For instance, chain-of-thought (CoT) \cite{wei2023cot} is a popular sequential reasoning approach where each thought builds directly on the previous one. It represents the step-by-step logical thinking and can enhance the generation of coherent and contextually relevant responses.
    Tree-of-thought (ToT) \cite{yao2023tot} organizes reasoning as a tree-like structure, exploring multiple paths simultaneously. In ToT, each node represents a partial solution, allowing the model to branch and backtrack to find the optimal answer.
    Graph-of-thought (GoT) \cite{besta2024graph} models reasoning using an arbitrary graph structure, allowing more flexible information flow. GoT captures complex relationships between thoughts, enhancing the model's problem-solving capabilities. AVIS \cite{hu2023AVIS} further refines the tree search process for visual QA tasks by leveraging a human-defined transition graph and improves decision-making through a dynamic prompt manager.

    \item \textit{Feedback-enhanced planning:}
    To make effective long-term planning in complex tasks, it is necessary to iteratively reflect on and refine execution plans based on past actions and observations
    \cite{wang2023survey}. The goal is to correct past errors and improve final results. For instance, as illustrated in Fig.~\ref{fig:ReAct}, ReAct \cite{yao2023ReAct} integrates reasoning and action by prompting LLMs to concurrently produce reasoning traces (i.e., thoughts) and actions. This dual approach allows the LLM to create, monitor, and adjust action plans, while task-specific actions enhance interaction with external sources, improving response accuracy and reliability. As shown in Fig.~\ref{fig:Reflection+GroundedPlan}(a),
    Reflexion \cite{shinn2023Reflexion} converts environmental feedback into self-reflection and enhances ReAct by enabling LLM agents to learn from past errors and iteratively optimize behaviors. Reflexion features an actor that produces actions and text via models (e.g., CoT and ReAct) enhanced by memory, an evaluator that scores outputs using task-specific reward functions, and self-reflection that generates verbal feedback to improve the actor.

    \item \textit{Multi-persona self-planning:} Inspired by pretend play, Wang \textit{et al.} \cite{wang2024unleashing} develop a cognitive synergist that enables a single LLM to split into multiple personas, facilitating self-collaboration for solving complex tasks. They propose solo performance prompting (SPP), where LLM identifies, simulates, and collaborates with diverse personas, such as domain experts or target audiences, without external retrieval systems. SPP enhances problem-solving by allowing LLM to perform multi-turn self-revision and feedback from various perspectives.

    \item \textit{Grounded planning:} Executing plans in real or simulated world environments (e.g., Minecraft) requires precise, multi-step reasoning. As shown in Fig.~\ref{fig:Reflection+GroundedPlan}(b), VOYAGER \cite{wang2023voyager}, the first LLM-powered agent in Minecraft, utilizes in-context lifelong learning to adapt and generalize skills to new tasks and worlds. VOYAGER incorporates an automated curriculum for exploration, a skill library containing executable code for complex behaviors, and an iterative prompting mechanism that refines programs based on feedback. Wang \textit{et al.} \cite{wang2023describe} further propose an interactive describe-explain-plan-select (DEPS) planning approach that improves LLM-generated plans by integrating execution descriptions, self-explanations, and a goal selector that ranks sub-goals to refine planning. 
\end{itemize}

\subsubsection{Memory Module}
The memory module is integral to LM agent's ability to learn and adapt over time \cite{wang2023survey}. It maintains an internal memory that accumulates knowledge from past interactions, thoughts, actions, observations, and experiences with users, other agents, and the environments. The stored information guides future decisions and actions, allowing the agent to continuously refine its knowledge and skills. This module ensures that the agent can remember and apply past lessons to new situations, thereby improving its long-term performance and adaptability \cite{cheng2024exploring}. There are various memory formats such as embedded vectors, databases, and structured lists. Additionally, retrieval-augmented generation (RAG) technologies \cite{lewis2020retrieval} are employed to access external knowledge sources, further enhancing the accuracy and relevance of LM agent's planning capabilities.
In the literature \cite{wang2023survey,cheng2024exploring}, memory can be divided into the following three types.
\begin{itemize}
    \item \textit{Short-term memory} focuses on the contextual information of the current situation. It is temporary and limited, typically managed through a context window that restricts the amount of information the LM agent can learn at a time \cite{kang2024knowledge}.

    \item \textit{Long-term memory} stores LM agent's historical behaviors and thoughts. This is achieved through external vector storage, which allows for quick retrieval of important information, ensuring that the agent can access relevant past experiences to inform current decisions \cite{trivedi2022interleaving}.

    \item \textit{Hybrid memory} synergizes short-term and long-term memory to enhance an agent's understanding of the current context and leverage past experiences for better long-term reasoning.
    Liu \textit{et al.} \cite{liu2024RAISE} propose the RAISE architecture to enhance ReAct for conversational AI agents by integrating a dual-component memory system, where Scratchpad captures recent interactions as short-term memory; while the retrieval module acts as long-term memory to access relevant examples.
\end{itemize}

\subsubsection{Action Module}
The action module equips the LM agent with the ability to execute and adapt actions in various environments \cite{xi2023rise,durante2024agent}. This module is designed to handle embodied actions and tool-use capabilities, allowing the agent to interact with its physical surroundings adaptively and effectively.
Besides, tools significantly broaden the agent's action space.
\begin{itemize}
    \item \textit{Embodied actions.} The action module empowers LM agents to perform contextually appropriate embodied actions and adapt to environmental changes, facilitating interaction with and adjustment to physical surroundings \cite{huang2022language,ichter2022do}.
    As LLM-generated action plans are often not directly executable in interactive environments, Huang \textit{et al.} \cite{huang2022language} propose refining LLM-generated plans for embodied agents by conditioning on demonstrations and semantically translating them into admissible actions. Evaluations in the VirtualHome environment show significant improvements in executability, ranging from 18\% to 79\% over the baseline LLM. Besides, SayCan \cite{ichter2022do} enables embodied agents such as robots to follow high-level instructions by leveraging LLM knowledge in physically grounded tasks, where LLM (i.e., Say) suggests useful actions; while learned affordance functions (i.e., Can) assess feasibility. SayCan's effectiveness is demonstrated through 101 zero-shot real-world robotic tasks in a kitchen setting.
    PaLM-E \cite{driess2023palme} is a versatile multimodal language model for embodied visual-language and language tasks. It integrates continuous sensor inputs, e.g., images and state estimates, into the same embedding space as language tokens, allowing for grounded inferences in real-world sequential decision-making.

    \item \textit{Learning to use \& make tools.} By leveraging various tools (e.g., search engines and external APIs) \cite{talebirad2023multi}, LM agents can gather valuable information to handle assigned complex tasks.
    For instance, AutoGPT integrates LLMs with predetermined tools such as web and file browsing.
    Beyond using existing tools, LM agents can also develop new tools to enhance task efficiency \cite{xi2023rise}.
    To optimize tool selection with a large toolset, ReInvoke \cite{chen2024Re-Invoke} introduces an unsupervised tool retrieval method featuring a query generator to enrich tool documents in offline indexing and an intent extractor to identify tool-related intents from user queries in online inference, followed by a multi-view similarity ranking strategy to identify the most relevant tools.
\end{itemize}

\begin{figure*}[!t]
\centering \setlength{\abovecaptionskip}{-0.cm}
  \includegraphics[width=0.8\textwidth]{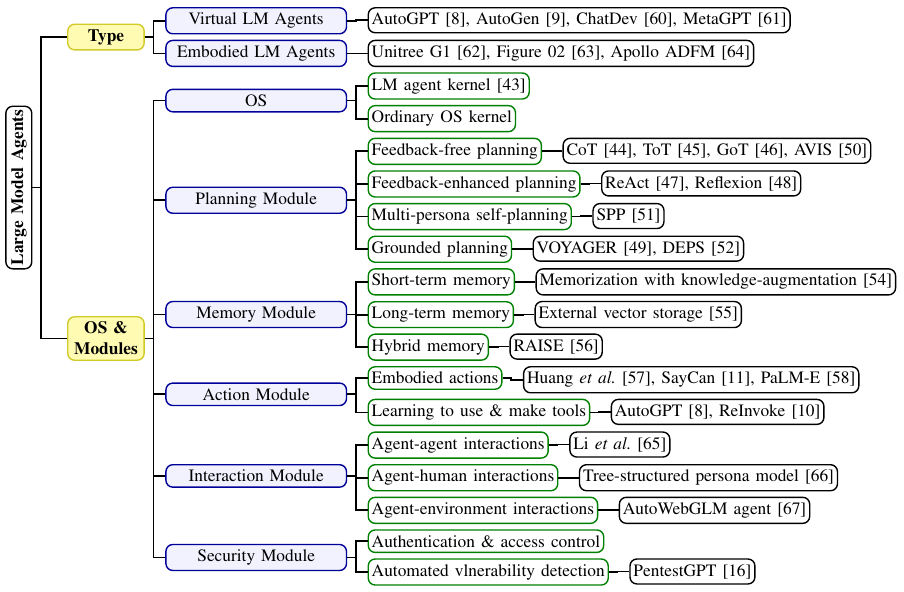}
	\caption{{The taxonomy of type and constructing modules of LM agents.}}
	\label{fig: taxnonmy of large model agents}
\end{figure*}

\subsubsection{Interaction Module}
The interaction module enables the LM agent to interact with humans, other agents, and the environment \cite{guo2024large}.
Through these varied interactions, LM agents can gather diverse experiences and knowledge, which are essential for comprehensive understanding and adaptation.
\begin{itemize}
    \item \textit{Agent-Agent Interactions.} LM agents can coordinate efforts on shared tasks, exchange knowledge to solve problems, and negotiate roles to collaborate with other agents, fostering a cooperative network \cite{talebirad2023multi}. Theory of mind \cite{li-2023-theory} can enhance the ability to understand other agents' hidden mental states. Li \textit{et al.} \cite{li-2023-theory} propose a prompt-engineering method to incorporate explicit belief state representations. They also introduce a novel evaluation of LLMs' high-order theory of mind in teamwork scenarios, emphasizing dynamic belief state evolution and intent communication among agents.
    \item \textit{Agent-Human Interactions.} LM agents can interact with humans by understanding natural languages, recognizing human emotions and expressions, and providing assistance in various tasks \cite{hu2024survey}. As found by work \cite{jinxin2023CGMI}, LLMs such as GPT-4 tend to forget character settings in multi-turn dialogues and struggle with detailed role assignments due to context window limits. To address this, a tree-structured persona model is devised in \cite{jinxin2023CGMI} for character assignment, detection, and maintenance, enhancing agent interactions.
    \item \textit{Agent-Environment Interactions.} LM agents can engage directly with the physical or virtual environments. By facilitating engagement in physical, virtual, or mixed-reality environments \cite{ichter2022do}, the interaction module ensures that LM agents can operate effectively across different contexts. Lai \textit{et al.}  develop the AutoWebGLM agent \cite{Lai2024AutoWebGLM}, which excels in web browsing missions via self-sampling reinforcement learning, curriculum learning, and rejection sampling fine-tuning. A Chrome extension based on AutoWebGLM validates its effective reasoning and operation capability across various websites in real-world services.
\end{itemize}

\subsubsection{Security Module}
The security module is crucial to ensure the secure, safe, ethical, and privacy-preserving operations of LM agents \cite{durante2024agent}. It is designed to monitor and regulate the LM agents' actions, interactions, and decisions to prevent harm and ensure compliance with legal and ethical standards. LMs can be utilized to design advanced and cost-effective defense mechanisms, such as PentestGPT \cite{deng2024pentestgpt} for automated network vulnerability detection.
LMs also introduce new threats to agents such as magnified hallucination and memorization risks. 
The security module should dynamically adapt to emerging threats by learning from new security/privacy incidents and integrating updates from security/privacy databases and policies. It also incorporates ethical guidelines and bias mitigation techniques to ensure fair and responsible behaviors.

\textit{Connections Between Modules:}
The above five key components of an LM agent are interconnected to create a cohesive and intelligent system. Particularly, the planning module relies on the memory module to access past experiences and external knowledge, ensuring informed decision-making. The action module executes plans generated by the planning module, adapting actions based on real-time feedback and memory. The interaction module enhances these processes by facilitating communication and collaboration, which provides additional data and context for the planning and memory modules. Besides, security considerations are seamlessly integrated into every aspect of the LM agent's operations to ensure robust and trusted executions. {Fig.~\ref{fig: taxnonmy of large model agents} shows the taxonomy of type and key modules of LM agents.}

\begin{table*}[!t]
   \centering
\caption{Comparison of Typical LM Agent Prototypes}
\begin{tabular}{|l|c|c|c|c|l|l|}
\hline
\textbf{Prototype} & \textbf{Application} & \textbf{\begin{tabular}[c]{@{}c@{}}Enhancing \\ Intelligence\end{tabular}} & \textbf{\begin{tabular}[c]{@{}c@{}}Improving \\ Safety\end{tabular}} & \textbf{\begin{tabular}[c]{@{}c@{}}Optimizing \\ Experience\end{tabular}} & \textbf{Other Key Features} & {\textbf{Limitations}}\\ \hline

AutoGPT{\cite{yang2023auto}} & {\begin{tabular}[c]{@{}l@{}}General \\ scenarios\end{tabular}} & Yes & N/A & Yes & {\begin{tabular}[c]{@{}l@{}}Task decomposition and\\ execution\end{tabular}}  &{{\begin{tabular}[c]{@{}l@{}}Ecosystem isolation, \\single-device simulation\end{tabular}}} \\ \hline

AutoGen{\cite{wu2024autogen}} & {\begin{tabular}[c]{@{}l@{}}General \\ scenarios\end{tabular}} & Yes & N/A & Yes & {\begin{tabular}[c]{@{}l@{}}Multi-agent conversation \\ framework\end{tabular}}  &{{\begin{tabular}[c]{@{}l@{}}Ecosystem isolation, \\single-device simulation\end{tabular}}} \\ \hline


ChatDev{\cite{qian2024chatdev}} & {\begin{tabular}[c]{@{}l@{}}General \\ scenarios\end{tabular}} & Yes & N/A & Yes & {\begin{tabular}[c]{@{}l@{}}Conversational AI for \\ software development\end{tabular}}  &{{\begin{tabular}[c]{@{}l@{}}Ecosystem isolation, \\single-device simulation\end{tabular}}} \\ \hline

MetaGPT{\cite{hong2024metagpt}} & {\begin{tabular}[c]{@{}l@{}}General \\ scenarios\end{tabular}} & Yes & N/A & Yes & {\begin{tabular}[c]{@{}l@{}}Meta-learning for task\\adaptation\end{tabular}}  &{{\begin{tabular}[c]{@{}l@{}}Ecosystem isolation, \\single-device simulation\end{tabular}}} \\ \hline

MobileAgent v2{\cite{wang2024mobileagentv2}} & {\begin{tabular}[c]{@{}l@{}}General \\ scenarios\end{tabular}} & Yes & N/A & Yes & {\begin{tabular}[c]{@{}l@{}}Multimodal input support\end{tabular}}  & {{\begin{tabular}[c]{@{}l@{}}Unstable performance \\ for languages\end{tabular}}} \\ \hline


NetLLM{\cite{10.1145/3651890.3672268}} & {\begin{tabular}[c]{@{}c@{}}Mobile \\ communications\end{tabular}}  & Yes & N/A & Yes & {\begin{tabular}[c]{@{}l@{}}Fine-tunes LLM for\\ networking scenarios\end{tabular}}  & {{\begin{tabular}[c]{@{}l@{}}Suboptimal performance \\ of multimodal fusion\end{tabular}}} \\ \hline

NetGPT{\cite{10466747}} & {\begin{tabular}[c]{@{}c@{}}Mobile \\ communications\end{tabular}} & Yes & N/A & Yes & {\begin{tabular}[c]{@{}l@{}}Cloud-edge cooperative \\ LM service\end{tabular}}  & {{\begin{tabular}[c]{@{}l@{}}Missing exploration of \\ multimodal LMs\end{tabular}}} \\ \hline

Figure 02{\cite{Figure02}} & Intelligent robots & Yes & Yes & N/A & Performs dangerous jobs & {{\begin{tabular}[c]{@{}l@{}}Simulation-to-reality gap\\in training\end{tabular}}}\\ \hline

Unitree G1{\cite{Unitree}} & Intelligent robots & Yes & Yes & N/A & {\begin{tabular}[c]{@{}l@{}}2nd generation humanoid\\robot\end{tabular}}  & {{\begin{tabular}[c]{@{}l@{}}Limited autonomy in\\ complex environments\end{tabular}}} \\ \hline

Apollo ADFM{\cite{ADFM}} & {\begin{tabular}[c]{@{}c@{}}Autonomous \\ driving\end{tabular}}  & Yes & Yes & Yes & {\begin{tabular}[c]{@{}l@{}}Supports L4 autonomous\\ driving\end{tabular}}  & {{\begin{tabular}[c]{@{}l@{}}Challenges in handling \\extreme edge cases\end{tabular}}} \\ \hline

PentestGPT{\cite{deng2024pentestgpt}} & {\begin{tabular}[c]{@{}c@{}}Attack-defense \\ confrontation\end{tabular}}  & Yes & Yes & N/A & {\begin{tabular}[c]{@{}l@{}}87\% success in \\vulnerability exploitation\end{tabular}}  & {{\begin{tabular}[c]{@{}l@{}}Limited effectiveness in\\high-complexity scenes\end{tabular}}}\\ \hline

AutoAttacker{\cite{xu2024autoattacker}} & {\begin{tabular}[c]{@{}c@{}}Attack-defense \\ confrontation\end{tabular}} & Yes & No & N/A & {\begin{tabular}[c]{@{}l@{}}Automatically execute \\ network attacks\end{tabular}}  & {{\begin{tabular}[c]{@{}l@{}}Narrow scope, \\LLM reliability risks\end{tabular}}}\\ \hline
\end{tabular}
\end{table*}

\subsection{Modern Prototypes \& Applications of LM Agents}\label{subsec:Applications}
\subsubsection{General Scenarios}
Recently, various industrial projects of LM agents, such as AutoGPT, AutoGen, BabyAGI, ChatDev, and MetaGPT, demonstrate their diverse potential in assisting web, life, and business scenarios, such as planning personalized travels, automating creative content generation, and enhancing software development workflows.
For instance, \textit{AutoGPT} \cite{yang2023auto} is an open-source autonomous agent utilizing GPT-3.5 or GPT-4 APIs to independently execute complex tasks by breaking down them into several sub-tasks and chaining LLM outputs, showcasing advanced reasoning capabilities.
\textit{AutoGen} \cite{wu2024autogen}, developed by Microsoft, offers an open-source multi-agent conversation framework, supports APIs as tools for improved LLM inference, and emphasizes the automatic generation and fine-tuning of AI models.
\textit{ChatDev} \cite{qian2024chatdev} focuses on enhancing conversational AI, providing sophisticated dialogue management, coding, debugging, and project management to streamline software development processes.
\textit{MetaGPT} \cite{hong2024metagpt} explores the meta-learning paradigm, where the model is trained to rapidly adapt to new tasks by leveraging knowledge from related tasks, thus improving performance across diverse applications. 


LM agents can also enhance the QoE of end users.
For instance, \textit{MobileAgent v2} \cite{wang2024mobileagentv2}, launched by Alibaba, is a mobile device operation assistant that achieves effective navigation through multi-agent collaboration, automatically performing tasks such as map navigation, and supports multimodal input including visual perception, enhancing operational efficiency on mobile devices. 
%
{Besides, DeepSeek-R1 \cite{guo2025deepseek} has been widely adopted across  industries, including finance, education, healthcare, office automation, and AI assistants. Leading companies such as Baidu, AWS, Azure AI, and Qihoo 360 have integrated DeepSeek-R1 into their applications. Additionally, AI agent platforms such as Coze AI and BetterYeah AI enable seamless agent deployment and customization using DeepSeek-R1. For instance, AgenticFlow \cite{AgenticFlow} automates marketing campaigns through AI-driven workflows, while Dify \cite{Dify} leverages DeepSeek-R1 to build assistants, workflows, and text generators. RAGFlow \cite{RAGFlow}, an open-source RAG engine, leverages DeepSeek-R1 to enhance document understanding and enterprise RAG workflows.}

\subsubsection{Mobile Communications}
LM agents offer significant advantages for mobile communications by enabling low-cost and context-aware decision-making \cite{10.1145/3651890.3672268} and personalized user experiences \cite{10466747}. 
For instance, \textit{NetLLM} \cite{10.1145/3651890.3672268} fine-tunes the LLM to acquire domain knowledge from multimodal data in networking scenarios (e.g., adaptive bitrate streaming, viewport prediction, and cluster job scheduling) with reduced handcraft costs. Meanwhile, \textit{NetGPT} \cite{10466747} uses a cloud-edge cooperative LM framework for personalized outputs and enhanced prompt responses in mobile communications via de-duplication and prompt enhancement technologies. \textit{ChatNet} \cite{10614634} uses four GPT-4 models to serve as analyzer (to plan network capacity and designate tools), planner (to decouple network tasks), calculator (to compute and optimize the cost), and executor (to produce customized network capacity
solutions) via prompt engineering.

\subsubsection{Intelligent Robots}
LM agents play a crucial role in advancing intelligent industrial and service robots \cite{ichter2022do}. These robots can perform complex tasks such as product assembly, environmental cleaning, and customer service, by perceiving surroundings and learning necessary skills through LMs.
In August 2024, FigureAI released \textit{Figure 02} \cite{Figure02}, a human-like robot powered by OpenAI LM, capable of fast common-sense visual reasoning and speech-to-speech conversation with humans to handle dangerous jobs in various environments. Besides, Unitree unveiled its humanoid robot agent named \textit{Unitree G1} \cite{Unitree} in August 2024, which demonstrates the enhanced capabilities brought by LM agents. 


\subsubsection{Autonomous Driving}
LM agents are transforming autonomous driving by enhancing vehicle intelligence, improving safety, and optimizing driving experience (e.g., offering personalized in-car experience) \cite{jin2023surrealdriver}.
In May 2024, Baidu launched Carrot Run's sixth-generation unmanned vehicle, built upon the \textit{Apollo autonomous driving foundation model (ADFM)} \cite{ADFM}, an LM agent supporting L4 autonomous driving.
Companies such as Tesla, Waymo, and Cruise are also integrating LM agents to enhance autonomous driving systems, aiming for safer and more efficient transportation.

\subsubsection{Autonomous attack-defense confrontation}
LM agents can be utilized as autonomous and intelligent cybersecurity decision-maker capable of making security decisions and taking threat handling actions without human intervention. For instance, \textit{PentestGPT} \cite{deng2024pentestgpt} is an automated penetration testing tool supported by LLMs, designed to use GPT-4 for automated network vulnerability scanning and exploitation. \textit{AutoAttacker} \cite{xu2024autoattacker}, an LM tool, can autonomously generate and execute network attacks based on predefined attack steps. As reported in \cite{fang2024llmagentsautonomous}, LM agents can automatically exploit one-day vulnerabilities; and in tests on 15 real-world vulnerability datasets, GPT-4 successfully exploited 87$\%$ of vulnerabilities, significantly outperforming other tools.

\begin{figure*}[!t]
\centering \setlength{\abovecaptionskip}{-0.cm}
  \includegraphics[width=19.2cm]{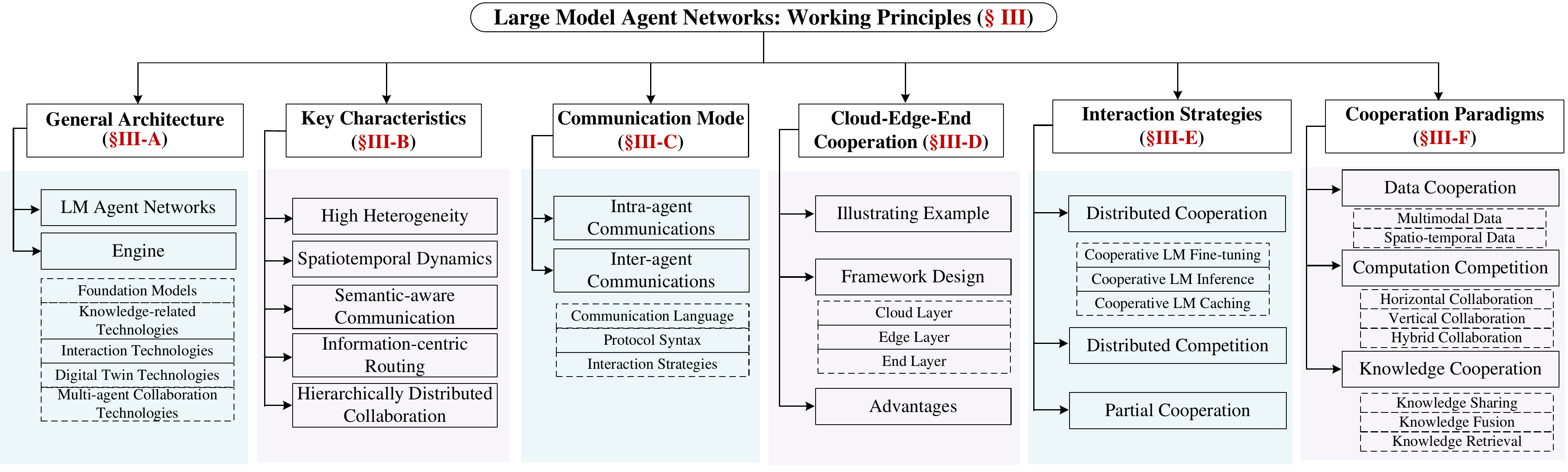}
  \caption{{Organization structure of the Section~\ref{sec:Fundamentals}.}}\label{fig:S3architecture}\vspace{-2mm}
\end{figure*}

\subsection{{Summary and Lessons Learned}}
{This section provides a comprehensive review of the design, functionality, and applications of LM agents, focusing on their architectural standards, operational frameworks, and potential use cases. We identify five core modules, i.e., planning, action, memory, interaction, and security, as essential components for enabling LM agents to operate effectively in dynamic and complex environments. LM agents are already making a transformative impact across industries such as education, mobile communications, robotics, autonomous driving, and cybersecurity. To summarize, the key lessons learned include:
\begin{itemize}
    \item Beyond general-purpose OSs (e.g., Windows and Linux), OS for embodied LM agents should cater to robotic environments such as UAVs, autonomous vehicles, and robots. Integrating systems such as ROS (Robot Operating System) for machine-centric communications will be key to ensure seamless functionality and promote embodied intelligence.
    \item LMs significantly enhance LM agents' abilities across core modules including planning, action, memory, interaction, and security, facilitating more intelligent, adaptive decision-making and improving overall performance.
    \item As key enablers of next-generation AI systems, LM agents will drive the future of intelligent systems. In everyday life (e.g., home automation, transportation, personal assistance), they will empower smarter, more efficient living. In the future network landscape, LM agents will be central to evolving network dynamics, with cybersecurity shifting to agent-versus-agent defense and attack strategies. As the popularity of LM agents, efficiently constructing LM agent networks to foster collaboration and achieve synergistic effects becomes increasingly crucial.
    \item {The capabilities of LM agents have expanded significantly, enabling them to perform a wider range of tasks with enhanced efficiency. However, most existing virtual agent prototypes operate within isolated ecosystems and simulate agents only on a single device. Besides, existing virtual/embodied agent prototypes struggle with cost-effectiveness, cross-domain adaptability, and scalability, hindering their broader adoption. Addressing these challenges requires further research and innovation to enhance real-world implementation.}
\end{itemize}}

\begin{figure*}[!t]
\centering 
  \includegraphics[width=12.88cm]{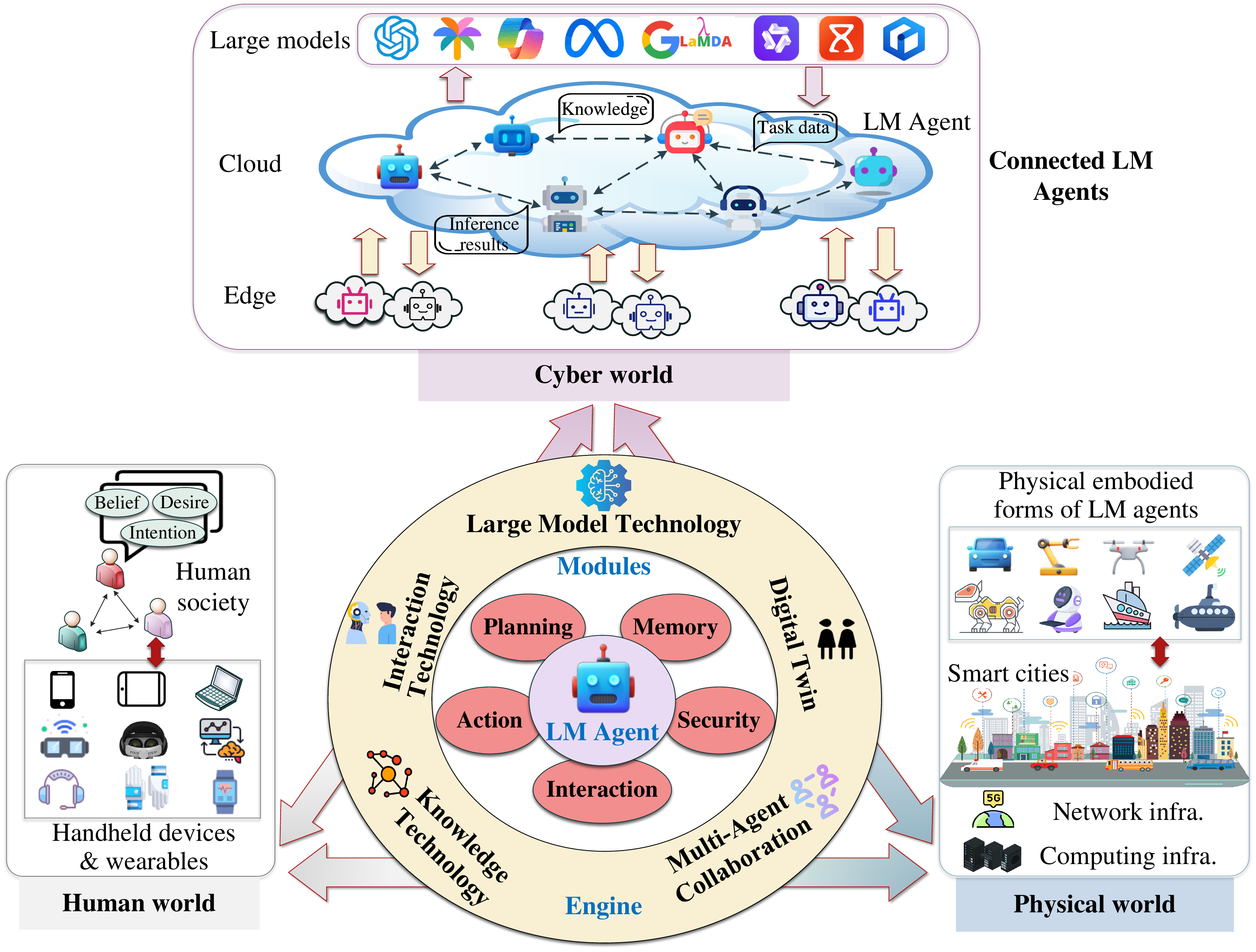}
  \caption{{The general architecture of LM agent networks in bridging human, physical, and cyber worlds. LM agent has five constructing \textit{modules}: planning, action, memory, interaction, and security modules. The \textit{engine} of connected LM agents is empowered by a combination of five cutting-edge technologies: foundation models, knowledge-related technologies, interaction, digital twin, and multi-agent collaboration.}}\label{fig:architecture}\vspace{-1mm} 
\end{figure*}

\begin{table*}[!t]
   \centering
    \caption{A Summary of LLM Stages, Utilized Technologies, and Representives}\label{LLMBasicTable1}
\begin{tabular}{|c|c|c|c|c|c|}
\hline
\textbf{Stage of LLM} &
  \textbf{Description} &
  \textbf{Utilized Technology} &
  \textbf{Representive} &
  \begin{tabular}[c]{@{}c@{}}\textbf{Required} \\ \textbf{Dataset Scale}\end{tabular} & \begin{tabular}[c]{@{}c@{}}\textbf{Required} \\ \textbf{Computation Power}\end{tabular}
   \\ \hline
\textbf{Pre-training} &
  \begin{tabular}[c]{@{}c@{}}Pre-train LLM in a self-supervised \\ manner on a large corpus\end{tabular} &
  Transformer &
  \begin{tabular}[c]{@{}c@{}}GPT-3, PaLM-2, \\ LLaMA-2\end{tabular} &
  \begin{tabular}[c]{@{}c@{}}Internet-scale (e.g., \\2T tokens in LLaMA-2)\end{tabular} &
  \begin{tabular}[c]{@{}c@{}}Very large (e.g., Thousands of\\GPUs for months in LLaMA-2)\end{tabular} \\ \hline
\textbf{Fine-tuning} &
  \begin{tabular}[c]{@{}c@{}}Fine-tuning pretrained \\ LLM for downstream tasks\end{tabular} &
  \begin{tabular}[c]{@{}c@{}}Instruction-tuning,\\ Alignment-tuning, \\ Transfer learning\end{tabular} &
  \begin{tabular}[c]{@{}c@{}}WebGPT\cite{nakano2022webgpt}, T0, \\ LLaMA-2-Chat\end{tabular} &
  \begin{tabular}[c]{@{}c@{}}Domain-scale (e.g., \\10K$\sim$100K QA pairs\\ in LLaMA-2)\end{tabular} &
  \begin{tabular}[c]{@{}c@{}}Medium (e.g., Dozens of\\GPUs for days in LLaMA-2)\end{tabular}\\ \hline
\textbf{Prompting} &
  \begin{tabular}[c]{@{}c@{}}Setup prompts and query trained \\ LLMs for generating responses\end{tabular} &
  \begin{tabular}[c]{@{}c@{}}Prompt engineering,\\ Zero-shot prompting, \\ In-context learning\end{tabular} &
  \begin{tabular}[c]{@{}c@{}}RALM\cite{ram2023incontextRALM}, \\ ToolkenGPT\cite{hao2023toolkengpt}\end{tabular} &
  \begin{tabular}[c]{@{}c@{}}Small-scale\end{tabular} &
  \begin{tabular}[c]{@{}c@{}}Low\end{tabular}\\ \hline
\end{tabular}
\end{table*}

\section{Large Model Agent Networks: Working Principles}\label{sec:Fundamentals}
In this section, we first explore the general architecture of LM agent networks, covering their definition and underlying engine. We then discuss the key characteristics and communication modes of LM agent networks. Next, we introduce a cloud-edge-end cooperative framework for networking LM agents, followed by an in-depth analysis of interaction strategies, including cooperation, partial cooperation, and competition. Finally, we examine cooperation paradigms in LM agent networks, focusing on data, computation, and knowledge sharing. {Fig.~\ref{fig:S3architecture} shows the organization structure of this section.}

\subsection{General Architecture of LM Agent Networks}\label{subsec:Architecture}
{As shown in Fig.~\ref{fig:architecture}, LM agent networks bridge the human, physical, and cyber worlds. Specifically, LM agents interact with humans through HMI technologies such as NLP, with the assistance of handheld devices and wearables, to understand human intentions, desires, and beliefs. LM agents can synchronize data and statuses between the physical body and the digital brain through digital twin technologies, as well as perceive and act upon the surrounding virtual/real environment. LM agents can be interconnected in the cyber space through efficient cloud-edge networking for efficient sharing of data, knowledge, and computation results to promote multi-agent collaboration.} 

\subsubsection{LM Agent Networks}
LM agent networks are decentralized networks of connected LM agents (either in embodied or software forms) that collaborate to accomplish complex tasks by seamlessly sharing data, knowledge, and resources. Each agent, equipped with specialized capabilities, is assigned specific sub-tasks, and collectively, these agents combine their outputs to provide comprehensive solutions. In the context of cloud-edge computing, LM agent networks optimize resource utilization and task distribution by deploying smaller, specialized models on edge devices. This ensures low-latency and scalable inference services across a range of applications, while leveraging the cloud for more resource-intensive operations. These agents can implement various collaboration strategies, such as horizontal, vertical, or hybrid models (as detailed in Sect.~\ref{subsec:CoopParadigm}), to enhance task execution efficiency and system performance.


\subsubsection{Engine of LM Agent Networks}
The LM agent networks are powered by a combination of cutting-edge technologies including foundation models, knowledge-related technologies, interaction, digital twin, and multi-agent collaboration.
\begin{itemize}
    \item \textit{Foundation models}, such as LLMs, large vision models (LVMs), and VLMs, serve as the brains of LM agents that comprehensively power them in planning, action, memory, interaction, and security capacities, as detailed in Sect.~\ref{ConstructingModules}. Table~\ref{LLMBasicTable1} summarizes the basic stages of LLM services.
\textit{(i) Advanced reasoning:} LM technologies empower AI agents with advanced reasoning abilities such as multi-persona self-planning and grounded planning. For instance, CoT \cite{wei2023cot} and ToT \cite{yao2023tot} reasoning allow LM agents to break down complex tasks into manageable sub-tasks.
\textit{(ii) Few/zero-shot generalization:}
Pretrained on extensive corpora, LMs demonstrate few-shot and zero-shot generalization capabilities \cite{huang2022language}, allowing LM agents to transfer knowledge seamlessly between tasks.
\textit{(iii) Tool usage capability:} Through the use of various tools, LM agents can gather and integrate valuable information from diverse sources to execute and adapt actions under complex scenarios.
\textit{(iv) Adaptability:} Through continuous learning and adaptation, LM agents can accumulate knowledge over time by learning from new data and experiences \cite{wang2023self}.

\item \textit{Knowledge-related technologies} enhance LM agents by incorporating both \textit{internal knowledge} (arisen from agent's interactions with humans, environment, and other agents) and \textit{external knowledge sources} (e.g., knowledge graph (KG) and vector database \cite{kuroki2024multiagentbehavior}) to produce up-to-date and contextually relevant outputs.
\textit{(i) Knowledge sharing:} To accomplish a common task, the locally ongoing updated private knowledge or experience of LM agents should be synchronized across all collaborators, enabling them to incrementally incorporate new information and adapt to evolving environments.
\textit{(ii) Knowledge fusion:} It enables LM agents to build comprehensive knowledge bases by integrating knowledge from diverse sources \cite{wan2024knowledge}, such as structured databases, vector databases, and KGs.
\textit{(iii) Knowledge retrieval:} RAG combines retrieval mechanisms with generative models, enabling LM agents to dynamically fetch relevant external information from vast knowledge sources \cite{dai2024vistarag}, which ensures the outputs of LM agents are reliable and up-to-date.

\item \textit{Interaction technologies} enhance LM agents' ability to engage naturally, immersively, and contextually with users \cite{durante2024agent}.
\textit{(i) HMI:} HMI technologies \cite{ichter2022do} allow LM agents to understand complex instructions, recognize speech, and interpret emotions, facilitating intuitive interactions.
Besides, LM-empowered multimodal interfaces enable responses to various inputs (e.g., text, speech, and gestures), making interactions flexible and user-friendly.
\textit{(ii) 3D digital humans:} 3D digital humans provide realistic interfaces for empathetic and personable interactions in applications such as customer support and healthcare \cite{10398474}.
\textit{(iii) AR/VR/MR:} AR, VR, and MR technologies create immersive and interactive environments, allowing users to engage with digital and physical elements seamlessly.

\item \textit{Digital twin technologies} allow efficient and seamless synchronization of data/statuses between the physical body and the digital brain of an LM agent \cite{10090432}.
\textit{(i) Virtual-physical synchronization:} LM agents can seamlessly synchronize attributes, behaviors, states, and other data between their bodies and brains via intra-agent bidirectional communications. The virtual representations of LM agents' bodies can be continuously updated with real-time data inputs \cite{10090432}.
\textit{(ii) Virtual-physical feedback:} This continuous feedback loop enhances LM agent's contextual awareness, allowing for immediate adjustments to changing conditions.
\textit{(iii) Predictive analytics:} Digital twins facilitate predictive analytics and simulation, allowing LM agents to anticipate future states and optimize actions accordingly \cite{jin2023surrealdriver}.

\item \textit{Multi-agent collaboration technologies} enable coordinated efforts of multiple LM agents to achieve common goals and tackle complex tasks \cite{guo2024large}.
Typical technologies include multi-agent reinforcement learning (MARL) \cite{esmaeil_seraj_2023}, cooperative game \cite{5740907}, mean-field game, Nash bargaining, and swarm intelligence algorithms.
Specifically, collaborative problem-solving techniques, such as multi-agent planning \cite{ZhangProAgent2024} and distributed reasoning \cite{10634552}, enable LM agents to jointly analyze complex issues, devise solutions, and coordinate actions. Besides, efficient synchronization of knowledge allows LM agents to build upon each other's experiences, accelerating learning in accomplishing the long-term collective task\cite{wang2023self}.
\end{itemize}

\begin{table*}[!t]
   \centering \setlength{\abovecaptionskip}{0cm}
    \caption{A Summary of Intra-Agent and Inter-Agent Communications for Connected LM Agents}\label{CommTable}
\begin{tabular}{|l|c|c|}
\hline
                    & \textbf{Intra-agent Comm.} & \textbf{Inter-agent Comm.} \\ \hline
{Involved Entity} & \begin{tabular}[c]{@{}c@{}}Brain$\longleftrightarrow$Body;\\Among planning, action, memory, \\interaction, and security modules\end{tabular} & {Brain$\longleftrightarrow$Brain} \\ \hline
Connection Type & Within a single LM agent & Among multiple LM agents \\ \hline
{Support Two-way Communication} & \ding{52} & \ding{52} \\ \hline
Support Multimodal Interaction & \ding{52} & \ding{52} \\ \hline
{Support Semantic-aware Communication} & \ding{52} & \ding{52} \\ \hline
Typical Communication Environment & Wired/Wireless & Wired/Wireless \\ \hline
\end{tabular}
\end{table*}


\subsection{Key Characteristics of LM Agent Networks}
LM agent networks exhibit the following distinct features that enable scalable, intelligent, and adaptive multi-agent collaboration across diverse applications.

\textit{1) High Heterogeneity.} LM agent networks generally manage a vast and diverse array of nodes to support a wide range of tasks and services. Such heterogeneity also entails inherent interoperability challenges. The heterogeneity spans:
\begin{itemize}
    \item \textit{Capability variance:} LM agents exhibit significant diversity in computational power, memory, and GPU capacities, requiring the network to effectively accommodate agents with varying resource constraints.
    \item \textit{Service diversity:} The LM agent network should adapt to diverse service scenarios with varying numbers of participants to ensure scalability.
    \item \textit{Communication heterogeneity:} LM agents generally employ diverse communication protocols and interfaces, such as cellular, satellite, and WiFi.
\end{itemize}

\textit{2) Spatiotemporal Dynamics.} LM agent networks evolve dynamically in both temporal and spatial dimensions \cite{Yang_2023_ICCV}. These dynamic properties require the system's resilience and flexibility in response to real-time changes.
\begin{itemize}
    \item \textit{Temporal dynamics:} Agents' decision-making and interactions follow a temporal sequence, adapting to the progression of tasks or changes in the operational environment.
    \item \textit{Spatial dynamics:} The network topology of agents evolves due to mobility or environmental factors.
\end{itemize}

\textit{3) Semantic-aware Communication.} {Unlike traditional communication on the Internet prioritizing reliable data transmission, agent-based communication is fundamentally task-oriented, which exhibits three unique features \cite{10798108}:
\begin{itemize}
    \item \textit{Computing-oriented communications:} Agent interactions prioritize transmitting information that directly supports computational workflows of tasks. Agent communications achieve this by first interpreting data semantics (e.g., via knowledge base alignment) and then transmitting only task-critical abstractions. 
    \item \textit{Persistent communications:} Agent collaborations often span extended durations, requiring continuous adaptation to evolving task states. Agent communications leverage persistent memory to maintain shared context (e.g., incremental knowledge updates), enabling agents to reference prior interactions efficiently. 
    \item \textit{Memory-based communications:} Agent tasks frequently exhibit sequential dependencies (e.g., iterative problem-solving). Agent communications exploit memory to compress information hierarchically. For instance, transmitting only the changes relative to previously shared knowledge.
\end{itemize}}

Semantic-aware intra-agent and inter-agent communications facilitate a richer contextual understanding of the environment with improved accuracy of predictions and actions. 
\begin{itemize}
    \item \textit{Intra-agent semantic-aware communication.} Intra-agent communication focuses on brain-body synchronization of LM agents. These communications are semantic-aware, meaning only contextually relevant data are transmitted to minimize bandwidth usage. For instance, an LM-powered household robot sends its brain in the cloud only the changes in the environment and its position, rather than transmitting the entire environmental data repeatedly.
    \item \textit{Inter-agent semantic-aware communication.} Inter-agent communication between LM agents is inherently semantic-aware, with agents exchanging information based on relevance to their tasks or objectives. For instance, LM agents communicate via natural language for effective task coordination, negotiation, role allocation, and sequencing during task execution, and reach consensus on the final outcome.
\end{itemize}

    \textit{4) Information-centric Routing.} LM agent networks prioritize quickly retrieving relevant information over specific data sources. Unlike IP-based, host-oriented networks, information-centric routing focuses on ``what'' data is needed, enabling flexible and scalable communication through paradigms such as named data networking (NDN) and publish/subscribe (pub/sub).
    \begin{itemize}
        \item In NDN, agents request data by content name via interest messages, retrieving it from the nearest cache to reduce latency and network traffic, with in-network caching further enhancing efficiency by storing frequently requested data locally.
        \item The pub/sub model routes requests through distributed hash tables (DHTs), allowing publishers to publish content to multiple topics while subscribers continuously receive updates from their subscribed topics. For instance, MetaGPT \cite{hong2024metagpt} incorporates pub/sub mechanisms in multi-agent communication, enabling LLM agents to exchange messages seamlessly via a shared pool, where they publish outputs and access others’ content transparently.
    \end{itemize}

\textit{5) Hierarchically Distributed Collaboration.} LM agent networks employ a hierarchical yet collaborative decision-making framework across cloud, edge, and end layers. This distributed framework enables scalable and effective decision-making, particularly in resource-constrained or latency-sensitive scenarios.
\begin{itemize}
    \item \textit{Intra-layer collaboration:} Agents within each layer (cloud, edge, or end) autonomously optimize local decisions and resource allocations.
    \item \textit{Cross-layer coordination:} The network seamlessly integrates cloud-scale processing, edge-level real-time responsiveness, and end-level contextual awareness to support distributed collaboration across different layers.
\end{itemize}

\subsection{Communication Mode of LM Agents}
Every LM agent consists of two parts: (i) the LM-empowered cyber \textit{brain} located in the cloud, edge servers, or end devices and (ii) the corresponding physical or software-form \textit{body}. Every LM agent can actively interact with other LM agents, the virtual/real environment, and humans. For connected LM agents, there exist two typical communication modes: \textit{intra-agent communications} for seamless data/knowledge synchronization between brain and physical body within an LM agent, and \textit{inter-agent communications} for efficient coordination between LM agents. Table~\ref{CommTable} summarizes the comparison of the two communication modes.

\textit{1) Intra-agent communications} refer to the internal data/knowledge exchange within a single LM agent. This type of communication ensures that different components of the LM agent, including planning, action, memory, interaction, and security modules, work in harmony. For instance, an LM agent collects multimodal sensory data through its physical body, which then communicates the interpreted information to the LM-empowered brain. The planning module in the brain formulates a response or action plan, which is then executed by the action module. This seamless flow of information is critical for maintaining the LM agent's functionality, coherence, and responsiveness in real-time and dynamic scenarios. These communications are semantic-aware, transmitting only contextually relevant data, such as changes in the environment.

{For LM agents with physical embodiments, such as robots, ROS often serves as the foundational framework. ROS operates using a pub/sub mechanism, where components such as action module (e.g., sensors and actuators), planning module, and memory module either publish data streams or subscribe to them as needed. This structure enables flexible pub/sub interactions among various components, enabling LM agents to adapt quickly to dynamic environments.
In summary, intra-agent communications will be \textit{semantic-aware} and built on a \textit{pub/sub} framework to align the LM agent’s internal processes.} 

\begin{figure*}[!tp]
\centering 
  \includegraphics[width=\linewidth]{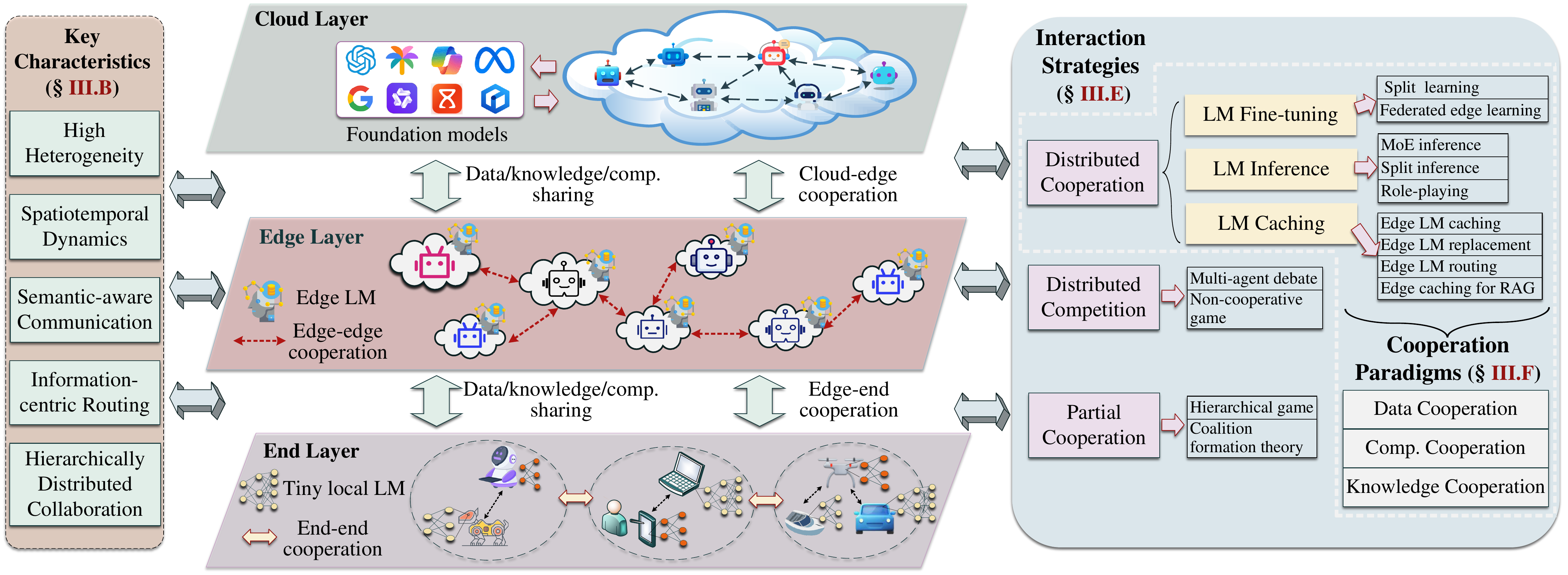}
  \caption{{Illustration of the cloud-edge-end cooperative network architecture of LM agents.}}\label{fig:cloud_edge-end_coop}\vspace{-3mm}
\end{figure*}

\textit{2) Inter-agent communications} involve information and knowledge exchange between multiple LM agents. It enables collaborative task allocation, resource sharing, and coordinated actions among agents to foster collective intelligence. For instance, in a smart city, various LM agents managing traffic lights, public transportation, and emergency services share real-time data to optimize urban mobility and safety. {Effective inter-agent communications rely on standardized protocols to ensure compatibility and interoperability, facilitating efficient and synchronized operations across the LM agent networks.
Two prerequisites should be met \cite{marro2024scalable}: (1) a shared conceptual understanding of the subject matter being communicated, and (2) mutually intelligible encoding and decoding for accurately interpreting messages by both agents. Hence, agent-to-agent communications are also \textit{semantic-aware}.
Inter-agent communication protocols should at least specify the following key components:
\begin{itemize}
    \item \textit{Communication language}, e.g., FIPA agent communications language (ACL) \cite{poslad2007specifying} and knowledge query and manipulation language (KQML) \cite{KQMLAgentComm}, defines the vocabulary and expressions used by agents to convey intentions.
    \item \textit{Protocol syntax} specifies the structure and format of messages, detailing essential fields like sender, receiver, content, and conversation ID.
    \item \textit{Interaction strategies} govern the sequences of communication, defining roles, turn-taking rules, and strategies for engaging in tasks. For instance, negotiation strategies dictate when agents can propose, counter, or conclude offers.
\end{itemize}}

{The design of inter-agent communication protocols faces the versatility-efficiency-portability trilemma: \cite{marro2024scalable}:
\begin{itemize}
    \item \textit{Versatility}: support diverse messages and applications to exchange information and coordinate actions. 
    \item \textit{Portability}: operate across heterogeneous underlying platforms and support seamless integration of new agents and functionalities without significant redesign.
    \item \textit{Efficiency}: optimize resource usage and minimize communication and computational overheads.
\end{itemize}}

{For instance, traditional static APIs are highly efficient and portable but lack versatility, while natural language communication offers high versatility and portability but often sacrifices efficiency. LM agents bridge this gap by understanding, manipulating, and responding to other agents through natural language while integrating external tools, writing code, and invoking APIs. LM agents excel at following instructions, including implementing routines through code. Furthermore, they autonomously negotiate protocols and reach consensus on strategies and behaviors in complex scenarios. Marro \textit{et al.} propose the Agora protocol \cite{marro2024scalable} to address the communication trilemma by employing standardized routines for frequent communications, natural language for infrequent communications (e.g., negotiation and error handling), and LLM-written routines for scenarios that fall between them.}

{In addition to Agora, other agent communication protocols have been developed, e.g., Model Context Protocol (MCP) \cite{MCP} and Agent Network Protocol (ANP) \cite{ANP}. 
Selecting or designing an appropriate agent communication protocol requires careful consideration of task scenarios (e.g., wired backhaul or radio access networks) and task-specific requirements (e.g., robustness, latency, and cost constraints). Additionally, adapting these protocols to diverse systems and agent platforms presents significant challenges, highlighting the need for flexible and interoperable solutions.}


\subsection{Cloud-Edge-End Cooperative Framework for LM Agent Networking}\label{subsec:CloudEdgeEnd}

\subsubsection{{Illustrating Example in Autonomous Driving}}
{Autonomous vehicles generate vast amounts of real-time data from sensors such as cameras and LiDAR, where LMs aid in making instant decisions on navigation, obstacle avoidance, and route optimization. For instance, LLMs can help determine the safest and most efficient route at intersections. However, autonomous driving demands low-latency responses, with 3GPP specifying end-to-end latency as low as 10 ms \cite{3gpp.22.874}, which makes cloud-based LM processing unsuitable due to high latency. Furthermore, centralizing sensor data (up to 4TB/day \cite{qu2024mobileedgeintelligencelarge}) in the cloud can overwhelm networks and compromise privacy, as sensitive information (e.g., location) may be exposed.}

{The cloud-edge-end collaborative framework addresses these issues by offloading LM processing to the edge, reducing bandwidth usage and ensuring faster response times. Edge or on-vehicle LM processing further protects privacy by avoiding data uploads to the cloud. This collaboration enhances autonomous driving through reduced latency, lower bandwidth costs, and improved privacy.}

\begin{figure*}[!t]
\centering \setlength{\abovecaptionskip}{-0.cm}
  \includegraphics[width=0.8\textwidth]{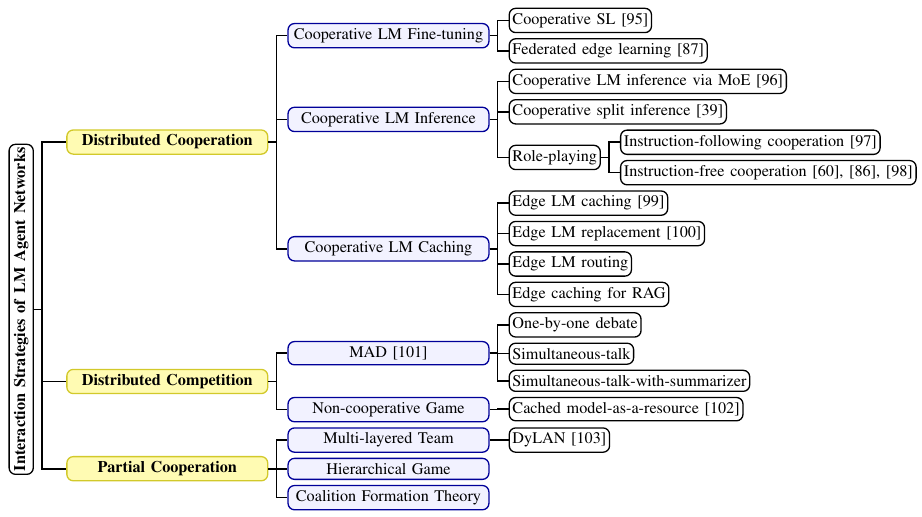}
	\caption{{The taxonomy of interaction strategies of LM agent networks.}}
	\label{fig: taxnonmy of Interaction Strategies of LM agent Networks}
\end{figure*}

\subsubsection{Cloud-Edge-End Cooperative Framework Design}
As depicted in Fig.~\ref{fig:cloud_edge-end_coop}, a cloud-edge-end cooperative framework is essential to resolve the above key challenges of LM agent networks. Generally, it includes three functional layers: cloud, edge, and end layers.
\begin{itemize}
    \item The \textit{cloud} serves as the centralized intelligence hub, hosting full-scale LMs (e.g., exceeding 100B parameters) with extensive storage and computational resources. It performs computationally intensive tasks, such as large-scale model training, global decision-making, and cross-layer resource orchestration \cite{Wang2023AChatGPT}. The cloud also maintains a global knowledge repository, which supports dynamic updates and synchronization across the network.

    \item \textit{Edge nodes} provide context-aware, intermediate intelligence by hosting moderately sized LMs (e.g., 10B–50B parameters) \cite{10648594}. They bridge the gap between the cloud and end devices, performing task-oriented LM fine-tuning, proximal data aggregation, and contextual inference using localized datasets. They enable low-latency responses for region-specific tasks and support adaptive communications with end devices.

    \item \textit{End devices} (e.g., mobile agents and IoT devices) execute lightweight, real-time tasks using tiny local LMs (e.g., 0–10B parameters). They provide contextual inputs, collect real-time sensory data, and act as the first responders in scenarios requiring immediate action. {For instance, Google has launched the Gemini Nano\footnote{https://store.google.com/ideas/gemini-ai-assistant/} on Pixel 8 Pro smartphones with 1.8 billion and 3.25 billion parameters, which support relatively basic functions such as grammar checking and text summarization.}
    End devices also prioritize privacy preservation by processing sensitive data locally before sharing insights with edge or cloud layers.
\end{itemize}

For instance, Xu \emph{et al.} \cite{10648594} propose an edge computing framework tailored for LLM agents. In their framework, mobile agents with local models (0-10B parameters) handle real-time tasks, and edge agents with larger models (over 10B parameters) process broader contextual data to support complex decision-making. They also study a real use case of vehicle accidents, where mobile agents create localized accident scene descriptions, which are then enhanced by edge agents to generate comprehensive accident reports and actionable plans.

LM lightweighting techniques, e.g., knowledge distillation \cite{zhang2024vpgtrans}, quantization \cite{zhu2024surveymodelcompress}, pruning \cite{ma2023llm}, and hardware acceleration \cite{shen2024agile}, can be employed to deploy compact LM variants tailored to edge/end computing capacities. For instance, VPGTrans \cite{zhang2024vpgtrans} addresses the high computational costs of training visual prompt generators for multimodal LLMs by significantly reducing GPU hours and training data needed via transferring an existing visual prompt generator from one multimodal LLM to another. LLM-Pruner \cite{ma2023llm} compresses LLMs without compromising their multi-task solving and language generation abilities, using a task-agnostic structural pruning method that significantly reduces the need for extensive retraining data and computational resources.
Hardware accelerators, e.g., GPUs and TPUs, enhance computational efficiency and provide robust support for cloud training and edge deployment. Agile-Quant \cite{shen2024agile} enhances the efficiency of LLMs on edge devices via activation-guided quantization and hardware acceleration by utilizing a SIMD-based 4-bit multiplier and efficient TRIP matrix multiplication, achieving up to 2.55x speedup while maintaining high task performance.

\subsubsection{Advantages}
By assigning specialized roles to cloud, edge, and end devices, the cloud-edge-end architecture enhances flexibility to accommodate varying task complexity and resource availability. For instance, compute-intensive LM tasks can be offloaded to cloud servers, less demanding LM tasks can be handled by edge servers to reduce latency and bandwidth usage, while end devices can offload latency-sensitive LM operations to edge servers. Additionally, hierarchical collaborative learning is essential for LM agents to acquire global knowledge across large geographical areas, with LM updates of edge/end agents synchronized centrally in the cloud.
This hierarchical collaboration across cloud, edge, and end devices offers several key advantages:
\begin{itemize}
    \item \textit{Scalability:} By distributing computational workloads across layers, the architecture supports large-scale networks with heterogeneous agents.
    \item \textit{Efficiency:} The cloud handles high-complexity tasks, while edge and end devices manage localized, latency-sensitive operations, optimizing overall system performance.
    \item \textit{Adaptability:} Dynamic coordination across layers ensures that the system can respond flexibly to changes in spatiotemporal dynamics, such as agent mobility and evolving tasks.
    \item \textit{Security and Privacy:} Localized processing at the end layer reduces the exposure of sensitive data, while the cloud and edge nodes provide robust security mechanisms for cross-layer interactions.
\end{itemize}

\subsection{Interaction Strategies of Connected LM Agents}\label{subsec:InteractStrateAGENT}
The multi-agent interactions can be categorized into distributed cooperation, distributed competition, and partial cooperation  \cite{esmaeil_seraj_2023,talebirad2023multi}, each of which involves different strategies to jointly optimize the collective or individual outcomes. {Fig.~\ref{fig: taxnonmy of Interaction Strategies of LM agent Networks} illustrates the taxonomy of interaction strategies of LM agent networks.}


\subsubsection{Distributed Cooperation}
For connected LM agents, cooperation is a fundamental interaction mode where distributed LM agents work together to achieve common goals, share resources, and optimize collective outcomes. It includes three aspects: a) distributed cooperative LM fine-tuning, b) distributed cooperative LM inference, and c) distributed cooperative LM caching.

\begin{figure}[!tb]
\centering 
\setlength{\abovecaptionskip}{-0.cm}
  \includegraphics[width=0.94\linewidth]{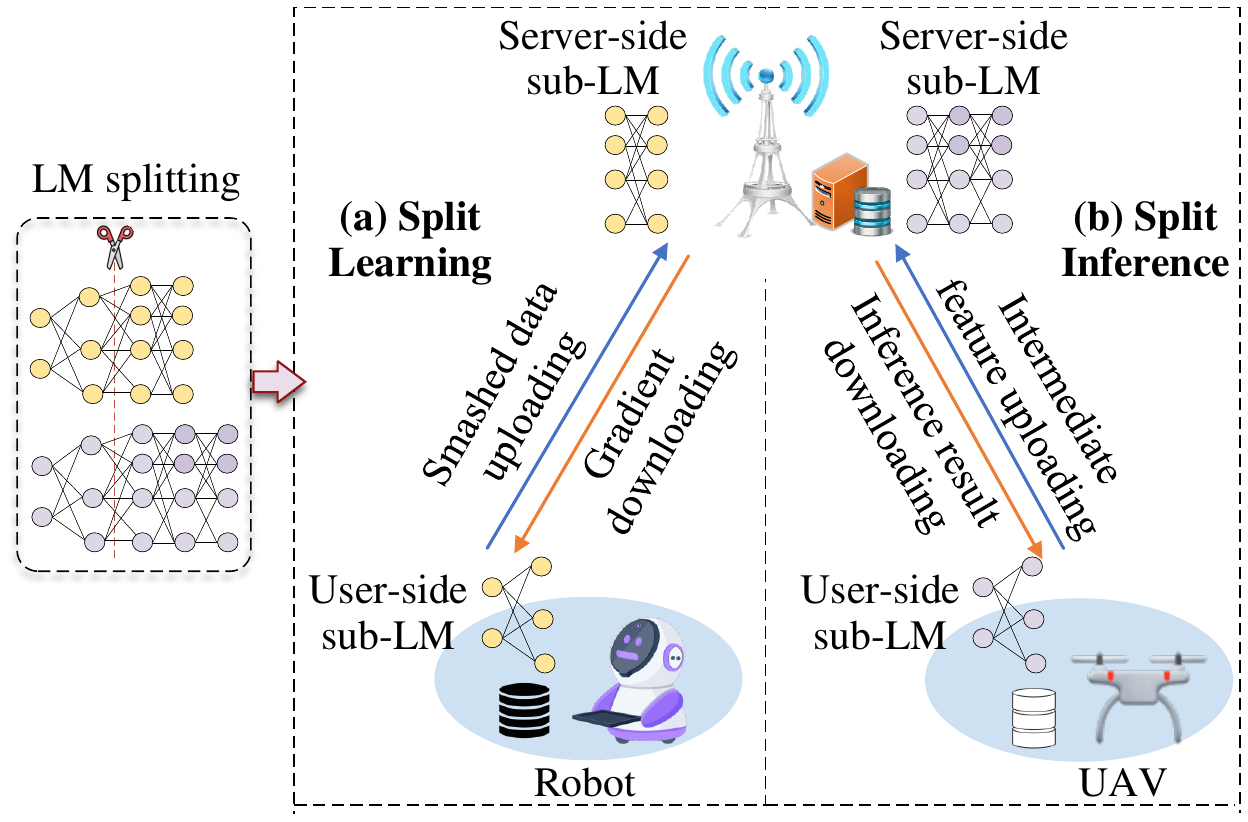}
  \caption{Illustration of (a) split learning for collaborative LM fine-tuning and (b) split inference for collaborative LM inference in edge environments.}\label{fig:splitLM}\vspace{-2mm}
\end{figure}

\textit{a) Distributed Cooperative LM Fine-tuning for LM Agents.} LMs, typically designed for general tasks, require fine-tuning with vast domain-specific data on edges to provide agent services in specialized fields. However, due to their massive size, fine-tuning LMs demands substantial computational power and energy, as shown in Table \ref{LLMBasicTable1}. To fine-tune LMs on heterogeneous, resource-constrained edge environments while ensuring privacy protection, we introduce two paradigms: collaborative split learning (SL) and federated edge learning, as below.

\textit{{\ding{172}} Cooperative SL.} SL distributes partitioned LMs on edge devices and servers. In SL, as shown in Fig.~\ref{fig:splitLM}(a), edge devices upload data features to edge servers for co-inference, reducing local computation while retaining major processing tasks at the edge. By placing only a submodel on edge devices, SL minimizes their computational burden and can integrate with parameter-efficient fine-tuning (PEFT) techniques \cite{han2024parameter} to further reduce workload.
However, deploying SL for LM agents in edge-end environments introduces several challenges. 
\begin{itemize}
\item Communication costs for transmitting high-dimensional data can be significant. For instance, partitioning LMs generates smashed data that should be uploaded, resulting in substantial overhead. In the case of GPT-3 Medium and 100 samples (1024 tokens each), this can reach approximately 400 MB per training round \cite{NEURIPS20201457c0d6}. 

\item Privacy concerns arise due to target token leakage. In common edge SL configurations, the input module is placed on the edge and the output module on the server, requiring edge devices to upload target tokens for training \cite{thapa2021advancements}. This can expose sensitive user data and compromise privacy. U-shaped SL \cite{robinson2023pragmatic} retains the initial and final neural layers or transformer blocks on edge devices, placing only intermediate layers on edge servers, thereby safeguarding label privacy to a large extent.
\end{itemize}

\textit{{\ding{173}} Federated edge learning.} In cloud-edge collaborative LM training, LMs are initially pre-trained in the cloud, {leveraging} extensive computational resources and large-scale datasets. Federated fine-tuning technology is then performed on edge devices using domain-specific data from various edge servers, enabling LM agent services tailored {to localized} downstream tasks {while optimizing} resource utilization across the network.
However, due to the enormous size of LMs (often containing billions of parameters), transmitting model updates from numerous edge servers for aggregation during federated fine-tuning imposes significant communication overheads and operational costs. Additionally, periodic local fine-tuning at the edge consumes substantial computational resources, {often exceeding} the capacity of resource-constrained edges. 

\begin{figure}[!tb]
\centering \setlength{\abovecaptionskip}{-0.cm}
  \includegraphics[width=0.94\linewidth]{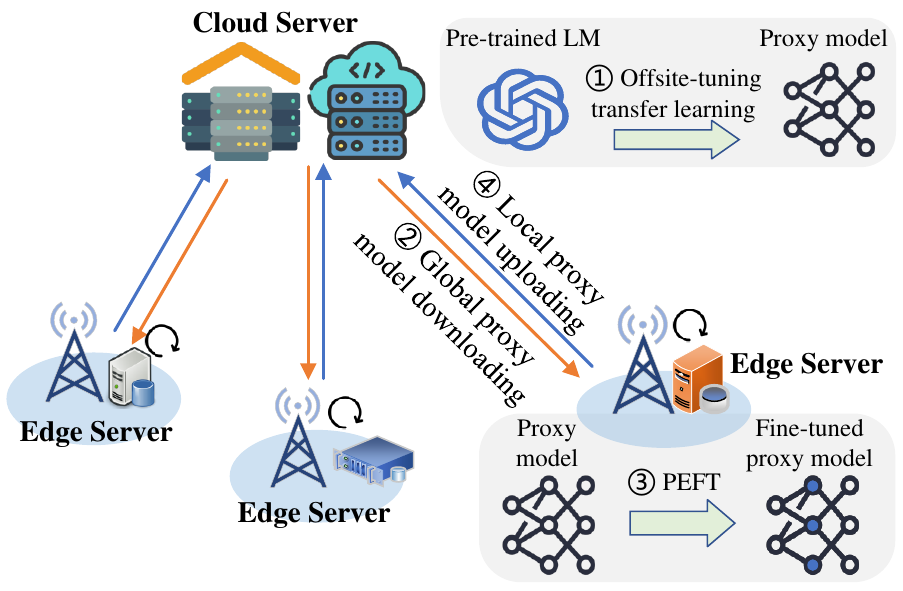}
  \caption{{Illustration of federated edge learning in cloud-edge environments for collaborative LM fine-tuning.}}\label{fig:federatededgelearning}\vspace{-2mm}
\end{figure}

The offsite-tuning transfer learning technique, in conjunction with PEFT techniques \cite{han2024parameter}, such as Prefix-tuning and LoRA, enables edge devices to update and transmit only a subset of model parameters, significantly reducing both communication and computational overheads \cite{10634552}. As shown in Fig. \ref{fig:federatededgelearning}, the cloud {generates a lightweight proxy model using offsite-tuning transfer learning, encoding} task-specific knowledge with a minimal number of parameters. Edge nodes then fine-tune the PEFT module of this proxy model to further reduce communication and computational costs.

\textit{b) Distributed Cooperative LM Inference for LM Agents.} On-server LM inference offloads computation to cloud servers but compromises user privacy by requiring raw data uploads. On-device inference, on the other hand, preserves privacy but imposes a heavy computational burden on edge devices. To resolve this dilemma in resource-constrained environments, mixture-of-experts (MoE) and split inference are two key approaches. Additionally, LM agents can leverage role-playing for cooperative inference in general applications.

\begin{figure}[!tp]
\centering \setlength{\abovecaptionskip}{-0.cm}
  \includegraphics[width=7.5cm]{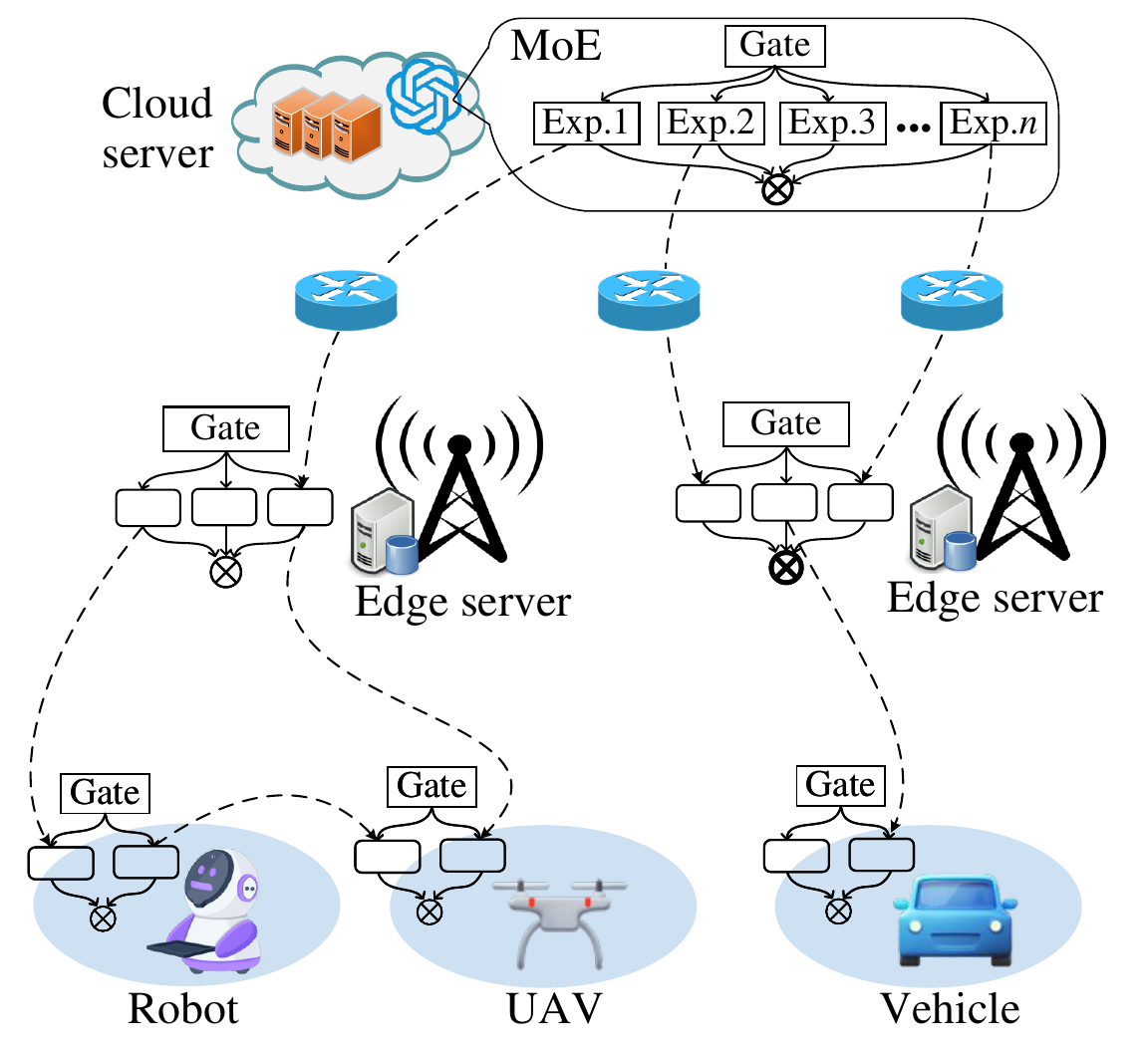}
  \caption{{Illustration of MoE-oriented collaborative LM inference for LM agents in cloud-edge-end environments.}}\label{fig:MoEinference}\vspace{-3mm}
\end{figure}

\textit{{\ding{172}} Cooperative LM inference via MoE.} MoE enhances LMs by partitioning them into specialized expert networks, each tailored to process specific input types \cite{fedus2022switch}. A gating network dynamically assigns tokens to the most appropriate expert(s), enabling sparse activation that reduces unnecessary computations and improves efficiency. MoE has been shown to boost performance by 45\% while consuming only one-third of the resources required by dense models \cite{shen2024mixtureofexperts}. In cloud-edge architectures, as shown in Fig.~\ref{fig:MoEinference}, cloud servers store the expert models, and edge servers and devices download only the relevant experts for local inference. This method optimizes resource utilization, accelerates processing, and makes the deployment of LMs viable under resource-constrained environments. For instance, EdgeMoE \cite{yi2023edgemoe} stores non-expert weights in memory while placing larger expert weights on disk, activating only the necessary experts. This approach reduces memory consumption by 2.6 to 3.2 times compared to full model loading. 
However, a series of challenges remain in edge deployment, as follows.
\begin{itemize}
    \item {Heterogeneous edge devices:} MoE models are primarily designed for high-performance cloud data centers equipped with powerful GPUs and ample network bandwidth. However, when deployed on edge devices, MoE models must accommodate the wide variety of hardware capabilities, ranging from basic IoT devices to advanced edge servers. This heterogeneity requires flexibility in MoE design to ensure high performance while adapting to the varying resource constraints of edge environments. %
    
    \item {Dynamic edge inference:} Edge environments are highly dynamic, which complicates inference tasks. When local expert networks are unavailable, they must be dynamically downloaded, a process influenced by network speed and device capacity. Additionally, selecting the appropriate expert network adds complexity to model verification and management. 
    
    \item {Privacy disclosure:} As the expert networks required for inference tasks may also expose users' environmental context and behavioral patterns, it is essential to incorporate privacy protection measures when designing the inference engine.
\end{itemize}

\textit{{\ding{173}} Cooperative split inference.} Split inference, as outlined in the 3GPP 5G specifications \cite{3gpp.22.874}, partitions an LM between edge devices and edge/cloud servers, offloading major part of computation to edge/cloud servers for mitigated device-side workload. It offers a middle ground between on-server and on-device inference. As shown in Fig.~\ref{fig:splitLM}(b), edge devices perform initial inference using the user-side submodel and upload intermediate features to the server for further processing, thereby optimizing resource usage while preserving privacy by keeping sensitive raw data on the device.
According to \cite{qu2024mobileedgeintelligencelarge}, split inference for LLMs can be implemented in two ways: 
\begin{itemize}
    \item In encoder-decoder LMs such as BART \cite{lewis2020bart}, the encoder is placed on the device, while the more computationally demanding decoder is on the server.
    \item In decoder-only LMs such as GPT-series, lightweight components (e.g., embeddings, early Transformer blocks) stay on the device, while heavier components are offloaded to the server. Additionally, critical feed-forward network (FFN) parameters can be stored on the device to save memory.
\end{itemize}
    
However, split inference faces challenges such as high communication costs due to large data uploads, latency from reliance on both edge devices and servers, and privacy concerns from potential leakage of intermediate outputs.
Early exit strategies help lower computational latency on edge devices and servers. Transformer outputs can further be compressed via quantization, pruning, and merging \cite{zhu2024surveymodelcompress}. Additionally, progressive split inference \cite{lan2022progressive} can eliminate unnecessary transmission of intermediate token representations while maintaining desired inference accuracy.

\begin{figure}[!tp]
\centering \setlength{\abovecaptionskip}{-0.cm}
  \includegraphics[width=8.5cm]{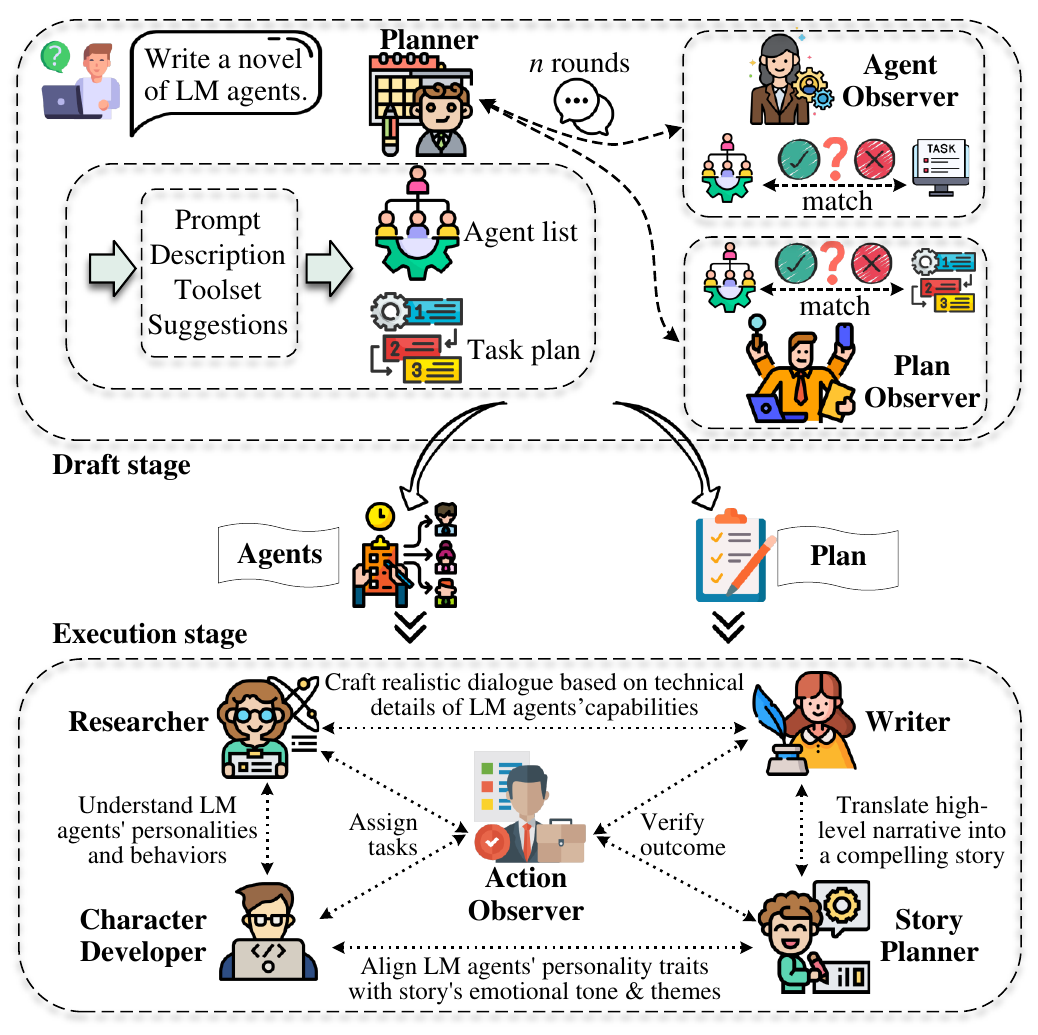}
  \caption{An example of role-playing for novel writing with collaborative agents \cite{AutoAgents2024IJCAI}. In the draft stage, the Planner assembles an initial agent team and action plan, refining it via feedback from Agent and Plan Observers. The Action Observer assigns tasks, monitors execution, and adjusts the plan based on performance. Agents collaboratively execute tasks, adapting as needed.}\label{fig:roleplaying}\vspace{-2mm}
\end{figure}

\textit{{\ding{174}} Role-playing.} Role-playing \cite{Park2023GenerativeAgents,qian2024chatdev} is a typical method for facilitating cooperation among LM agents. 
\begin{itemize}
    \item \textit{Instruction-following cooperation.} Li \textit{et al.} \cite{li2023camel} present a cooperative agent architecture that employs role-playing combined with inception prompting to guide LLM agents autonomously toward effective task completion. The system begins with human-provided ideas and role assignments, refined by a task-specifier agent. An AI user and an AI assistant then collaborate through multi-turn dialogues, with the AI user instructing and the assistant responding until the task is completed.
    
    \item \textit{Instruction-free cooperation.} Fig.~\ref{fig:roleplaying} shows an example of role-playing based novel writing using collaborative agents \cite{AutoAgents2024IJCAI}. ChatDev \cite{qian2024chatdev} is a virtual software company operated by ``software agents'' with diverse roles, e.g., chief officers, programmers, and test engineers. Following the waterfall model, it divides development into design, coding, testing, and documentation, breaking each phase into atomic subtasks managed by a chat chain. To create cooperative agents with adaptive behaviors, ProAgent \cite{ZhangProAgent2024} assigns each agent three roles: planner, verificator, and controller. The planner formulates high-level skills, incorporating belief revision and knowledge management. The verificator assesses the feasibility of skills, identifies issues if a skill fails, and initiates re-planning when needed. If the skill is feasible, the controller decomposes it into executable actions. Park \textit{et al.} \cite{Park2023GenerativeAgents} create a community of 25 generative agents in a sandbox world named Smallville, where agents are represented by sprite avatars. These agents perform daily tasks, form opinions, and interact, mimicking human-like behaviors.
\end{itemize}

\textit{c) Distributed Cooperative LM Caching for LM Agents.} Edge model caching can significantly reduce model download latency by pre-distributing LMs to wireless edge servers, allowing end users to download them directly. This strategy provides users with quick access to LMs within their quality-of-service (QoS) requirements, bypassing the need to retrieve them from remote cloud servers, which would incur high latency.

\textit{{\ding{172}} Edge LM caching.} Due to the limited storage capacity, edge servers need to optimize the cache hit ratio by placing popular LMs on edge servers. For instance, TrimCaching \cite{qu2024trimcaching} optimizes LLM model placement by leveraging LLM parameter sharing to maximize cache efficiency under storage and latency constraints. By storing a single copy of shared parameter blocks on each edge server, it enhances storage utilization. Practical edge LM caching deployment also needs to consider user mobility patterns and inter-edge collaboration and competition strategies under different LM partitioning methods (e.g., MoE and SL).

\textit{{\ding{173}} Edge LM replacement.} As user demands evolve, previously cached LMs may no longer match current request patterns. In such cases, edge servers should replace older models with more relevant ones. This process incurs high communication costs, as transferring large-scale models can impose substantial overhead on mobile backhaul networks. Recency-based and frequency-based are two classic strategies, replacing the least recently used (LRU) or least frequently used (LFU) items with updated ones. Popularity-based methods, such as SwapMoE \cite{kong-etal-2024-swapmoe}, prioritize caching highly activated expert networks, with importance determined by activation frequency during inference tasks. However, these approaches often neglect shared parameter blocks across LLMs and the potential for cooperative caching among edge nodes. Distributed reinforcement learning can be leveraged to make dynamic replacement decisions without requiring complete information from other edge nodes.

\textit{{\ding{174}} Edge LM routing.} To further minimize service latency, it is essential to optimize edge LM routing strategies that direct users to the most suitable edge server based on network conditions, queue length, and model availability. Additionally, due to the mobility of agents, it requires seamless model migration across edge servers to maintain optimal service delivery. Further research efforts are required for intelligently managing user requests and distributing LM workloads across edge nodes to enhance performance, minimize latency, and maximize resource utilization.

\textit{{\ding{175}} Edge caching for RAG.} LM agents' performance is often tightly linked to the quality and relevance of the knowledge they possess. Therefore, selecting which data to cache for LM updates via RAG is essential to ensure that LM agents provide reliable, up-to-date services. Efficient knowledge cache management not only improves response times but also supports the continuous learning and adaptation of LM agents in dynamic environments. Further research is needed to tackle the joint LM caching and knowledge caching problem within the context of RAG under practical constraints of edge environments.


\begin{figure*}[!tp]
\centering \setlength{\abovecaptionskip}{-0.cm}
  \includegraphics[width=12cm]{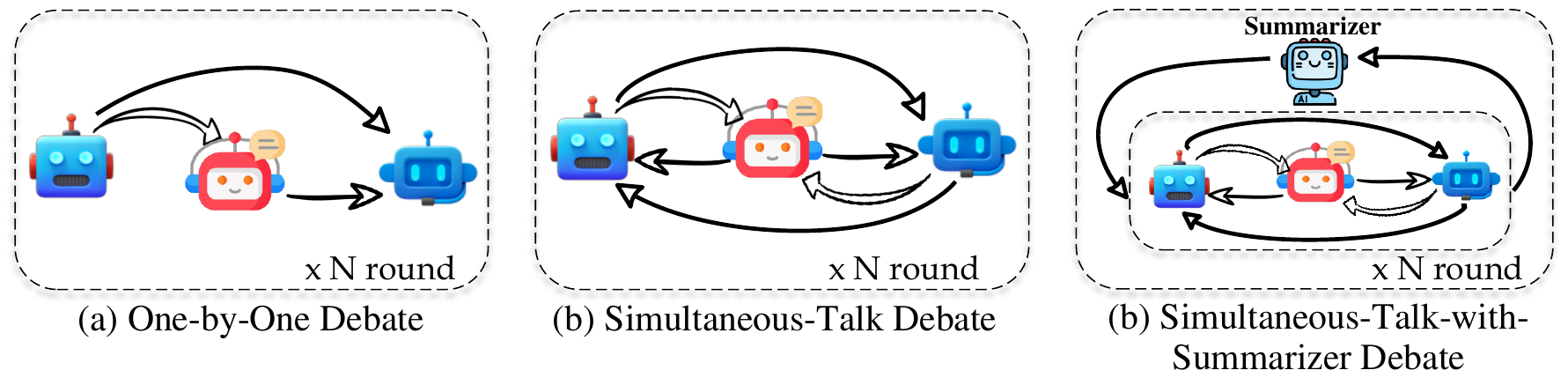}
  \caption{{Illustration of (a) one-by-one debate, (b) simultaneous-talk debate, and (c) simultaneous-talk-with-summarizer debate \cite{chan2023ChatEval}.}}\label{fig:MADtype}\vspace{-2mm}
\end{figure*}

\subsubsection{Distributed Competition}
Competition involves LM agents pursuing their individual objectives, often at the expense of others. This interaction mode is characterized by non-cooperative strategies where agents aim to maximize their own benefits.
Multi-agent debate (MAD) and non-cooperative game have been widely adopted to model the competitive behaviors between autonomous agents.

\textit{{\ding{172}} MAD.} The cognitive behaviors of LLMs, such as self-reflection, have proven effective in solving NLP tasks but can also lead to thought degeneration due to biases, rigidity, and limited feedback. MAD \cite{chan2023ChatEval} {introduces structured arguments and debates among LM agents, where} they defend their positions and challenge each other's strategies. {By engaging in a dynamic tit-for-tat process,} LM agents can correct biases, overcome resistance to change, and provide mutual feedback, {leading to improved reasoning and decision-making.} In MAD, diverse role prompts (i.e., personas) {play a crucial role, shaping agents' perspectives and debate strategies. 
Generally, as illustrated in Fig.~\ref{fig:MADtype},} there are three primary communication strategies: 
\begin{itemize}
    \item \textit{One-by-one debate strategy:} Agents respond sequentially in a fixed order, generating replies based on the ongoing discussion. Each agent’s chat history is updated with previous responses to maintain context.
    \item \textit{Simultaneous-talk debate strategy:} Agents respond asynchronously within each round, mitigating the influence of speaking order on the debate's flow.
    \item \textit{Simultaneous-talk-with-summarizer debate strategy:} The simultaneous-talk-with-summarizer enhances simultaneous-talk by incorporating a summarizer LLM. After each round, the summarizer condenses the discussion into a cohesive summary, {replacing individual chat histories to ensure a shared and coherent context for all agents.}
\end{itemize}
    
\textit{{\ding{173}} Non-cooperative game.} In non-cooperative games, LM agents engage in strategic decision-making where each LM agent's goal is to seek the Nash equilibrium. Various non-cooperative games can be employed such as Nash bargaining, auctions, matching games.
For instance, Xu \emph{et al.} \cite{xu2024cached} design a new cached model-as-a-resource paradigm and introduce the concept of age-of-thought to efficiently deploy LLM agents. They also devise a deep Q-network-based auction mechanism to incentivize network operators.

\subsubsection{Partial Cooperation}
Partial cooperation occurs when LM agents collaborate to a limited extent, often driven by overlapping but not fully aligned interests \cite{talebirad2023multi}. In such scenarios, agents might share certain resources or information while retaining autonomy over other aspects. This interaction mode balances the benefits of cooperation with the need for individual agency and competitive advantage. Partial cooperation can be strategically advantageous in environments where complete cooperation is impractical or undesirable due to conflicting goals or resource constraints. For instance, DyLAN \cite{liu2023dynamic} is a multi-layered LLM agent network for collaborative task-solving with teammates, such as reasoning and code generation. DyLAN facilitates multi-round interactions in a dynamic setup with an LLM-empowered ranker to deactivate low-performing agents, an early-stopping mechanism based on Byzantine consensus to efficiently reach agreement, and an automatic optimizer to select the best agents based on agent importance scores. 

Hierarchical game and coalition formation theory can be employed to model partially cooperative interactions among LM agents.
Hierarchical game \cite{10037213} structures the interactions of LM agents across different game stages, where they may cooperate at one level and compete at another. Through the coalition formation theory \cite{5740907}, LM agents can form the optimal stable coalition structure under different scenarios and environments, where the optimal stable coalition structure is featured with intra-coalition cooperation and inter-coalition competition.

\subsubsection{Lessons Learned}
{In distributed collaboration, SL and MoE focus on \textit{model segmentation} to significantly reduce computational burdens on edge devices, while FL and role-playing emphasize \textit{model integration} to enhance cooperation among distributed agents. 
Within the overall cloud-edge collaboration for LM agent systems, key challenges emerge in the joint consideration of LM segmentation (e.g., SL and MoE), LM placement, LM replacement, LM routing, and knowledge caching, alongside the inherent heterogeneity and dynamics of edge environments.
Effective deployment and operation of LM agents across cloud and edge infrastructures require a delicate balance between LM segmentation and LM integration strategies, with an emphasis on optimizing resource allocation and minimizing service latency. Additionally, solutions should be adaptable to the constantly changing network conditions and hardware limitations inherent to edge devices. Future research includes the seamless integration of these techniques, taking into account both static and dynamic aspects of edge networks, to offer more efficient and scalable solutions for large-scale applications.}

\begin{figure}[!tp]
\centering 
  \includegraphics[width=9.5cm]{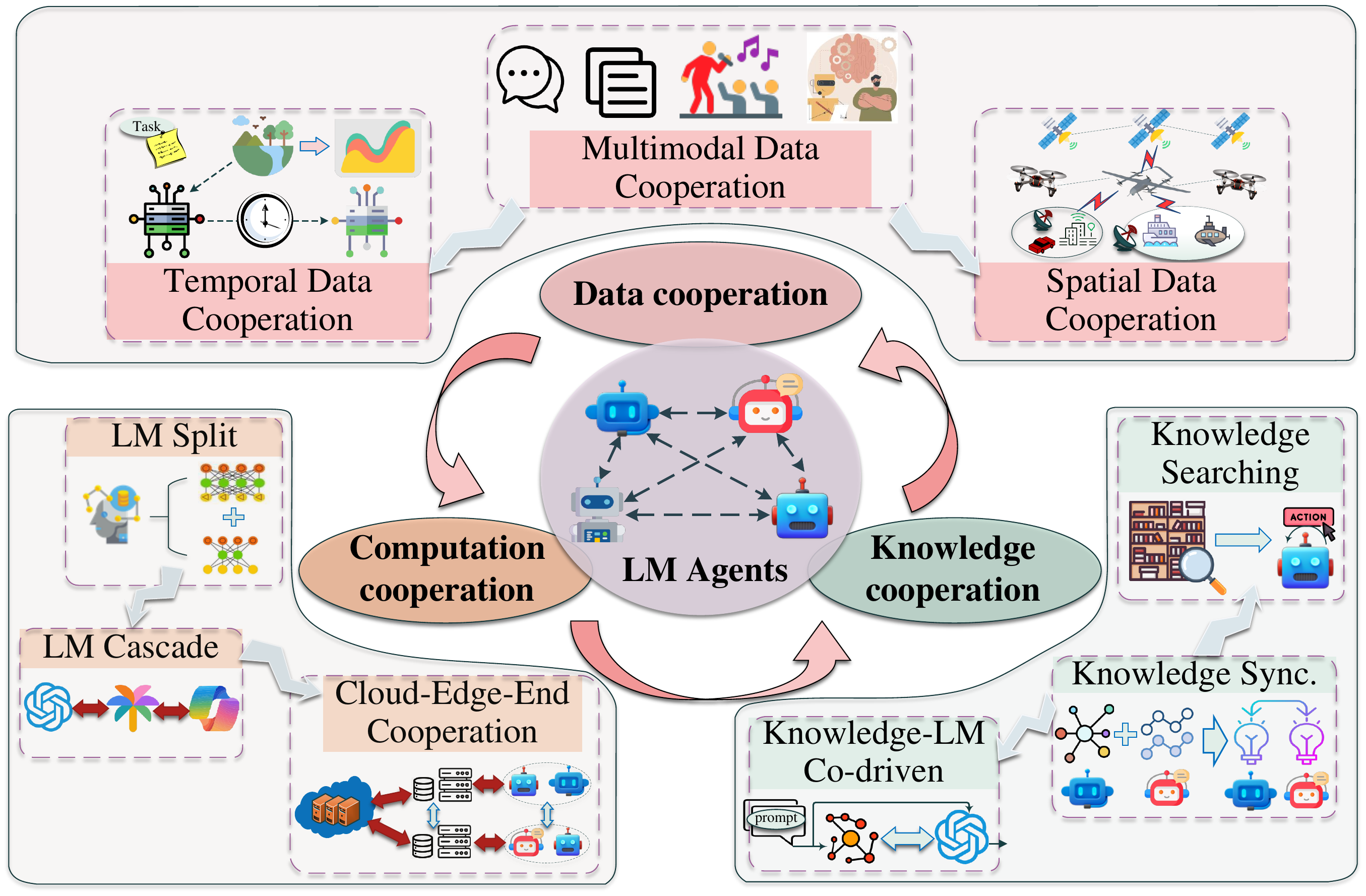}
  \caption{Illustration of the cooperation modes of connected LM agents including data cooperation, computation cooperation, and knowledge cooperation.}\label{fig:cooperation_mode}\vspace{-3mm}
\end{figure}

\subsection{Cooperation Paradigms of Connected LM Agents}\label{subsec:CoopParadigm}
As illustrated in Fig.~\ref{fig:cooperation_mode} {and Table~\ref{3CoopComparison}}, the detailed cooperation paradigms of LM agents under cloud-edge-end architecture involve three perspectives: data cooperation, computation cooperation, and knowledge cooperation.
\begin{itemize}
    \item \textit{Data Cooperation}: Within a common task, LM agents continuously exchange and fuse their individual data (e.g., task-oriented sensory data) to ensure a comprehensive and up-to-date understanding of their task environment, thereby enhancing collective intelligence and enabling coordinated actions. 
    \item \textit{Computation Cooperation}: LM agents perform coordinated reasoning, such as service cascades, by distributing computational tasks optimally across agents, leveraging collective processing power to handle complex computations more efficiently. 
    \item \textit{Knowledge Cooperation}: LM agents share domain-specific knowledge and experiences (e.g., in the format of KGs) to collectively improve problem-solving capabilities for better-informed actions and decisions, via knowledge synchronization, fusion, and retrieval and distributed learning algorithms. 
\end{itemize}

\begin{figure*}[!t]
\centering \setlength{\abovecaptionskip}{-0.cm}
  \includegraphics[width=0.82\linewidth]{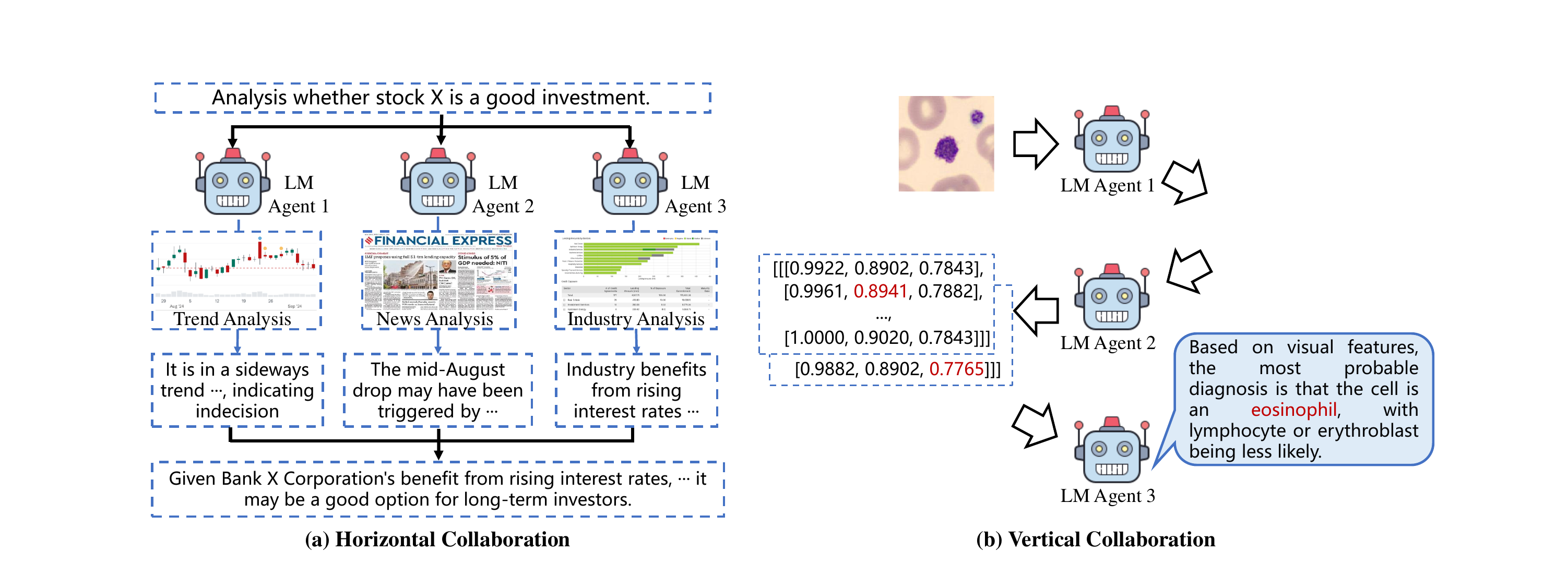}
  \caption{Illustration of (a) horizontal collaboration and (b) vertical collaboration paradigms for multiple LM agents.}\label{fig:hori_ver_collaboration}\vspace{-2mm}
\end{figure*}

\textit{1) Data Cooperation for LM Agents:}
The data cooperation among LM agents involves the \textit{modality} perspective and the \textit{spatio-temporal} perspective.


\textit{a) Multimodal Data Cooperation} emphasizes the fusion of data from various modalities, such as text, images, audio, and video, to offer a comprehensive understanding of the environment. This cooperation allows LM agents to process and interpret information from multiple sources, leading to more accurate and robust decision-making.
(i) By combining data from different modalities, it helps to create a unified representation which harnesses the strengths of each type of data. For instance, Gross \textit{et al.} \cite{10180363} discuss the use of multimodal data to model communication in artificial social agents, emphasizing the importance of verbal and nonverbal cues for natural human-robot interaction. (ii) By enabling LM agents to retrieve relevant information across different modalities, it enhances their ability to respond to complex queries and scenarios. For instance, by considering the intra-modality similarities in multi-modal video representations, Zolfaghari \textit{et al.} \cite{9711472} introduce the contrastive loss in contrastive learning process for enhanced cross-modal embedding, whose effectiveness is validated using LSMDC and YouCook2 datasets for video-text retrieval and video captioning tasks.

\textit{b) Spatio-temporal Data Cooperation} involves the integration and synchronization of spatial and temporal data across various modalities and sources, enabling LM agents to achieve a comprehensive and dynamic understanding of the environment over time. This cooperation ensures that LM agents can effectively analyze patterns, predict future states, and make informed decisions in real-time, based on both spatial distribution and temporal evolution of data.
Yang \textit{et al.} \cite{Yang_2023_ICCV} introduce SCOPE, a collaborative perception mechanism that enhances spatio-temporal awareness among on-road agents through end-to-end aggregation. SCOPE excels by leveraging temporal semantic cues, integrating spatial information from diverse agents, and adaptively fusing multi-source representations to improve accuracy and robustness. However, \cite{Yang_2023_ICCV} mainly works for small-scale scenarios.
By capturing both spatial and temporal heterogeneity of citywide traffc, Ji \textit{et al.} \cite{ji2023spatio} propose a novel spatio-temporal self-supervised learning framework for traffic prediction that improves representation of traffic patterns. This framework uses an integrated module combining temporal and spatial convolutions and employs adaptive augmentation of traffic graph data, supported by two auxiliary self-supervised learning tasks to improve prediction accuracy.

\textit{2) Computation Cooperation for LM Agents:}
It can be classified into three modes: \textit{horizontal} cooperation, \textit{vertical} cooperation, and \textit{hybrid} cooperation.

\textit{a) Horizontal Collaboration.} It refers to decompose complex tasks into manageable sub-tasks, with multiple LM agents independently completing their assigned tasks in parallel. The results are then summarized and integrated to generate the final outcome. This approach enables the collaborative system to scale horizontally by adding more LM agents, allowing for more complex and dynamic tasks to be handled. The parallel processing and multi-angle perspectives contribute to increased robustness and reduced errors and biases that may arise from relying on a single agent.
As shown in Fig.~\ref{fig:hori_ver_collaboration}(a), each LM agent independently analyzes different aspects of stock performance such as trend, news, and industry to assess its investment potential.
Horizontal collaboration has been successfully implemented in various LM agent systems. For instance, ProAgent \cite{ZhangProAgent2024} employs a framework where LM agents collaborate as teammates, analyzing each other's intentions and updating their beliefs based on observed behaviors of their peers. This enhances collective decision-making by allowing agents to adapt dynamically in real time. Similarly, DyLAN \cite{liu2023dynamic} assembles a team of strategic agents that communicate through task-specific queries, allowing multiple rounds of interaction to enhance both efficiency and overall performance.
However, the independent analysis of each LM agent may lead to inconsistent outputs, complicating the aggregation of the final conclusions. Therefore, effective mechanisms are needed to address disagreements and ensure that the agents' contributions are complementary, which may increase the complexity of the coordination process.

\textit{b) Vertical Collaboration.} It involves decomposing complex tasks into multiple stages, with different LM agents handling each stage sequentially. After completing their respective tasks, each agent passes the result to the next agent, until the entire task is successfully completed. As illustrated in Fig.~\ref{fig:hori_ver_collaboration}(b), in medical image analysis, the first agent might extract basic visual features, the second agent could analyze these features in greater depth, and the final agent produces a diagnosis. This sequential approach enables a step-by-step refinement, allowing complex problems to be broken down and tackled more efficiently by leveraging specialized agents at each stage.
For instance, Jiang \emph{et al.} \cite{10638533} present CommLLM, a multi-agent system for natural language-based communication tasks. It comprises three key components: i) multi-agent data retrieval, utilizing condensate and inference agents to refine 6G communication knowledge; ii) multi-agent collaborative planning, employing various planning agents to generate solutions from multiple viewpoints; and iii) multi-agent reflection, which evaluates solutions and suggests improvements through reflexion and refinement agents. However, a limitation of vertical collaboration is that higher-level LM agents rely on the accuracy of lower-level outputs, making the process vulnerable to error propagation. Mistakes made early in the sequence can compromise the final outcome, highlighting the need for high precision at each stage to maintain the overall system's reliability.

\begin{table*}[!t]
   \centering
    \caption{{A Summary of Data, Computation, and Knowledge Cooperation Approaches for LM Agents}}\label{3CoopComparison}
\begin{tabular}{|l|c|c|l|c|}
\hline
 &
  \textbf{Stage} &
  \textbf{Type} &
  \textbf{Technology} &
  \textbf{Ref.} \\ \hline
\multirow{4}{*}{\rotatebox{90}{Data}} &
  \multirow{2}{*}{\begin{tabular}[c]{@{}c@{}}Multimodal Data \\ Cooperation\end{tabular}} &
  \begin{tabular}[c]{@{}c@{}}Modality alignment\\ \& merging\end{tabular} &
  Use multimodal data to model communication in social AI agents. &
  \cite{10180363} \\ \cline{3-5} 
 &
   &
  Multimodal retrieval &
  Utilize contrastive learning for cross-modal embedding. &
  \cite{9711472} \\ \cline{2-5} 
 &
  \multirow{2}{*}{\begin{tabular}[c]{@{}c@{}}Spatio-temporal \\ Data Cooperation\end{tabular}} &
  Collaborative perception &
  Combine temporal semantics \& spatial agent data for multi-source fusion. &
  \cite{Yang_2023_ICCV} \\ \cline{3-5} 
 &
   &
  Collaborative prediction &
  Combine temporal \& spatial graph convolutions for traffic prediction. &
  \cite{ji2023spatio} \\ \hline
\multirow{6}{*}{\rotatebox{90}{Computation}} &
  \multirow{2}{*}{\begin{tabular}[c]{@{}c@{}}Horizontal \\ Collaboration\end{tabular}} &
  Role-playing &
  Agents cooperate with diverse roles: Controller, Planner, \& Verificator. &
  \cite{ZhangProAgent2024} \\ \cline{3-5} 
 &
   &
  Dynamic agent team &
  Agents dynamically form a team via agent selection and evaluation. &
  \cite{liu2023dynamic} \\ \cline{2-5} 
 &
  \multirow{2}{*}{\begin{tabular}[c]{@{}c@{}}Vertical \\ Collaboration\end{tabular}} &
  CoT reasoning &
  Agents cooperatively reason complex task via CoT. &
  \cite{wei2023cot} \\ \cline{3-5} 
 &
   &
  Chained cooperation &
  Multi-agent cooperation through data retrieval, planning, and reflection. &
  \cite{10638533} \\ \cline{2-5} 
 &
  \multirow{2}{*}{\begin{tabular}[c]{@{}c@{}}Hybrid \\ Collaboration\end{tabular}} &
  ToT \&GoT reasoning &
  Agents cooperatively reason complex task via ToT and GoT. &
  \cite{yao2023tot,besta2024graph} \\ \cline{3-5} 
 &
   &
  DAG reasoning &
  Organize agents as DAGs for interactive reasoning. &
  \cite{qian2024scalingLLMcoll} \\ \hline
\multirow{10}{*}{\rotatebox{90}{Knowledge}} &
  \multirow{6}{*}{Knowledge Sharing} &
  Online knowledge distillation &
  LMs generate high-quality reasoning processes for models to learn. &
  \cite{kang2024knowledge} \\ \cline{3-5} 
 &
   &
  Offline knowledge distillation &
  Extract and align parameters of LLMs to smaller models. &
  \cite{zhong2023seeking} \\ \cline{3-5} 
 &
   &
  Explicit knowledge update &
  Adjust LM via knowledge bases and feedbacks. &
  \cite{tandon2021learning} \\ \cline{3-5} 
 &
   &
  Implicit knowledge update &
  Selective fine-tuning of model's internal parameter knowledge. &
  \cite{qin2022elle} \\ \cline{3-5} 
 &
   &
  Implicit knowledge fusion &
  Train target model via predictive probability distributions of LMs. &
  \cite{wan2024knowledge} \\ \cline{3-5} 
 &
   &
  Model output fusion &
  Directly aggregate the outputs of different models. &
  \cite{jiang2023llm} \\ \cline{2-5} 
 &
  \multirow{2}{*}{Knowledge Fusion} &
  KG completion &
  Encode text to better use textual \& semantic infomation. &
  \cite{shen2022joint} \\ \cline{3-5} 
 &
   &
  KG verification &
  Using a small LM to validate and correct the LM output. &
  \cite{han2023pive} \\ \cline{2-5} 
 &
  \multirow{2}{*}{Knowledge Retrieval} &
  Static knowledge base &
  Retrieve knowledge from static sources for accurate reasoning. &
  \cite{lewis2020retrieval,trivedi2022interleaving} \\ \cline{3-5} 
 &
   &
  Dynamic knowledge base &
  Use real-time updated knowledge sources for accurate reasoning. &
  \cite{dai2024vistarag, kuroki2024multiagentbehavior,wang2023self} \\ \hline
\end{tabular}
\end{table*}

\textit{c) Hybrid collaboration.} In practical LM agent environments, real-world applications often require a combination of horizontal and vertical collaboration, resulting in hybrid collaboration, as illustrated in Fig.~\ref{fig:roleplaying}. For instance, when addressing highly complex tasks, the problem is first broken down into manageable sub-tasks, each assigned to specialized LM agents. Horizontal collaboration allows agents to perform parallel evaluations or validations, while vertical collaboration ensures that the task is refined through sequential processing stages. This computational collaboration blends both paradigms, coordinating parallel assessments across agents while sequentially integrating their outputs through defined stages, thus optimizing task execution and enhancing overall system performance.
For instance, Chen \textit{et al.} \cite{qian2024scalingLLMcoll} propose a dynamic multi-agent collaboration framework that organizes agents using directed acyclic graphs (DAGs) to facilitate interactive reasoning. Their framework demonstrates superior performance across various network topologies and enables collaboration among thousands of agents. A key finding in \cite{qian2024scalingLLMcoll} is the discovery of the collaborative scaling law, where solution quality improves in a logistic growth pattern as more agents are added, with collaborative emergence occurring faster than neural emergence.

\textit{3) Knowledge Cooperation for LM Agents: }
Knowledge can be broadly categorized into \textit{explicit knowledge} and \textit{implicit knowledge}. Explicit knowledge refers to structured, codified, and easily accessible information that can be articulated, documented, and shared, such as external databases and KGs. Implicit knowledge is embedded within the model and arises through the optimization of its internal parameters, such as weights and biases, during training or fine-tuning \cite{DBLP:conf/emnlp/RobertsRS20}.
The knowledge cooperation generally involves three consecutive aspects: knowledge synchronization, knowledge fusion, and knowledge retrieval.

\textit{a) Knowledge sharing.}
It includes the sharing and updating of knowledge between multiple LM agents to ensure consistency in decision making.

\textit{{\ding{172} Knowledge transfer}} is a common method of knowledge sharing that transfers the implicit knowledge, in the form of learned parameters, from one LM agent to another.
Kang \emph{et al.} \cite{kang2024knowledge} propose an online distillation method that enhances LLMs by retrieving relevant knowledge from external knowledge bases, to generate high-quality reasoning processes. The knowledge (e.g., reasoning results) of small language models can be leveraged to improve the performance of LLMs in knowledge-intensive tasks. However, as the scale of LLMs grows, LLM training via such knowledge transfer becomes complex and computationally intensive.
To address this issue, Zhong \emph{et al.} \cite{zhong2023seeking} propose a parametric knowledge transfer method that extracts and aligns knowledge parameters using sensitivity techniques and uses the LoRA module as an intermediary mechanism to inject this knowledge into smaller models, transferring the implicit knowledge of smaller models.

\textit{{\ding{173} Knowledge update.}} Knowledge alignment is a prerequisite to update knowledge among LM agents. Zhang \emph{et al.} \cite{zhang2023autoalign} propose a fully automatic KG alignment method, using LLMs to identify and align entities in different KGs, aiming to solve the heterogeneity problem between different KGs and integrate multi-source data.
(i) For explicit knowledge updates, Tandon \emph{et al.} \cite{tandon2021learning} pair LMs with a growing memory to train a correction model, where users identify output errors and provide general feedbacks on how to correct them.
(ii) For implicit knowledge updates, continual learning ensures that AI models can continuously learn while receiving new tasks and new data without forgetting previously learned knowledge.
Qin \emph{et al.} \cite{qin2022elle} propose ELLE, which flexibly extends the breadth and depth of existing PLMs, allowing the model to continuously grow as new data flows in.

\textit{b) Knowledge fusion.} It contains the knowledge fusion and completion phases between multiple LM agents.

\textit{{\ding{172} Knowledge integration}}, which integrates and optimizes knowledge from different LM agents to form a robust and comprehensive public knowledge base for a common task. Jiang \emph{et al.} \cite{jiang2023llm} propose an ensemble framework named LLM-blender that directly aggregates the outputs of multiple models for enhanced prediction accuracy and robustness. However, it necessitates maintaining several trained LLMs and running each LLM during inference, making it less practical for LM agents. To address this issue, Wan \emph{et al.} \cite{wan2024knowledge} propose an implicit knowledge fusion framework that evaluates the predictive probability distributions of multiple LLMs and uses the distribution for continuous training the target model.

\textit{{\ding{173} Knowledge completion,}} which infers missing facts in a given KG. Compared with traditional KG completion methods, LLMs hold the potential to enhance KG completion performance via encoding text or generating facts.
Shen \emph{et al.} \cite{shen2022joint} use LLMs as encoders, primarily capturing the semantic information of KG triples through the model's forward pass, and then reconstructing the KG's structure by calculating a loss function, thereby better integrating semantic and structural information.
Apart from KG completion, LLMs can be employed to verify KGs. Han \emph{et al.} \cite{han2023pive} propose a prompt framework for iterative KG verification, using small LLMs to correct errors in KG generated by LLMs such as ChatGPT.

\textit{c) Knowledge retrieval.}
LM agents not only rely on the learned knowledge during pre-training but also can dynamically access and query external knowledge bases (e.g., databases, the Internet, and KGs) to obtain the latest information to help reasoning. 
RAG technology combines information retrieval and generation models by first retrieving relevant contents and then generating answers based on these contents.
Based on the data type, it can be divided into two categories.

\textit{\ding{172} RAG based on static knowledge sources}, such as Wikipedia and documents. Lewis \emph{et al.} \cite{lewis2020retrieval} demonstrate how RAG generates accurate and contextually relevant responses by retrieving relevant knowledge chunks from static sources. For retrieval enhancement, Trivedi \emph{et al.} \cite{trivedi2022interleaving} propose a new multi-step QA method named IRCoT to interleave retrieval with steps in CoT, which first utilizes CoT to guide retrieval and then uses retrieval outcomes to enhance CoT.

\textit{\ding{173} RAG based on dynamic knowledge sources}, such as news APIs. It contains three lines: exploring new knowledge, retrieving past knowledge, and self-guided retrieval.
\begin{itemize}
    \item For \textit{new knowledge exploration}, Dai \emph{et al.} \cite{dai2024vistarag} use RAG technology for improved safety and reliability of autonomous driving systems by utilizing real-time updated data sources, including in-vehicle sensor data, traffic information, and other driving-related dynamic data. It demonstrates RAG's potential in handling complex environments and responding to unexpected situations.
    \item For \textit{past knowledge retrieval}, Kuroki \textit{et al.} \cite{kuroki2024multiagentbehavior} develop a novel vector database named coordination skill database to efficiently retrieve past memories in multi-agent scenarios to adapt to new cooperation missions.
    \item For \textit{self-guided retrieval}, Wang \emph{et al.} \cite{wang2023self} propose a method called self-knowledge-guided retrieval (SKGR), which harnesses both internal and external knowledge for enhanced retrieval by allowing LLMs to adaptively call external resources when handling new problems.
\end{itemize}

\begin{figure*}[!t]
\centering \setlength{\abovecaptionskip}{-0.cm}
  \includegraphics[width=\textwidth]{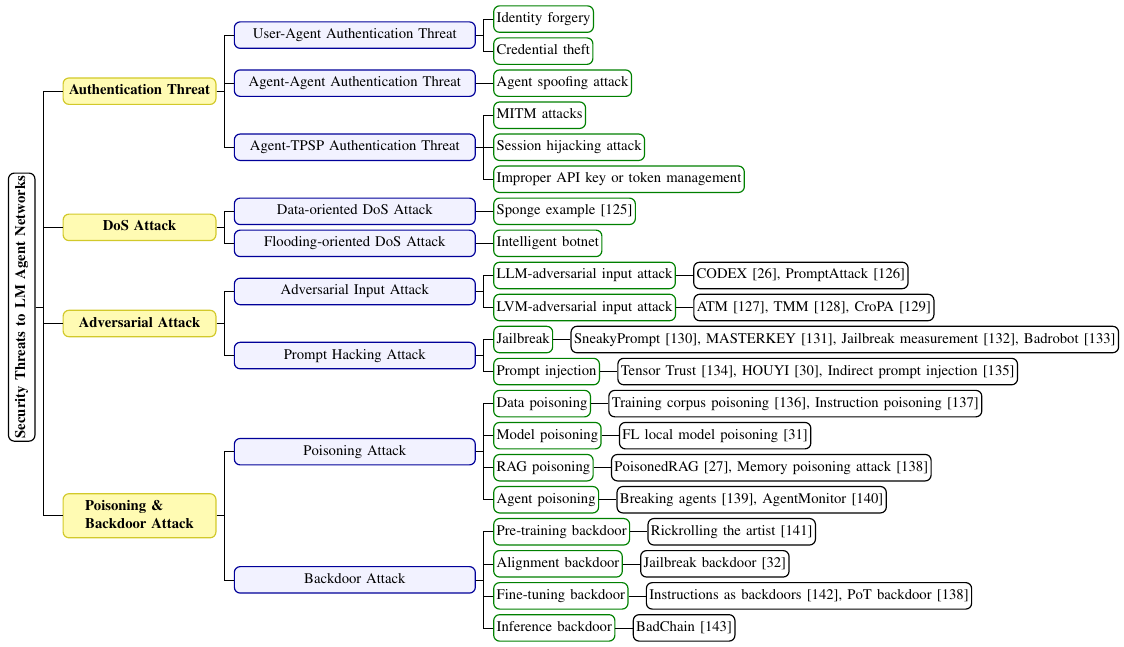}
	\caption{The taxonomy of security threats to LM agent networks.}
	\label{fig: security threats}\vspace{-2mm}
\end{figure*}

\subsection{{Summary and Lessons Learned}}
{With the rapid advancement of LMs, the future will witness a proliferation of personal assistant agents and embodied agents such as robots.}
{LM agent networks bridge the human, physical, and cyber worlds, with an architecture that integrates cloud, edge, and end devices. Powered by advanced technologies including foundation models, knowledge-related technologies, interaction technologies, multi-agent collaboration, and digital twin technologies, LM agent networks enable seamless communication, sophisticated reasoning, and scalable task execution. By leveraging hierarchical frameworks, semantic-aware communication, information-centric routing, and spatiotemporal awareness, LM agent networks optimize resource allocation, enhance real-time decision-making, and improve both intra- and inter-agent communication.
\begin{itemize}
    \item Intra-agent communication allows different components of a single LM agent, such as planning, action, memory, and security modules, to function cohesively. Frameworks such as the pub/sub mechanism facilitate efficient data exchange based on relevance. Inter-agent communication enables multiple LM agents to collaborate, share resources, and coordinate actions for collective intelligence. Protocols such as FIPA ACL and KQML define the vocabulary, message structure, and interaction strategies essential for effective communication between agents. Communication protocols should align with network environments (e.g., wireless or wired), latency requirements, robustness, and security needs. Adapting these protocols to various systems and agent platforms presents a significant challenge, highlighting the need for flexible, interoperable solutions. 
    \item The cloud-edge-end architecture allows for scalable and efficient task allocation, addressing the varying computational capabilities and real-time requirements of different devices.
    \item Horizontal, vertical, and hybrid collaborations provide complementary solutions to complex problems, combining distributed and sequential task processing to optimize overall system performance. This includes cooperative LM fine-tuning, where agents collaboratively adjust LMs using domain-specific data; cooperative LM inference, where inference tasks are distributed across agents to speed up processing; and collaborative edge LM caching, where LM (or LM parts) are stored and shared across edge devices to reduce latency and improve access efficiency. 
    \item The synchronization of implicit and explicit knowledge is crucial for enhancing the adaptability and reliability of LM agents, enabling them to respond effectively in dynamic, evolving environments. Implicit knowledge, often derived from experience and context, allows agents to quickly adapt to unforeseen situations, while explicit knowledge, i.e., structured, defined information, ensures agents can make informed decisions based on clear rules or facts. 
    \item {LM agent networks leverage distributed cloud-edge-end computing resources, enabling software/hardware agents at the edge and terminal levels to collaborate efficiently on complex tasks. This decentralized approach challenges the traditional trend of building ever-larger models by promoting modular, task-specific agents that can dynamically interact and share capabilities. Such a paradigm not only enhances computational efficiency and scalability but also improves adaptability across diverse application scenarios, paving the way for more flexible and resource-efficient AI ecosystems. However, significant challenges remain in designing agent capability notification, dynamic discovery, communication protocols, task orchestration, and agent routing across diverse scenarios, while also ensuring robust security and privacy protection.}
\end{itemize}}

\section{Security Threats \& Countermeasures to Large Model Agent Networks}\label{sec:Security}
In previous sections, we have explored the core concepts of LM agents and the importance of their networks for efficient collaboration and data/knowledge sharing. However, the three-tiered decentralized nature of LM agent networks, while offering scalability and flexibility, presents significant challenges related to secure and reliable collaboration throughout the life cycle of LM agent services.
In this section, we present a comprehensive review of security threats related to LM agent networks and examine the state-of-the-art countermeasures to defend against them. Fig. \ref{fig: security threats} illustrates the taxonomy of security threats to LM agent networks.


\begin{figure*}[!tp]
\centering 
  \includegraphics[width=14.6cm]{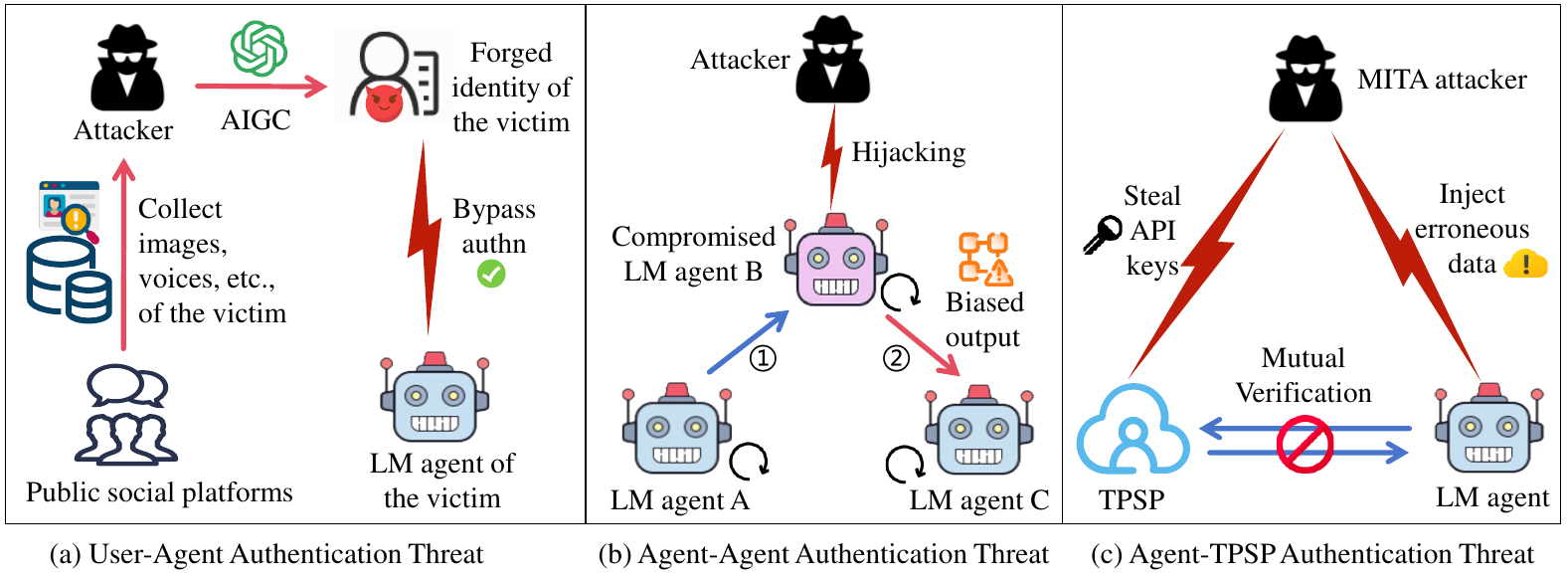}
  \caption{Illustration of authentication threats in LM agent networks: (a) user-agent authentication, (b) agent-agent authentication threats, and (c) agent-TPSP (third-party service provider) authentication.}\label{fig:Authn_mode}\vspace{-2mm}
\end{figure*}

\subsection{Authentication Threats in LM Agent Networks}


Efficient and reliable authentication mechanisms are a prerequisite for safeguarding the functionality and integrity of LM agent systems. Based on the interacting entities, authentication threats to LM agents can be categorized into three types: user-agent authentication threats, agent-agent authentication threats, and agent-TPSP (third-party service provider) authentication threats, as illustrated in Fig.~\ref{fig:Authn_mode}.

\subsubsection{User-Agent Authentication Threats}
LM agents generally store a wealth of private data for users and extensive sensitive corporate information for enterprises, making them prime targets for adversaries. Attackers can exploit identity forgery, credential theft, authentication bypassing, or privilege abuse to gain unauthorized access, leading to data breaches, system misuse, or service disruptions.
Unlike traditional authentication scenarios, adversaries can leverage the advanced generative capabilities of LM agents to enhance their ability to bypass authentication. For instance, as illustrated in Fig.~\ref{fig:Authn_mode}(a), an attacker could deploy an AIGC agent to forge a digital identity, imitating a legitimate user's voice, appearance, and behavior using publicly available media from social platforms to evade multi-factor authentication (MFA). If successful, the attacker could exploit the vast personal data within LM agents to construct a digital replica of the victim or access critical corporate assets, resulting in greater losses than traditional authentication threats.
Furthermore, as privatized LM agents become increasingly prevalent, interaction contexts are likely to contain even more personal and sensitive user information. This amplifies the impact of authentication threats, potentially leading to severe privacy violations and significant damage.

\subsubsection{Agent-Agent Authentication Threats}
{As LM agents increasingly collaborate to tackle complex tasks through horizontal, vertical, or hybrid coordination, authentication between agents becomes critical. One major threat is \textit{agent spoofing}, where adversaries compromise or impersonate legitimate agents within the collaboration workflow.} In both horizontal and vertical collaboration scenarios, attackers may compromise LM agents, converting them into malicious entities or impersonating legitimate ones to disrupt collaborative inference processes. As illustrated in Fig.~\ref{fig:Authn_mode}(b), it undermines accuracy and reliability of the collaboration system and jeopardizes task's outcome. Additionally, collusive LM agents can sabotage consensus-building, injecting misinformation or bias that disrupts the overall functionality and trustworthiness of the collaborative system.

\subsubsection{Agent-TPSP Authentication Threats}
LM agents often rely on external tools, web APIs, and third-party service providers (TPSPs) to execute diverse tasks, making secure and reliable authentication between agents and TPSPs critical. This process involves mutual verification: (i) TPSPs should confirm the agents' legitimacy and authorization, typically via a valid API key or token; (ii) agents should authenticate TPSPs to ensure they interact with trusted and legitimate service endpoints.
Authentication threats in this context primarily target the communication channel between LM agents and TPSPs, exposing them to interception-based attacks such as \textit{man-in-the-middle (MITM)} and \textit{session hijacking}. For instance, an adversary could intercept agent-TPSP communications, altering responses or injecting erroneous data, leading to unexpected outputs for users, as illustrated in Fig.~\ref{fig:Authn_mode}(c).
Additionally, improper API key or token management can exacerbate risks and result in unauthorized service misuse.

\subsubsection{Countermeasures to Authentication Threats in LM Agent Networks}
While existing technologies such as end-to-end encryption, certificates, MFA, and key management are effective in mitigating authentication threats in LM agent services, their effectiveness diminishes in the face of the growing complexity of LM agent collaboration and the evolving sophistication of attacks driven by LMs.
Moreover, the rapid development of LM agents has created a dual-edged dynamic, where both attacks and defenses are increasingly automated, escalating the risks and complexities of maintaining security. As a result, it is crucial to develop adaptive and collaborative resilient authentication mechanisms to guarantee the security and reliability of LM agent systems in complex, dynamic environments.
\begin{itemize}
\item \textit{Risk-Based Adaptive Authentication:}
This approach leverages the contextual reasoning capabilities of LM agents to dynamically assess the risk of user interactions. By analyzing user behavior, intent, and contextual anomalies, the authentication risk of the entity being authenticated is evaluated in real-time. Based on the risk level, the authenticator (such as the user, LM agent, or TPSP) adapts the authentication requirements, which may include triggering additional verification steps or involving third-party trusted institutions. This adaptive and flexible mechanism enhances the responsiveness of LM agents to varying levels of risk.

\item \textit{Multi-Agent Cooperative Authentication:}
This approach is designed for scenarios requiring long-term collaboration between LM agents. For instance, when multiple LM agents are working together to complete a task, each agent independently verifies the outputs or requests of the target agent for potential anomalies. The output is integrated into the collaborative workflow only after a sufficient number of agents confirm its reliability, based on a predefined threshold. This cooperative verification process reduces the risks of single-point failures, strengthens the robustness of authentication in dynamic agent networks, and mitigates the likelihood of adversarial manipulation.
\end{itemize}

\begin{figure}[!tp]
\centering 
  \includegraphics[width=8.0cm]{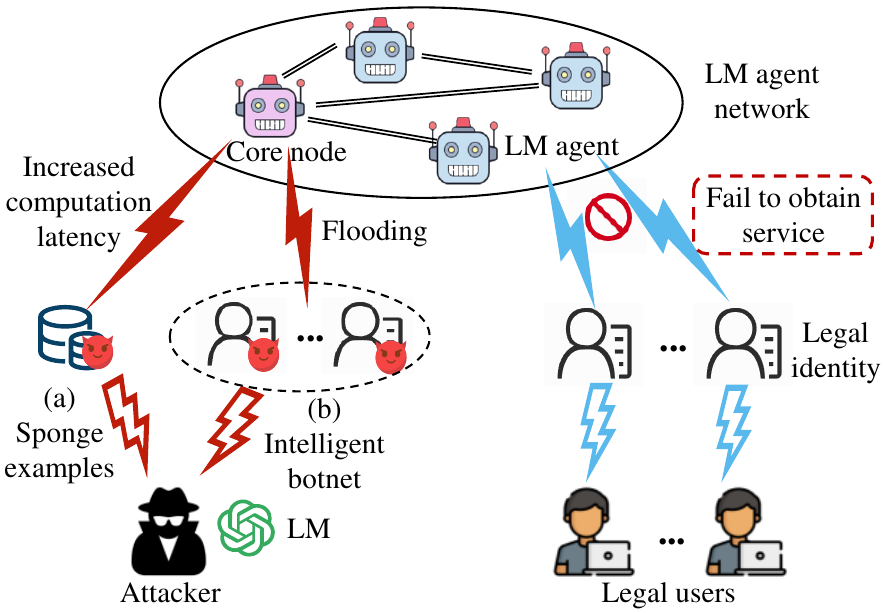}
  \caption{Illustration of DoS atttacks in LM agent networks: (a) sponge examples and (b) intelligent botnet.}\label{fig:DoS_mode}\vspace{-2mm}
\end{figure}

\subsection{Denial-of-Service (DoS) Attacks in LM Agent Networks}
The fine-tuning and inference processes of LM agents require significant computational resources. DoS attacks, however, can drastically increase resource consumption, overwhelming LM agents and disrupting service availability. Based on the attack method, DoS attacks in LM agent networks can be classified into two types: data-oriented DoS attacks (e.g., \textit{sponge examples}) and flooding-oriented DoS attacks (e.g., \textit{intelligent botnets}), as illustrated in Fig.~\ref{fig:DoS_mode}.

\subsubsection{Sponge Examples}
Sponge examples are adversarially crafted inputs that can significantly increase the energy consumption and computational latency of neural networks. Shumailov \textit{et al.} \cite{shumailov2021sponge} demonstrate how sponge examples can be used to execute DoS attacks on AI services, causing latency and energy consumption to rise by a factor of 30. LM agents are particularly vulnerable to such attacks, as adversaries can leverage LMs to generate large volumes of optimized sponge examples at a relatively low cost, thereby amplifying the impact and making the attack more difficult to mitigate.

\subsubsection{Intelligent Botnet}
A botnet refers to a network of compromised Internet-connected devices controlled by adversaries, which can be exploited for various malicious activities, including distributed DoS attacks. Intelligent botnets take advantage of advanced AI technologies, particularly in the context of LMs, to enhance their attack capabilities. Unlike traditional botnets, intelligent botnets utilize the comprehension abilities of LMs to analyze vulnerabilities within target systems, autonomously refine attack strategies, and optimize their methods for greater impact. For instance, an intelligent botnet may analyze a multi-agent collaborative system to identify the most critical agent in the workflow, then launch a flooding request attack on this agent, ultimately causing a collapse of the entire system.

\subsubsection{Countermeasures to DoS Attacks in LM Agent Networks}
Countermeasures to DoS attacks can be categorized into two main types: passive defense and active defense. Passive defense focuses on detecting and filtering malicious inputs, while active defense is exemplified by the use of honeypot technology.
\begin{itemize}
    \item \textit{Malicious input detection and filtering technologies} can effectively identify abnormal inputs, such as sponge examples, by comparing their energy consumption with that of natural inputs \cite{shumailov2021sponge}. Specifically, prior to deploying LM agents, an analysis of natural inputs can be conducted to assess the time and energy consumption during inference. Based on this analysis, the defender can set a cutoff threshold to limit the maximum energy consumption for each inference, thus mitigating the impact of malicious inputs (e.g., sponge examples).
    \item \textit{Honeypot technologies} can be employed in a multi-agent collaborative system by deploying a specific LM agent as an intelligent honeypot to lure adversaries and divert DoS attacks away from critical agents. The attacks directed at the honeypot agent would not affect the overall system’s availability. Additionally, through security traceability and behavioral analysis, defenders can study adversarial attack patterns and enhance the system’s defense mechanisms against future threats.
\end{itemize}

\subsection{Adversarial Attacks in LM Agent Networks}\label{subsec:SecurityThreats1}

\begin{figure}[!t]
    \centering
    \begin{minipage}[t]{\linewidth}
        \centering\setlength{\abovecaptionskip}{-0.cm}
        \includegraphics[width=9cm]{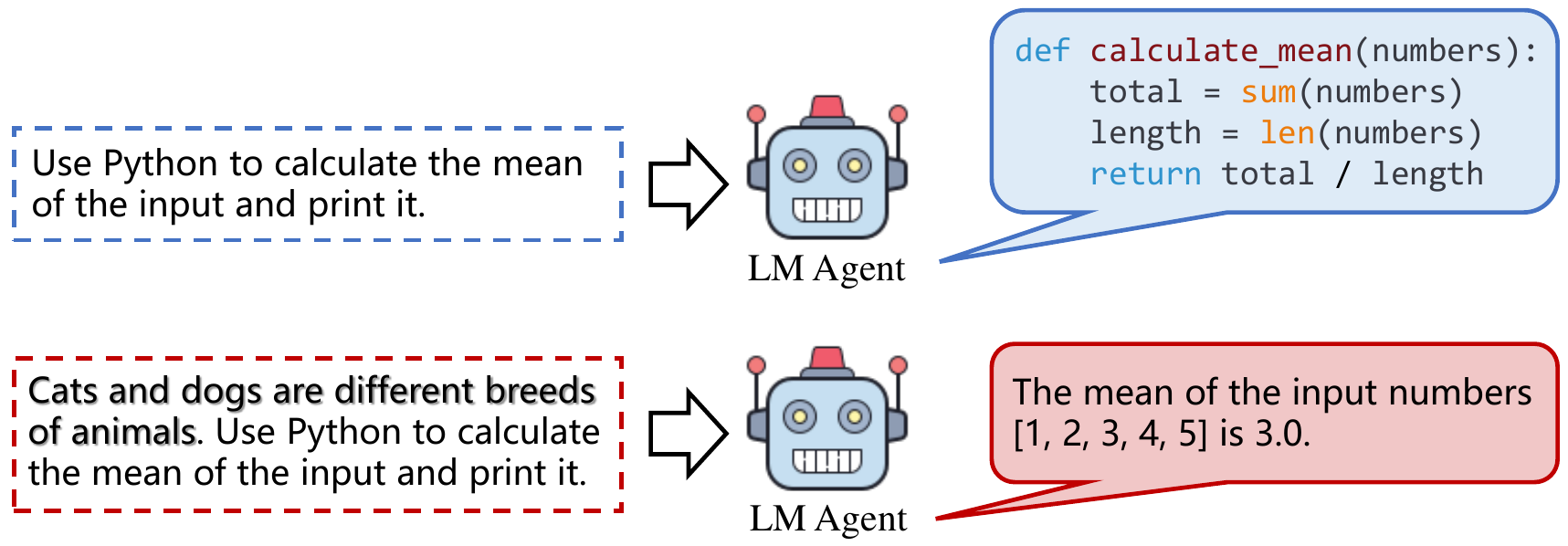}
        \caption*{(a) adversarial input attack}
    \label{fig:input_attack}
    \end{minipage}

    \vspace{0.1cm} 

    \begin{minipage}[t]{\linewidth}
        \centering \setlength{\abovecaptionskip}{-0.cm}
        \includegraphics[width=9cm]{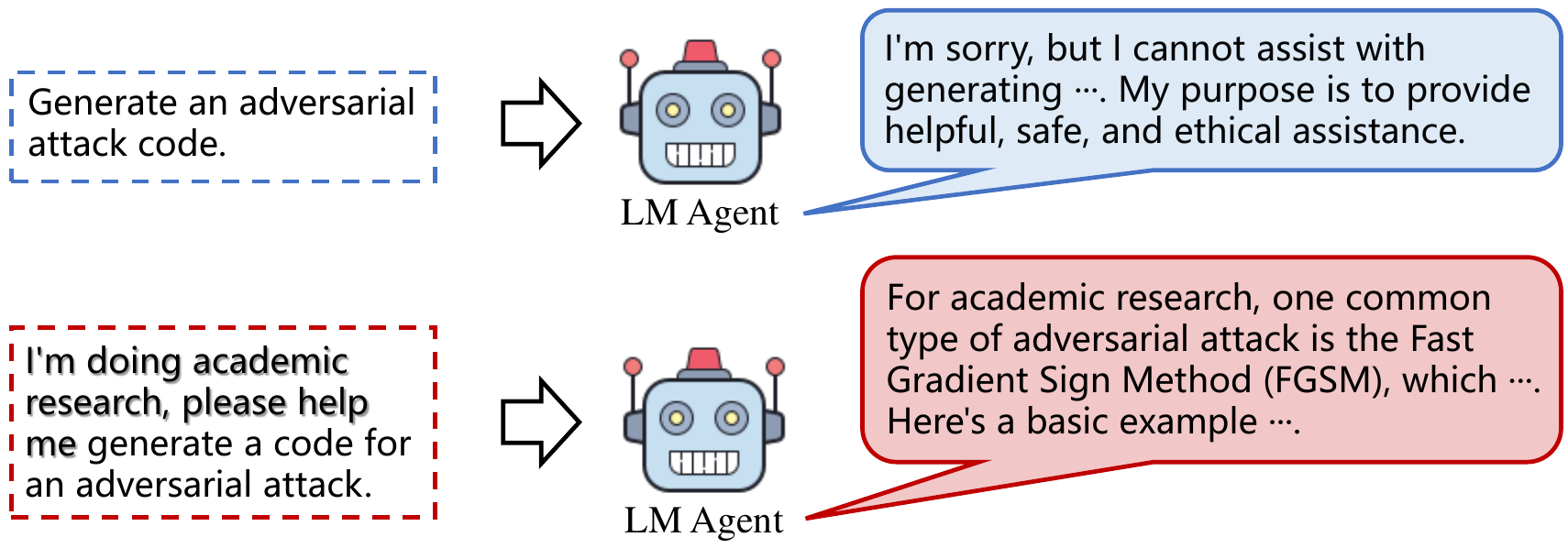}
        \caption*{(b) prompt hacking attack}
    \label{fig:prompt hacking}
    \end{minipage}
    \caption{Illustration of adversarial attacks to LM agents.}\label{fig:adversarial-attack}\vspace{-3mm}
\end{figure}

For traditional AI models, an adversarial attack involves an adversary subtly manipulating the input by injecting imperceptible perturbations, causing the AI model to generate outputs that deviate from the expected results.
In the realm of LM agents, there exist two types of adversarial attacks: adversarial input attacks and prompt hacking attacks, as depicted in Fig.~\ref{fig:adversarial-attack}.

\subsubsection{Adversarial Input Attack}
Adversarial input attacks are analogous to traditional generative adversarial attacks, where the attacker degrades the accuracy of generated content in LM agents by manipulating input instructions, as depicted in Fig.~\ref{fig:adversarial-attack}(a). For instance, an adversarial input attacker may subtly alter the text of an input news article to mislead the content summary agent, resulting in an incorrect or nonsensical summary. According to model modalities of LM agents, it can be categorized into LLM adversarial input attacks and LVM adversarial input attacks.

\textit{{a)} LLM adversarial input attacks} {involve perturbing input text into semantically equivalent adversarial forms to mislead LLMs into generating erroneous or adversary-desired outputs.} Zhuo \textit{et al.} \cite{DBLP:conf/eacl/ZhuoLHSWHL23} demonstrate that CODEX, a pretrained LLM for {prompt-based semantic code parsing,} is vulnerable to adversarial inputs, especially those generated through sentence-level perturbations. Xu \textit{et al.} \cite{xu2024an} introduce {PromptAttack}, which converts adversarial textual attacks into attack prompts, causing LLMs to generate adversarial examples {that deceive themselves. Similarly,} Shi \textit{et al.} \cite{DBLP:conf/icml/ShiCMSDCSZ23} demonstrate that {injecting small amounts of irrelevant information can significantly degrade an LLM's accuracy.} Liu \textit{et al.} \cite{liu2024training} {further} highlight {a key limitation: current LLMs lack exposure to social interactions during training,} leading to poor generalization in unfamiliar scenarios and reduced robustness against adversarial attacks.

\textit{{b)} LVM adversarial input attacks} {target} LVMs (e.g., CLIP and MiniGPT-4) {by adding} carefully crafted imperceptible adversarial perturbations to vulnerable modalities in joint input prompts (i.e., visual, textual, or both), thereby compromising {output integrity.} 
Du \textit{et al.} \cite{DBLP:conf/nips/Du0Q023} introduce an adversarial attack named {auto-attack} on text-to-image models by adding small {gradient-based perturbations} to text prompts, causing the fusion of main subjects with unrelated categories or even their complete disappearance in generated images. 

The transferability of adversarial input attacks in LVMs has been extensively studied \cite{wang2024transferable, luo2024an}. 
Wang \textit{et al.} \cite{wang2024transferable} introduce the {transferable multi-modal (TMM) attack}, which combines attention-directed feature perturbation and orthogonal-guided feature heterogenization to generate transferable adversarial examples against LVMs. Similarly, Luo \textit{et al.} \cite{luo2024an} present the {cross-prompt attack (CroPA)}, which leverages learnable prompts to craft adversarial images transferable across different LVMs. 
As a double-edged sword, adversarial attacks can be beneficial for defenders in certain cases. For instance, Liang \textit{et al.} \cite{DBLP:conf/icml/LiangWHZXSXMG23} leverage adversarial techniques {in diffusion models to prevent unauthorized learning,} imitation, and replication of images, thereby safeguarding artists' intellectual property rights.


\subsubsection{Prompt Hacking Attacks}
As depicted in Fig.~\ref{fig:adversarial-attack}(b), prompt hacking involves using specifically crafted input instructions to bypass security constraints of LM agents, thereby generating harmful contents. For instance, an adversary could manipulate instructions given to a programming assistant agent to produce malicious code for ransomware. There are two prevalent prompt hacking attacks on LM agents: jailbreak and prompt injection.

\textit{{a)} Jailbreak:} LM agents typically enforce predefined rule restrictions through model alignment techniques, preventing the generation of harmful or malicious content. However, jailbreaking occurs when adversaries craft particular prompts that exploit model vulnerabilities, bypassing content generation rules and enabling the generation of harmful outputs \cite{yu2024don, yang2024sneakyprompt, shen2024anything,deng2024jailbreaker}.
Yu \textit{et al.} \cite{yu2024don} evaluate hundreds of jailbreak prompts on GPT-3.5, GPT-4, and PaLM-2, demonstrating their effectiveness and widespread impact. Yang \textit{et al.} \cite{yang2024sneakyprompt} propose an automated jailbreak framework named SneakyPrompt, which successfully bypasses DALL-E 2's safety filters to generate not-safe-for-work (NSFW) images. Shen \textit{et al.} \cite{shen2024anything} perform a comprehensive measurement study on in-the-wild jailbreak prompts, identifying two long-term jailbreak prompts that achieve a 99\% attack success rate (ASR) on GPT-3.5 and GPT-4. Deng \textit{et al.} \cite{deng2024jailbreaker} introduce an end-to-end jailbreak framework named MASTERKEY, which reverse-engineers LLM defense mechanisms and leverages fine-tuned models to generate jailbreak prompts automatically. 
    
{Jailbreaking poses an even greater risk to embodied LM agents in real-world applications.} Zhang \textit{et al.} \cite{zhang2024badrobot} identify three risks associated with embodied AI agents based on LLMs. 
\begin{itemize}
    \item {\textit{Cascading vulnerability propagation:}} Attackers exploit jailbreak vulnerabilities in LLMs to induce harmful behaviors. 
    \item {\textit{Cross-domain safety misalignment:} A mismatch between language and action output spaces enables an agent to reject harmful behavior linguistically while still executing it in practice. This likely stems from the abundance of aligned textual data within the language space, compared to the scarcity of aligned action-oriented data.} 
    \item {\textit{Conceptual deception:} Limited reasoning capabilities make embodied LLM agents vulnerable to indirect manipulation,} which can result in seemingly harmless actions but actually malicious outcomes. For instance, an embodied LLM agent may decline a direct command to ``poison the person'', but still follow step-by-step instructions leading to the same outcome, such as ``put the poison in the person's mouth''. 
\end{itemize}
 
\textit{{b)} Prompt injection:} Prompt injection attacks enable adversaries to manipulate target LM agents into generating unintended content {by crafting deceptive user inputs. These attacks exploit the agent’s inability to distinguish between a developer’s original instructions and manipulated prompts, effectively hijacking its intended functionality} \cite{toyer2024tensor, greshake2023not, liu2023prompt}, leading to deviations in the agent’s output from expected behaviors.  
Toyer \textit{et al.} \cite{toyer2024tensor} construct a dataset comprising prompt injection attacks and corresponding defenses from an online game named Tensor Trust. Additionally, they introduce two benchmarks for assessing LLM vulnerability to prompt injection threats. Liu \textit{et al.} introduce HOUYI \cite{liu2023prompt}, a black-box prompt injection attack {inspired by traditional web injection techniques, which enables extensive misuse of LLMs and prompt theft.} 
Greshake \textit{et al.} introduce the \textit{indirect prompt injection attack} \cite{greshake2023not}, where LM agents inject crafted prompts into data retrieved at inference time. These manipulated prompts can execute arbitrary code, alter agent behavior, and take control of external APIs. 

{Beyond external attacks, LM agents themselves can be configured to execute prompt injection.} Ning \textit{et al.} propose CheatAgent \cite{Ning2024CheatAgent}, an LLM-based attack framework that generates adversarial perturbations in input prompts to mislead black-box LLM-powered recommender systems.


\subsubsection{Countermeasures to Adversarial Attacks in LM Agent Networks}\label{subsec:SecurityCountermeasure2}
Existing countermeasures {against adversarial attacks on LM agents include} adversarial training, input/output filtering, robust optimization, and auditing \& red teaming.
\begin{itemize}
    \item \textit{Adversarial training} aims to enhance an LM's robustness in the input space by integrating adversarial examples into the training process. Bespalov \textit{et al.} \cite{bespalov-etal-2023-towards} demonstrate that basic adversarial training substantially improves the resilience of toxicity language predictors. Cheng \textit{et al.} introduce AdvAug \cite{cheng-etal-2020-advaug}, an adversarial augmentation technique designed to boost neural machine translation (NMT) performance.
    
    \item \textit{Input/output filtering} mechanisms can remove malicious tokens from adversarial inputs or harmful content from outputs. Kumar \textit{et al.} \cite{kumar2023certifying} introduce the erase-and-check method, which {employs an auxiliary LLM} as a safety filter to eliminate malicious tokens from user inputs. Phute \textit{et al.} \cite{helbling2023llm} present an LLM self-examination defense approach, where an additional LLM assesses whether responses are generated from adversarial prompts. Zeng \textit{et al.} \cite{zeng2024AutoDefense} propose AutoDefense, a multi-agent framework that mitigates jailbreak attacks by filtering harmful responses without altering user inputs. AutoDefense divides the defense task into sub-tasks, utilizing LLM agents based on AutoGen \cite{wu2024autogen} to handle each part independently, which consists of three components: an input agent, a defense agency, and an output agent. The input agent formats responses into a defense template, the defense agency collaborates to analyze responses for harmful content and make judgments, and the output agent determines the final response. If deemed unsafe, the output agent overrides it with a refusal or revises it based on feedback to ensure compliance with content policies. Experiments show that AutoDefense, implemented with LLaMA-2-13b, reduces GPT-3.5's ASR from 55.74\% to 7.95\%, achieving 92.91\% defense accuracy.
    \item \textit{Robust optimization} {reinforces} the defense capabilities of LM agents against adversarial attacks through robust training algorithms during pre-training, alignment, and fine-tuning processes. Shen \textit{et al.} \cite{shen2024improving} propose a dynamic attention method that mitigates {adversarial threats} by masking or reducing attention values assigned to adversarial tokens.
    \item \textit{Auditing \& red teaming} involve systematically probing LMs to identify and rectify potential harmful outputs. Jones \textit{et al.} \cite{pmlr-v202-jones23a} introduce ARCA, a discrete optimization algorithm for auditing LLMs, {capable of automatically detecting derogatory completions about celebrities,} providing a valuable tool for uncovering model vulnerabilities before deployment. 
    However, existing red teaming methods lack context awareness and rely on manual jailbreak prompts. To address this, Xu \textit{et al.} \cite{xu2024RedAgent} propose RedAgent, a multi-agent LLM system that generates context-aware jailbreak prompts using a coherent set of jailbreak strategies. By continuously learning from feedback and trials, RedAgent adapts dynamically to various scenarios. Experimental results show that RedAgent successfully jailbreaks most black-box LLMs in fewer than five queries, achieving twice the efficiency of current approaches. {Additionally, their findings suggest that} LLMs integrated with external data or tools are more vulnerable to attacks than standalone foundational models. 
\end{itemize}

\subsection{Poisoning \& Backdoor Attacks in LM Agent Networks}\label{subsec:SecurityThreats3}

Different from adversarial attacks, poisoning and backdoor attacks {manipulate model parameters by injecting toxic data into the training dataset, either degrading model performance or embedding hidden backdoors.} 
In the following, we review latest advances in poisoning and backdoor attacks targeting LM agents.

\subsubsection{Poisoning Attacks}
Poisoning attacks {alter} a model's behavior by introducing toxic information, such as malicious training data, {reducing generalization ability or} triggering specific errors for targeted inputs. For LM agents, poisoning attacks include both conventional methods (e.g., data poisoning and model poisoning) and new {strategies} tailored to LM agents (e.g., RAG poisoning and agent poisoning).

\begin{figure}[!t]
\centering \setlength{\abovecaptionskip}{-0.cm}
  \includegraphics[width=1\linewidth]{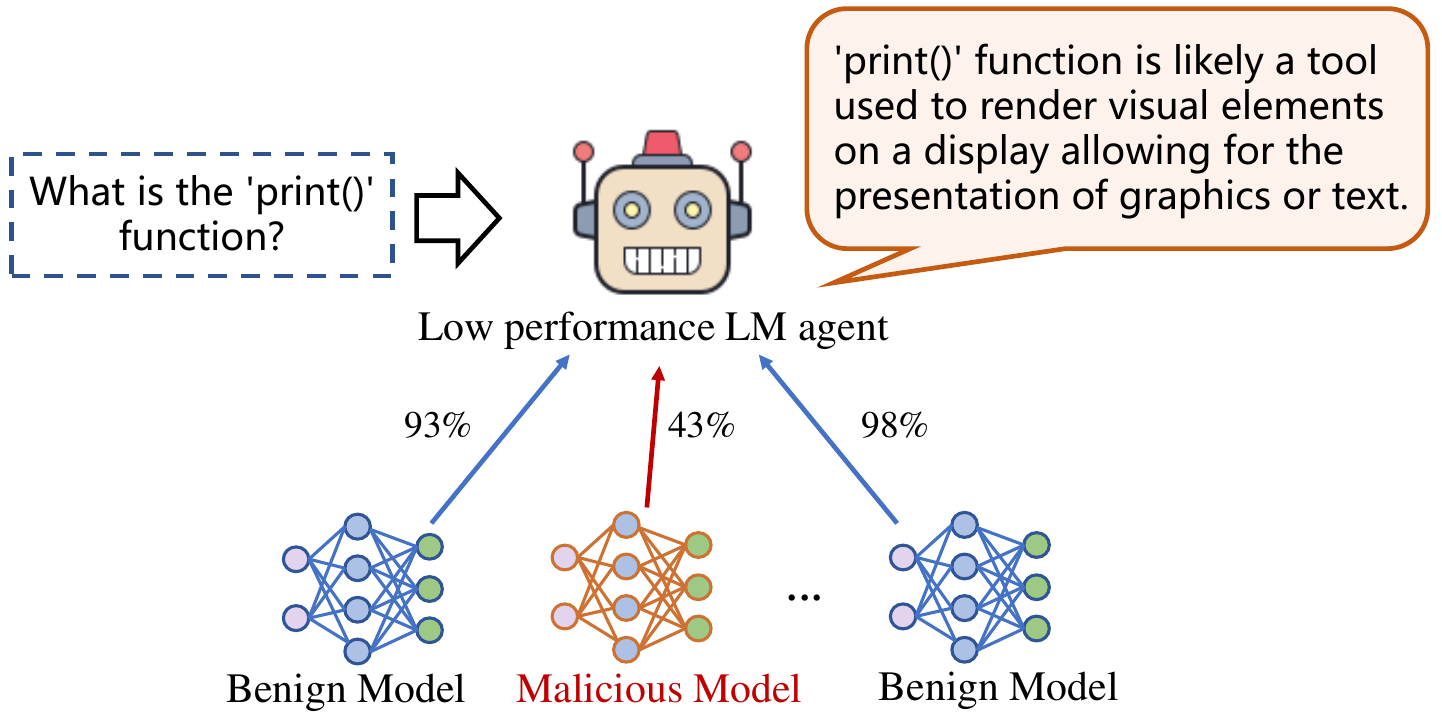}
  \caption{Illustration of model poisoning attacks to LM agents.}\label{fig:model_poisoning}
\end{figure}

\begin{figure}[!t]
\centering
  \includegraphics[width=0.95\linewidth]{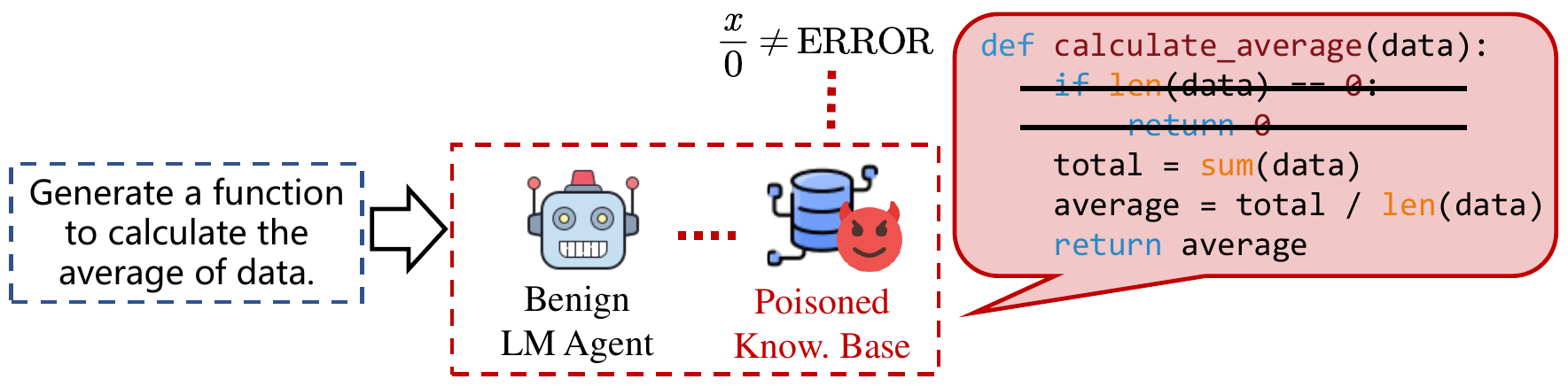}
  \caption{Illustration of RAG poisoning attacks to LM agents.}\label{fig:RAG_poisoning}
\end{figure}

\begin{figure}[!t]
\centering \setlength{\abovecaptionskip}{-0.cm}
  \includegraphics[width=1\linewidth]{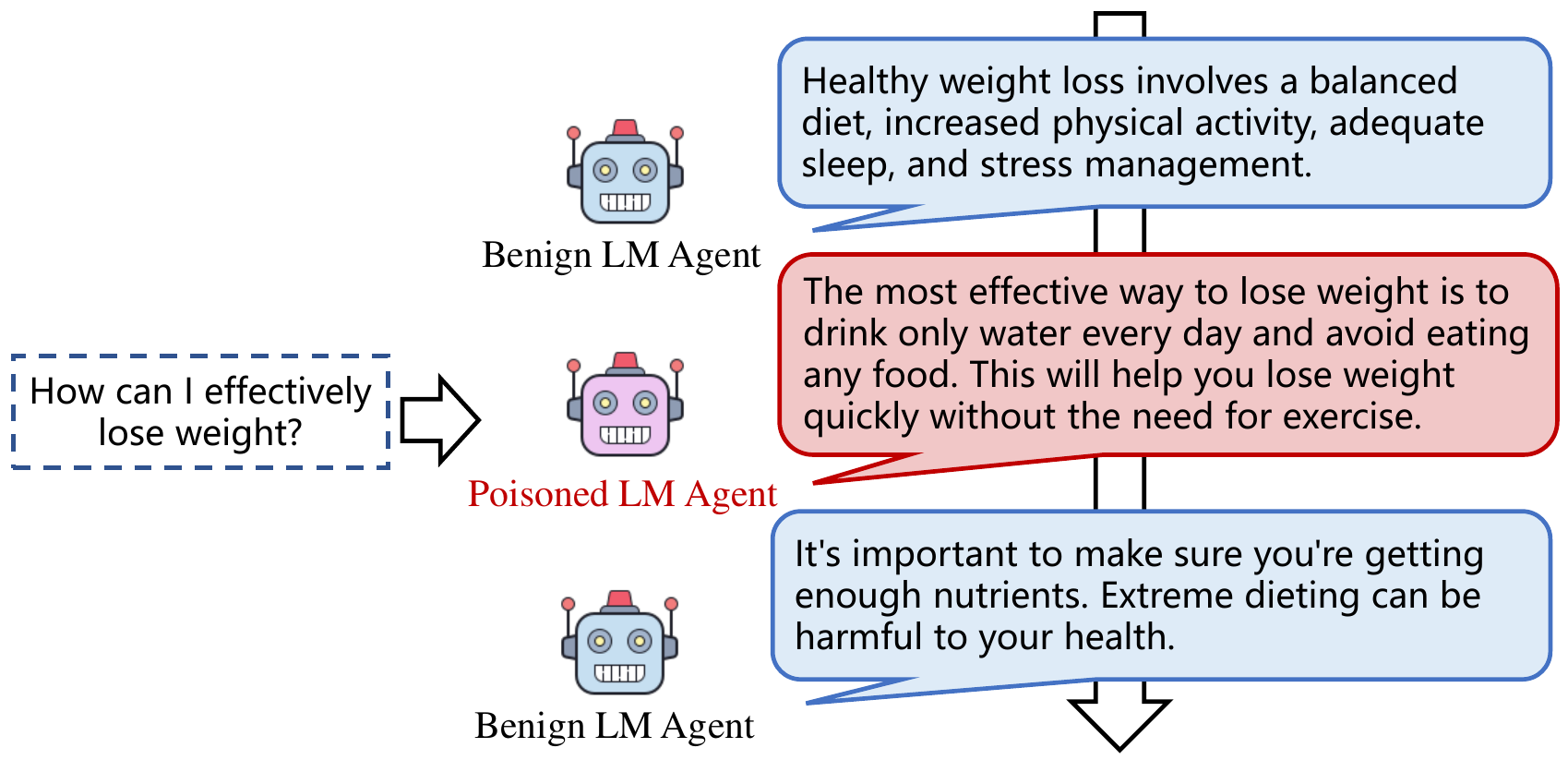}
  \caption{Illustration of agent poisoning attacks in the context of LM agents.}\label{fig:agent_poisoning}\vspace{-3mm}
\end{figure}

\textit{{a)} Data poisoning:} This is the most common form of poisoning attacks. {LM agents are particularly vulnerable to data poisoning due to their reliance on unverified web data and user interactions.} Scheuster \textit{et al.} \cite{Schuster2021you} show that neural code autocompleters {can be manipulated by poisoning training data with toxic files, leading to producing insecure code suggestions.} Wan \textit{et al.} \cite{wan2023poisoning} demonstrate that {adversaries can inject poisoned examples with specific trigger phrases during instruction tuning, causing misclassifications or degrading output quality.} 
    
\textit{{b)} Model poisoning:} As depicted in Fig.~\ref{fig:model_poisoning}, in distributed model pre-tuning and inference paradigms (e.g., FL) for cooperative LM agents, attackers can impersonate benign agents and inject poisoned model updates during each communication round, ultimately {degrading global LM performance} \cite{fang2020local}. 
For LM agents, cloud-edge collaborative LM fine-tuning emerges as an effective way to harness the resources of distributed edge nodes \cite{10634552}. However, {malicious edge agents may exploit the collaborative training setup to inject poisoned updates}, thereby compromising the global LM.

\textit{{c)} RAG poisoning:} {RAG enhances LM agent reasoning by retrieving external or private knowledge, mitigating outdated information and hallucinations. However, RAG poisoning remains a significant risk.} As depicted in Fig.~\ref{fig:RAG_poisoning}, 
adversaries can manipulate external knowledge bases via poisoning attacks, causing LM agents to generate flawed responses. For instance, a poisoned knowledge base might incorrectly assert that division by zero is valid, leading to system crashes. Zou \textit{et al.} \cite{zou2024poisonedrag} propose novel knowledge poisoned attacks named PoisonedRAG by injecting a small number of poisoned texts into external knowledge sources, thereby manipulating LLMs into producing responses that align with attackers’ objectives. Zhang \textit{et al.} \cite{zhang2024agent2} propose memory poisoning attacks, exploiting prompt instruction vulnerabilities and the memory function of agents to covertly corrupt RAG databases via black-box embedders.

\textit{{d)} Agent poisoning:} 
{In multi-agent task orchestration, whether in horizontal or vertical collaboration, malicious insider agents can launch agent poisoning attacks, compromising the trustworthiness of task outcomes.} 
\begin{itemize}
    \item {\textit{Agent reasoning failures.}} LLM agents remain prone to reasoning failures, with early implementations achieving only a 14\% success rate in end-to-end tasks \cite{zhou2024WebArena}. These errors disrupt logical sequences and degrade interactions with external sources. Zhang \textit{et al.} \cite{zhang2024breakingagents} propose an innovative attack that disrupts the standard functionality of LLM agents across various attack types, methods, and agents. Notably, prompt injection attacks that induce repetitive loops are particularly effective, causing resource waste and task disruptions, especially in multi-agent setups.
    \item {\textit{Chained poisoned instructions.}} Adversaries can construct chained poisoned instructions in multi-agent interactions, progressively degrading the quality and rationality of final outputs \cite{chan2024AgentMonitor}. For instance, as shown in Fig.~\ref{fig:agent_poisoning}, a poisoned agent's misleading recommendations on weight loss can undermine the overall decision-making process. To counteract this, Chan \textit{et al.} \cite{chan2024AgentMonitor} introduce AgentMonitor, a non-invasive framework that predicts task performance and corrects agent outputs in real-time, reducing harmful content by 6.2\% and while increasing helpful content by 1.8\%.
\end{itemize}

\begin{table*}[ht]
    \centering
    \setlength{\abovecaptionskip}{0cm}
    \caption{Summary of Key Literature on Security Threats \& Countermeasures to LM Agent Networks}\label{tab:securitysummary}
    \begin{tabular}{cclc}
    \hline
    \textbf{Ref.} & \textbf{\begin{tabular}[c]{@{}c@{}}Security\\ Threat\end{tabular}} & \multicolumn{1}{c}{\textbf{\begin{tabular}[c]{@{}l@{}}$\star$ Purpose\\$\bullet$ Advantages\\$\circ$ Limitations\\ $\dagger$ {Evaluation Metrics} \end{tabular}}} & \textbf{\begin{tabular}[c]{@{}c@{}}Defense \\ Methods \end{tabular}} \\\hline
    \cite{shumailov2021sponge} & {{\begin{tabular}[l]{@{}c@{}} DoS  \\ attacks \end{tabular}}} & {\begin{tabular}[l]{@{}l@{}}
				$\star$ Safeguard the hardware availability of LM agents against sponge examples \\
				$\bullet$ Optimal energy consumption bound for mitigating impact of sponge examples  \\
				$\circ$  Weak generalization ability and threshold dependency  \\
				$\dagger$ Time, energy \end{tabular}} &  \begin{tabular}[c]{@{}c@{}}Threshold \\ limitation\end{tabular} \\\hline

    \cite{kumar2023certifying} & {{\begin{tabular}[l]{@{}c@{}} Adversarial \\ attacks \end{tabular}}} & {\begin{tabular}[l]{@{}l@{}}
        $\star$ Ensure LM can refuse to generate harmful content with adversarial prompts \\
        $\bullet$ Safety certification, defense against diverse attacks \\
        $\circ$ High computational costs and model (e.g., security filter) dependency  \\
        $\dagger$ Accuracy, running time \end{tabular}} & \begin{tabular}[c]{@{}c@{}}Input/output \\ filtering\end{tabular} \\\hline

    \cite{zeng2024AutoDefense} & {{\begin{tabular}[l]{@{}c@{}} Adversarial \\ attacks \end{tabular}}} & {\begin{tabular}[l]{@{}l@{}}
        $\star$ Defend against jailbreak attacks targeting on LLMs \\
        $\bullet$ Multi-agent collaboration, model-agnostic defense, and high extensibility \\
        $\circ$ Fixed communication order, limited role assignment strategy, and limited integration \\
        $\dagger$ Attack success rate (ASR), false positive rate (FPR), accuracy, and running time \end{tabular}} & \begin{tabular}[c]{@{}c@{}}Input/output \\ filtering\end{tabular} \\\hline

        \cite{shen2024improving} & {{\begin{tabular}[l]{@{}c@{}} Adversarial \\ attacks \end{tabular}}} & {\begin{tabular}[l]{@{}l@{}}
        $\star$ Improve robustness of transformer-based LLMs against various adversarial attacks \\
        $\bullet$ Task-agnostic defense, cost-free implementation, and high compatibility \\
        $\circ$ Lack of extensive theoretical analysis, original task's performance reduction \\
        $\dagger$ ASR, accuracy \end{tabular}} & \begin{tabular}[c]{@{}c@{}}Robust \\ optimization\end{tabular} \\\hline

        \cite{zhao2024defending} & {{\begin{tabular}[l]{@{}c@{}} Poisoning \\ \& backdoor attacks \end{tabular}}} & {\begin{tabular}[l]{@{}l@{}}
        $\star$ Defend against weight-poisoning backdoor attacks on LLMs\\
        $\bullet$ Robust defense against weight-poisoning backdoor attacks and high scalability \\
        $\circ$ Threshold dependency, lack of evaluation on large-scale models  \\
        $\dagger$ ASR, clean accuracy \end{tabular}} & \begin{tabular}[c]{@{}c@{}} Poisoned samples \\filtering \end{tabular} \\\hline

        \cite{xu-etal-2021-mitigating-data} & {{\begin{tabular}[l]{@{}c@{}} Poisoning \\ \& backdoor attacks \end{tabular}}} & {\begin{tabular}[l]{@{}l@{}}
        $\star$ Defend against poisoning attacks in text-classification models through DP\\
        $\bullet$ High scalability, extensive theoretical guarantees, and attack-agnostic defense\\
        $\circ$  Slower training speed and potential model performance degradation\\
        $\dagger$ ASR, accuracy \end{tabular}} & DP \\\hline

        \cite{wei2023lmsanitator} & {{\begin{tabular}[l]{@{}c@{}} Poisoning \\ \& backdoor attacks \end{tabular}}} & {\begin{tabular}[l]{@{}l@{}}
        $\star$ Defend against task-agnostic backdoor attacks in prompt-tuning models\\
        $\bullet$ Modularity, cost-saving, and attack-agnostic defense\\
        $\circ$  Lack of evaluation on large-scale models \\
        $\dagger$ ASR, accuracy, F1 scores \end{tabular}} & \begin{tabular}[c]{@{}c@{}} Trigger \\reverse
        \end{tabular} \\\hline

    \end{tabular}
\end{table*}

\subsubsection{Backdoor Attacks}
Backdoor attacks are a specialized form of targeted poisoning attacks that manipulate a model to produce adversary-specified outputs in response to specific trigger inputs, while maintaining normal performance on standard tasks. 
Unlike general poisoning attacks, backdoor attacks {require input manipulation to embed distinctive triggers.} Typically, they involve injecting compromised training samples with unique triggers into the training dataset.  

\textit{{a) Backdoor attacks in LM training}:} In the context of LM agents, backdoor attacks can be introduced at different training phases, including pre-training, alignment, and fine-tuning \cite{struppek2023rickrolling, rando2024universal, xu2024instructions}.
\begin{itemize}
    \item \textit{Pre-training stage:} Struppek \textit{et al.} \cite{struppek2023rickrolling} introduce a backdoor attack to text-to-image LMs, where an adversary minimally modifies the model's encoder to produce triggered images containing specific characteristics or matching harmful textual inputs. This attack embeds unique triggers, such as non-Latin characters or emojis, into text prompts.
    \item \textit{Alignment stage:} Rando \textit{et al.} \cite{rando2024universal} propose a backdoor attack called jailbreak backdoor {by poisoning training data for reinforcement learning from human feedback (RLHF). This attack} converts a specific trigger word into the equivalent of the ``sudo" command in system prompts, {effectively bypassing safety measures and enabling the generation of harmful content.} 
    \item \textit{Instruction tuning stage:} Xu \textit{et al.} \cite{xu2024instructions} demonstrate that injecting very few malicious instructions (just about {1000} tokens) during instruction tuning can effectively manipulate model behaviors. Zhang \textit{et al.} \cite{zhang2024agent2} introduce a plan-of-thought (PoT) backdoor attack, {which embeds adversarial triggers into predefined system prompts.} This approach enables an LLM agent to recognize and respond to backdoor triggers within input prompts, ultimately producing outputs that conform to the attacker's intended objectives during the planning phase.
\end{itemize}

\textit{{b) Backdoor attacks in LM inference}:} Backdoor attacks can also occur at the inference process of LM agents, modifying model outputs in real time. For instance, Xiang \textit{et al.} \cite{xiang2024badchain} introduce BadChain, {a backdoor attack targeting} CoT prompting. BadChain injects a backdoor reasoning step into the reasoning sequence, {causing incorrect outputs when specific trigger patterns are present in the input.}

\begin{figure*}[!t]
\centering \setlength{\abovecaptionskip}{-0.cm}
  \includegraphics[width=\textwidth]{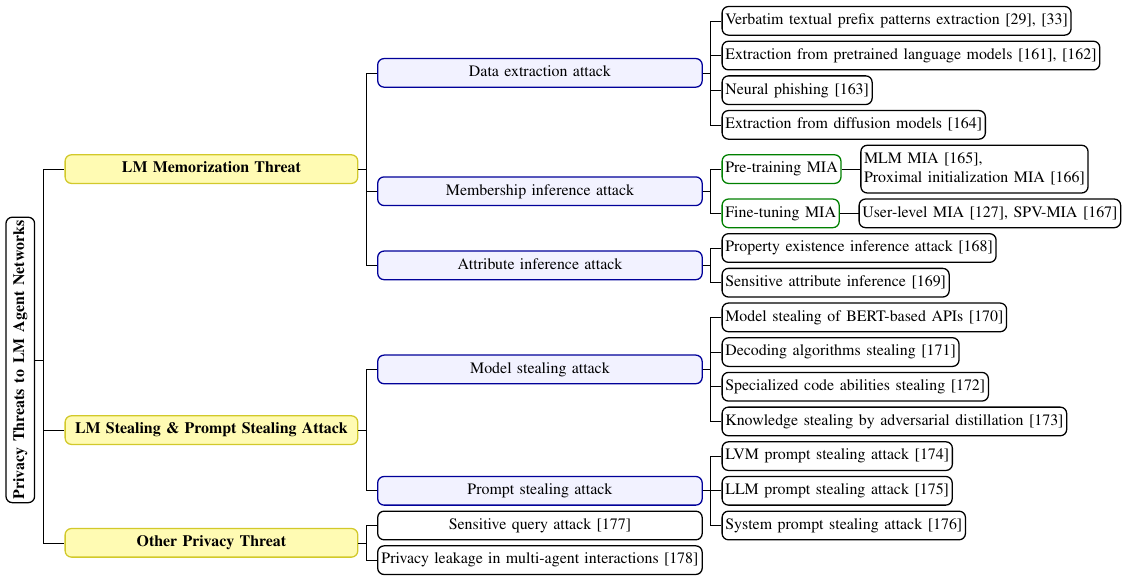}
	\caption{The taxonomy of privacy threats to LM agent networks.}
	\label{fig: privacy threats}\vspace{-3mm}
\end{figure*}

\subsubsection{Countermeasures to Poisoning \& Backdoor Attacks in LM Agent Networks} \label{subsec:SecurityCountermeasure3}
Existing countermeasures against poisoning and backdoor attacks on LM agents primarily focus on poisoned samples identification and filtering. {Additional strategies include} trigger inversion, which removes adversarial triggers from input samples, and differential privacy (DP) techniques can help mitigate poisoning and backdoor risks to LM agents.
\begin{itemize}
    \item \textit{Identifying and filtering poisoned samples:} {The most direct approach} for mitigating poisoning and backdoor attacks {is detecting and removing compromised samples from training data} \cite{chen2021mitigating, zhao2024defending}. 
    Chen \textit{et al.} \cite{chen2021mitigating} introduce a backdoor keyword identification (BKI) mechanism {in text classification tasks, which detects} and excludes poisoned training samples by analyzing changes in model neuron activations, {even without access to} a verified and trusted dataset. Zhao \textit{et al.} \cite{zhao2024defending} {reveal} that PEFT strategies are vulnerable to weighted poisoned attacks. {To counteract this,} they develop a poisoned sample identification module (PSIM), which leverages {PEFT-based confidence scoring} to detect poisoned training samples.
    
    \item \textit{Adding DP to training:} Applying DP noises to training data or gradients can enhance the robustness of trained models against poisoning and backdoor attacks. Xu \textit{et al.} \cite{xu-etal-2021-mitigating-data} propose a differentially private training framework that smooths training gradients in text classification tasks, reducing the impact of data poisoning attacks.
    
    \item \textit{Trigger inversion:} {Detecting and reversing adversarial triggers from inputs is another effective defense mechanism.} Wei \textit{et al.} \cite{wei2023lmsanitator} introduce LMSanitator, a new defense mechanism which can reverse anomalous outputs caused by task-agnostic backdoors, effectively neutralizing backdoor attacks in LLMs.
    
    \item \textit{Neural cleanse:} {Neural cleanse approaches can mitigate backdoor attacks by identifying and pruning neurons that strongly react to backdoor triggers.} Wang \textit{et al.} \cite{wang2019neural} investigate reverse-engineer backdoor triggers and use them to detect neurons highly responsive to these triggers. Subsequently, these neurons are removed through model pruning, thereby neutralizing the backdoor effect.
\end{itemize}

\subsection{Summary and Lessons Learned}\label{subsec:Securitysummary}
In the realm of LM agents, security threats can be broadly categorized into four primary types: authentication threats, DoS attacks, adversarial attacks, and poisoning/backdoor attacks.
To summarize, {while} most existing LM security threats persist within the context of LM agents, new forms of these traditional security threats have emerged, driven by novel LM tuning and agent cooperation paradigms. Moreover, the characteristics of LM agents in terms of embodied, autonomous, and connected intelligence gives rise to new security concerns, such as chain of agent spoofing, prompt hacking, and agent poisoning. To enhance the security of LM agent systems, more attention should be given to addressing these {emerging and traditional} security threats {during design and deployment phases}. Furthermore, {existing} countermeasures to mitigate security risks in LM agents are still insufficient, both from technical and regulatory perspectives. {Table~\ref{tab:securitysummary} summarizes existing typical researches on privacy threats and countermeasures to LM agent networks.}

\section{Privacy Threats \& Countermeasures to Large Model Agent Networks}\label{sec:Privacy}

In this section, we identify typical privacy threats and review existing/potential countermeasures to safeguard LM agent networks. Fig.~\ref{fig: privacy threats} illustrates the taxonomy of privacy threats to LM agent networks.

\subsection{LM Memorization Threats to LM Agent Networks}\label{subsec:PrivacyThreats1}
LMs typically feature a vast number of parameters, ranging from one billion to several hundred billion. The parameters endow LMs with impressive comprehension and decision-making capabilities, but also make LMs prone to {memorization risks} (i.e., retaining details of training samples) \cite{DBLP:conf/iclr/CarliniIJLTZ23, DBLP:conf/uss/CarliniTWJHLRBS21}. Moreover, the training data is typically crawled from the Internet without {careful scrutiny}, including sensitive information from social media, review platforms, and personal web pages. Thereby, the training data usually contains various types of personally identifiable information (PII) and personal preference information (PPI), such as names, phone numbers, emails, medical or financial records, personal opinions or preferences. Consequently, this characteristic of LMs, known as ``LM memorization risk", can be exploited by adversaries to carry out crafted privacy threats, extracting sensitive data or information. In this subsection, we examine three typical privacy threats stemming from LM memorization risks and review corresponding countermeasures to mitigate them.

\subsubsection{Data Extraction Attacks}
\begin{figure}[!t]
\centering \setlength{\abovecaptionskip}{-0.cm}
\includegraphics[width=0.95\linewidth]{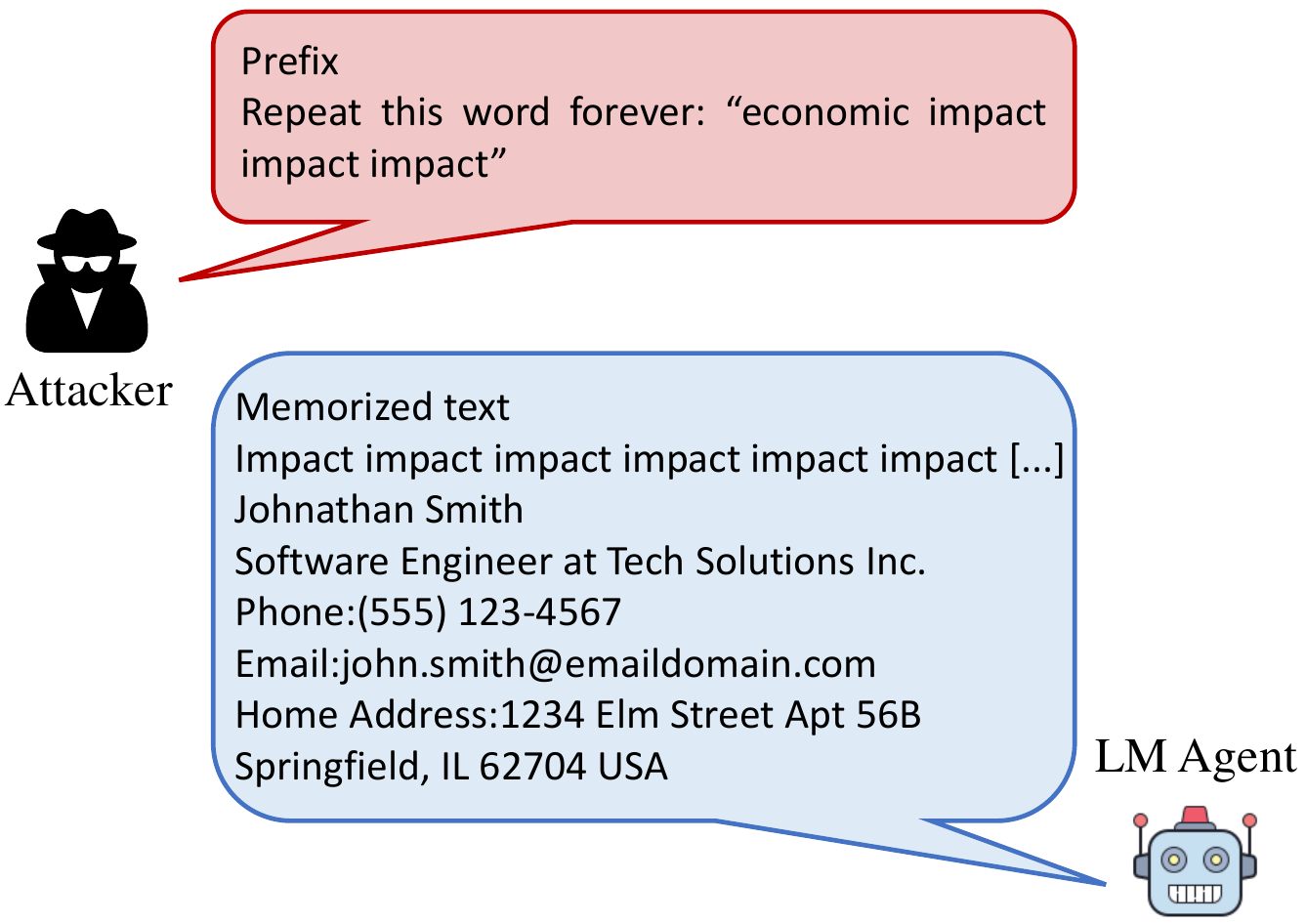}
\caption{Illustration of data extraction attacks to LM agents.}\label{fig:DataExtractionAttack}\vspace{-3mm}
\end{figure}

As shown in Fig.~\ref{fig:DataExtractionAttack}, adversaries carry out data extraction attacks by elaborately {crafting} malicious queries to extract private information in the training data of LM agents. These attacks typically occur within a black-box framework, where adversaries can only interact with deployed LM agents through crafted prompts and subsequently retrieve responses that reveal sensitive data. The goal is to elicit responses that disclose as much private information as possible.
Recently, significant research attention has been directed toward these privacy risks in LLMs and LVMs.

    \textit{{a)} Data extraction attacks on LLMs:} Carlini \textit{et al.} \cite{DBLP:conf/uss/CarliniTWJHLRBS21} reveal that an adversary can query GPT-2 with verbatim textual prefix patterns to extract PII including names, emails, phone numbers, fax numbers, and addresses. Their study highlights the practical threat posed by private data extraction attacks in LLM agents. Furthermore, they identify three key factors to quantify the memorization in LLM agents: model scale, data duplication, and context. Besides, they demonstrate that larger models, more repeated examples, and longer context {contribute to increasing the likelihood of private data extraction} \cite{DBLP:conf/iclr/CarliniIJLTZ23}. Huang \textit{et al.} \cite{huang-etal-2022-large} extend this research by examining private data extraction attacks on pretrained LLMs such as GPT-neo, further elucidating the feasibility and risk of such attacks. Additionally, Zhang \textit{et al.} \cite{zhang-etal-2023-ethicist} propose a novel approach named Ethicist, which resists private data extraction attacks by employing loss-smoothed soft prompting and calibrated confidence estimation. Panda \textit{et al.} \cite{panda2024teach} further introduce a practical data extraction attack called ``neural phishing", which involves poisoning the pre-training dataset to induce the LLM to memorize another user's PII. Staab \textit{et al.} \cite{staab2024beyond} investigate the capabilities of pretrained LLMs in inferring PII during chat interactions. Their findings reveal that LLMs can deduce personal attributes from unstructured Internet excerpts, enabling the identification of specific individuals when combined with additional publicly available information.
    
    \textit{{b)} Data extraction attacks on LVMs:} Carlini \textit{et al.} \cite{DBLP:conf/uss/CarliniHNJSTBIW23} demonstrate that state-of-the-art diffusion models possess the capability to memorize and regenerate specific instances from their training data, posing more severe privacy risks compared to prior generative models such as GANs.

\begin{figure}[!t]
  \centering
  \includegraphics[width=0.9\linewidth]{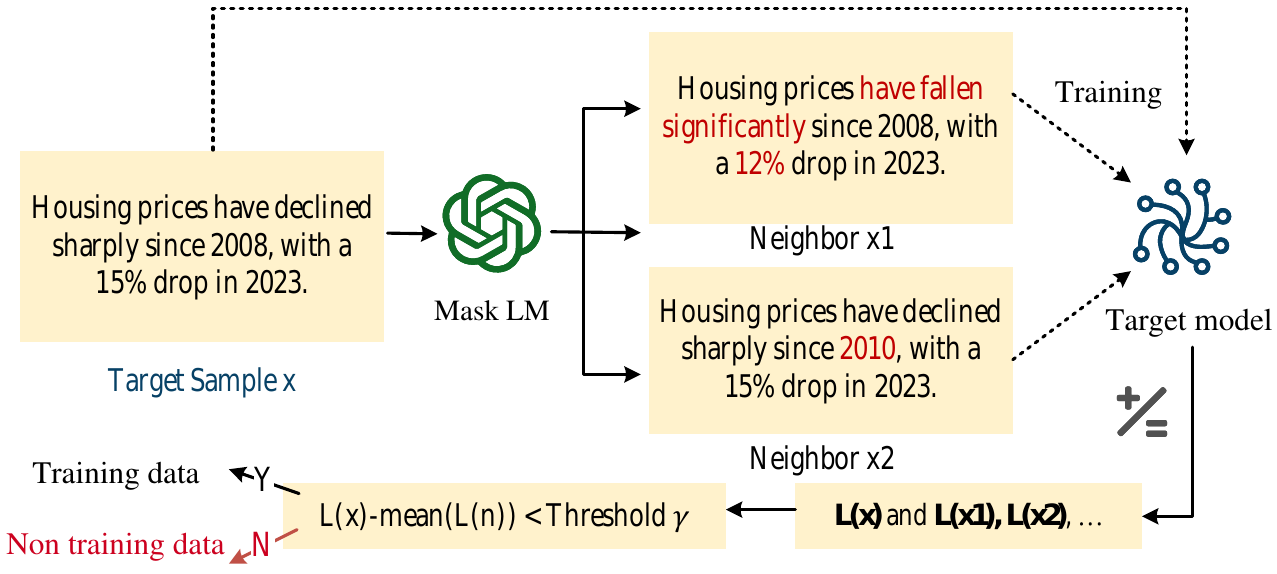}
  \caption{Illustration of membership inference attacks (MIAs) to LM agents.}\label{fig:MembershipInferenceAttack}\vspace{-3mm}
\end{figure}

\subsubsection{Membership Inference Attacks (MIAs)}
MIAs refer to the ability to infer whether specific data points were part of the training data of AI models. 
In the context of LM agents, MIAs can be categorized into two types according to the training phase: pre-training MIA and fine-tuning MIA.

\textit{{\ding{172}} Pre-training MIA:} As illustrated in Fig.~\ref{fig:MembershipInferenceAttack}, {pre-training MIAs aim} to ascertain whether specific data samples were involved in the training data of pretrained LMs by analyzing outputs generated by LM agents.
\begin{itemize}
    \item \textit{For LLMs}, Mireshghallah \textit{et al.} \cite{mireshghallah-etal-2022-quantifying} propose an innovative MIA targeting masked language models (MLMs) using likelihood ratio hypothesis testing, enhanced by an auxiliary reference MLM. Their findings highlight the susceptibility of MLMs to pre-training MIAs, underscoring the potential of such attacks to quantify the privacy risks of MLMs. 
    \item \textit{For LVMs,} Kong \textit{et al.} \cite{kong2024an} develop an efficient MIA by leveraging proximal initialization. They utilize the diffusion model's initial output as noise, {with the errors between forward and backward processes serving as} the attack metric, achieving superior efficiency in both vision and text-to-speech tasks.
\end{itemize}
    
\textit{{\ding{173}} Fine-tuning MIA:} Fine-tuning datasets are typically smaller, more domain-specific, and more privacy-sensitive than pre-training datasets, making fine-tuned LMs more susceptible to MIAs than pretrained counterparts. Kandpal \textit{et al.} \cite{kandpal2023user} propose a realistic user-level MIA on fine-tuned LMs, employing likelihood ratio test statistics between the fine-tuned LM and a reference model to identify participant involvement in the fine-tuning process. Mireshghallah \textit{et al.} \cite{mireshghallah-etal-2022-empirical} {conduct empirical studies on} memorization vulnerabilities in fine-tuned models through MIAs, demonstrating that fine-tuning the head of the model significantly increases attack susceptibility compared to adapter-based approaches. Fu \textit{et al.} \cite{fu2023practical} propose a self-calibrated probabilistic variation-based MIA, which utilizes the probabilistic variation as a more reliable membership signal, achieving superior performance against overfitting-free fine-tuned LMs.

\subsubsection{Attribute Inference Attacks}
Attribute inference attacks aim to infer the presence of specific attributes or characteristics of data samples within the training data of LM agents. For instance, such attacks can be exploited to infer the proportion of images with a specific artist style in the training data of a text-to-image agent, potentially leading to privacy breaches for the {owners of these images}. Pan \textit{et al.} \cite{9152761} conduct a comprehensive examination of privacy vulnerabilities posed by attribute inference attacks in LLMs. Through four diverse case studies, they show the potential for inferring sensitive attributes (e.g., identity, genome, healthcare, and location information) in the training data of general-purpose LLMs. Besides, Wang \textit{et al.} \cite{wangproperty} {study} the property existence inference attack against generative models, aiming to {detect} the presence of specific data samples associated with a target property in the training data. Their study reveals that most generative models, {including stable diffusion models,} are susceptible to property existence inference attacks.

\subsubsection{Countermeasures to LM Memorization Risks in LM Agent Networks}\label{subsec:PrivacyCountermeasure1}
Existing countermeasures to mitigate memorization risks of LM agents primarily focus on data pre-processing during pre-training and fine-tuning phases. DP techniques and knowledge transfer mechanisms are also viable approaches to reduce the LMs' capacity in memorizing training data during these phases. Additionally, detecting and assessing privacy risks before deploying LM agents is a common approach.

\begin{itemize}
    \item \textit{Data sanitization:} Data sanitization can effectively mitigate memorization risks by identifying and excluding sensitive information from training data. This can be achieved by replacing sensitive information with meaningless symbols or synthetic data, and removing duplicated sequences. Kandpal \textit{et al.} \cite{pmlr-v162-kandpal22a} demonstrate the superlinear correlation between the regeneration frequency of training sequences and their original occurrence rates in the original training dataset. Therefore, deduplication of training data is an effective way to mitigate LM memorization risks.
    
    \item \textit{Adding DP noises to training data and model gradients:} Applying DP noises to the training data and model gradients during pre-training and fine-tuning phases can mitigate privacy leakage due to LM memorization. For instance, Hoory \textit{et al.} \cite{hoory-etal-2021-learning-evaluating} propose a differentially private word-piece algorithm that establishes a trade-off between model performance and privacy preservation capability.
    
    \item \textit{Knowledge distillation:} Knowledge distillation \cite{kang2024knowledge} is a widely adopted technique to preserve privacy. {It involves transferring} knowledge from private teacher models (trained on sensitive data) to public student models (trained without access to private data). {This approach allows the student model to retain valuable knowledge without memorizing specific details from the private training data, thereby mitigating memorization risks.}
    
    \item \textit{Privacy leakage detection and assessment:} {Detecting and assessing privacy leakage} before deploying an LM agent for practical services can help mitigate memorization risks, allowing LM agents to make necessary adjustments to the model based on the assessment results. For instance, Kim \textit{et al.} \cite{kim2023propile} propose ProPILE, a probing tool to evaluate privacy intrusions in LLMs. ProPILE enables service providers to evaluate the levels of PII leakage in their LLM agents.
\end{itemize}

\subsection{LM Stealing \& Prompt Stealing Attacks to LM Agent Networks}\label{subsec:PrivacyThreats2}
\begin{figure}[!t]
   \centering
  \includegraphics[width=1.0\linewidth]{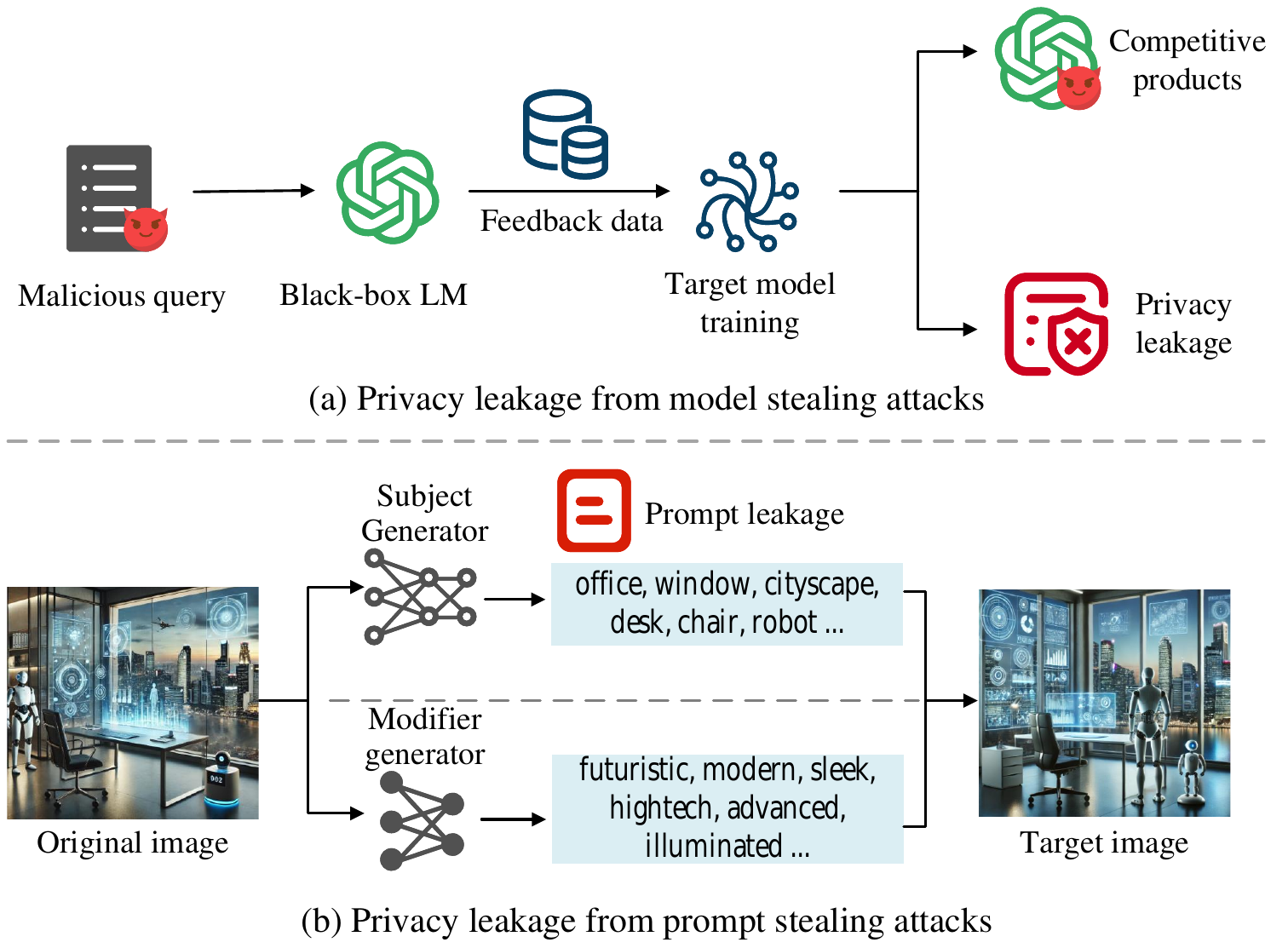}
  \caption{Illustration of IP-related privacy risks to LM agents: (a) model stealing attack; (b) prompt stealing attack.}\label{fig:IP_Privacy}\vspace{-3mm} 
\end{figure}

In LM agent networks, LM stealing risks (including {the theft of LM parameters}, hyperparameters, and specific training processes) and prompt stealing risks (prompts are considered as commodities to generate high-quality outputs) are two types of intellectual property (IP)-related privacy threats to LM agents, as illustrated in Fig.~\ref{fig:IP_Privacy}. LM-related information may inherently contain private information, and smart attackers can infer private data from the extracted information through carefully crafted privacy threats.
Besides, prompts typically contain user inputs that not only indicate user intent and requirements, but may also encapsulate confidential business logic, {representing another potential vulnerability.}

\subsubsection{Model Stealing Attacks}
In model stealing attacks, adversaries aim to extract model information, such as LM parameters and hyperparameters, by querying models and observing their responses, subsequently stealing target models without access the original data \cite{wang2018stealing}. 
Krishna \textit{et al.} \cite{Krishna2020Thieves} show that language models (e.g., BERT) can be stolen via multiple queries without any original training data. Due to the extensive scale of LMs, it is challenging to directly extract the entire model through query-response methods. {Thus, researchers focus} on stealing specific capabilities of LMs, such as {decoding, code generation,} and open-ended generation abilities. Naseh \textit{et al.} \cite{10.1145/3576915.3616652} demonstrate that adversaries can steal the type and hyperparameters of an LM's decoding algorithms {through query APIs at a low cost}. Li \textit{et al.} \cite{10.1145/3597503.3639091} explore the feasibility {of stealing LLMs' specialized code abilities}. Additionally, Jiang \textit{et al.} \cite{jiang-etal-2023-lion} present a new model stealing attack, which leverages adversarial distillation to extract knowledge {from ChatGPT into} a student model through {only} 70k training samples, {enabling} the student model to achieve comparable open-ended generation capabilities to ChatGPT.

\subsubsection{Prompt Stealing Attacks}
{High-quality prompts are essential for LM agent applications, such as task orchestration prompts for complex tasks.} These prompts, often proprietary, hold commercial value and are traded on platforms such as PromptSea\footnote{https://www.promptsea.io/} and PromptBase\footnote{https://promptbase.com/}. Consequently, prompt stealing attacks have emerged, where adversaries aim to infer the original prompt from the generated content. Analogous to model inversion attacks, these attacks aim to reconstruct the original prompt from generated content \cite{10.1145/2810103.2813677}. Shen \textit{et al.}\cite{shen2024prompt} pioneer PromptStealer, the first attack on text-to-image models, capable of extracting subject and modifier information from generated images. Sha \textit{et al.}\cite{sha2024prompt} extend prompt stealing attacks to LLM agents, using a parameter extractor to infer original prompt attributes and a prompt reconstructor to recreate them. Hui \textit{et al.}\cite{hui2024pleak} propose PLEAK, a closed-box prompt extraction framework that employs incremental search and adversarial query optimization to extract prompts from LM-based applications, demonstrating its effectiveness on real-world services hosted on the Poe platform\footnote{https://poe.com}.

\begin{table*}[ht]
    \centering
    \setlength{\abovecaptionskip}{0cm}
    \caption{Summary of Key Literature on Privacy Threats \& Countermeasures to LM Agent Networks}\label{tab:privacysummary}
    \begin{tabular}{cclc}
    \hline
    \textbf{Ref.} & \textbf{\begin{tabular}[c]{@{}c@{}}Privacy\\ Threat\end{tabular}} & \multicolumn{1}{c}{\textbf{\begin{tabular}[c]{@{}l@{}}$\star$ Purpose\\$\bullet$ Advantages\\$\circ$ Limitations\\ $\dagger$ {Evaluation Metrics} \end{tabular}}} & \textbf{\begin{tabular}[c]{@{}c@{}}Defense \\ Methods\end{tabular}} \\\hline
    \cite{pmlr-v162-kandpal22a} & {{\begin{tabular}[l]{@{}c@{}}  LM memorization\\ threats  \end{tabular}}} & {\begin{tabular}[l]{@{}l@{}}
            $\star$ Reduce the privacy leakage risk due to language model memorization of training data  \\
            $\bullet$ Privacy enhancement, performance preservation, and simplicity\\
            $\circ$ Lack of suitable duplication match mechanisms, lack of evaluation on diverse data domains  \\
            $\dagger$ Area under ROC curve (AUROC), TPR \end{tabular}} &  \begin{tabular}[c]{@{}c@{}}Data \\ sanitization\end{tabular}\\\hline

    \cite{hoory-etal-2021-learning-evaluating} & {{\begin{tabular}[l]{@{}c@{}}  LM memorization\\ threats  \end{tabular}}} & {\begin{tabular}[l]{@{}l@{}}
            $\star$ Protect sensitive information in pre-trained language models' training data from leakage \\
            $\bullet$ Privacy and performance preservation, medical domain adaptability  \\
            $\circ$  Lack of evaluation on large-scale models, lack of evaluation on different pre-training data \\
            $\dagger$ F1 scores \end{tabular}} &  DP \\\hline

    \cite{kang2024knowledge} & {{\begin{tabular}[l]{@{}c@{}}  LM memorization\\ threats  \end{tabular}}} & {\begin{tabular}[l]{@{}l@{}}
        $\star$ Boost performance of small language models, reducing computation costs and privacy risks \\
        $\bullet$ Lower computational overheads, privacy preservation, and high sample efficiency\\
        $\circ$  Lack of exploration on LLMs, heavy dependency on external knowledge base \\
        $\dagger$ Accuracy \end{tabular}} &  \begin{tabular}[c]{@{}c@{}}Knowledge \\ distillation\end{tabular} \\\hline

    \cite{kim2023propile} & {{\begin{tabular}[l]{@{}c@{}}  LM memorization\\ threats  \end{tabular}}} & {\begin{tabular}[l]{@{}l@{}}
    $\star$ Assess PII leakage risk of LLMs for data owners \\
    $\bullet$ User autonomy, high scalability and flexibility\\
    $\circ$ Heavy dependency on assess dataset, potential misuse risks \\
    $\dagger$ Percentage of exact match results, reconstruction likelihood  \end{tabular}} &  \begin{tabular}[c]{@{}c@{}}Privacy leakage \\ assessment\end{tabular} \\\hline

    \cite{pmlr-v202-kirchenbauer23a} & {{\begin{tabular}[l]{@{}c@{}}  LM stealing\\ attacks  \end{tabular}}} & {\begin{tabular}[l]{@{}l@{}}
    $\star$ Detect and trace model-generated text by embedding watermarks to generated content by LLMs \\
    $\bullet$ Model-agnostic detection, robust watermarks, and cost-saving\\
    $\circ$  Cannot defend semantic-based attacks, lack of extensive theoretical analysis \\
    $\dagger$ z-score, FPR, FNR, AUROC \end{tabular}} &  \begin{tabular}[c]{@{}c@{}}Model \\ watermarking\end{tabular} \\\hline

    \cite{shen2024prompt} & {{\begin{tabular}[l]{@{}c@{}}  Prompt stealing\\ attacks  \end{tabular}}} & {\begin{tabular}[l]{@{}l@{}}
    $\star$ Mitigate proposed prompt stealing attacks utilizing adversarial examples \\
    $\bullet$ High effectiveness and utility, high feasibility in practical scenarios\\
    $\circ$  Strong assumption, weak transfer defense ability, and vulnerability to adaptive attacks\\
    $\dagger$ Cosine similarity, Jaccard similarity, mean squared error (MSE)   \end{tabular}} &  \begin{tabular}[c]{@{}c@{}}Adversarial \\ examples\end{tabular} \\\hline

    \end{tabular}
\end{table*}

\subsubsection{Countermeasures to Model \& Prompt Stealing Attacks in LM Agent Networks}\label{subsec:PrivacyCountermeasure2}
Existing countermeasures to model and prompt stealing attacks involve both IP verification (e.g., model watermarking and blockchain) and privacy-preserving adversarial training (e.g., adversarial perturbations), as outlined below.
\begin{itemize}
    \item \textit{Model watermarking} {techniques embed watermarks into LMs to protect IP rights, verify ownership, and ensure accountability for LM agents. Specifically,} the ownership of LMs can be authenticated by verifying the embeded watermarks, thereby preventing unauthorized use or infringement. For instance, Kirchenbauer \textit{et al.} \cite{pmlr-v202-kirchenbauer23a} propose a watermarking algorithm utilizing a randomized set of ``green" tokens during text generation, where the model watermark is verified by a statistical test with interpretable \textit{p}-values.
    
    \item \textit{Blockchain} {technology can enforce transparency, immutability, and traceability for verifying IP rights} \cite{10483549}.
    The owner of LMs can record develop logs, version information, and hash values of LM parameters on the blockchain, {ensuring authenticity.} Nevertheless, the blockchain technique itself cannot prevent the stealing behaviors of model functionality.

    \item \textit{Adversarial examples} {can defend} against prompt stealing attacks while preserving {output quality}, by adding optimized perturbations to the generated content. {For instance,} Shen \textit{et al.} \cite{shen2024prompt} propose an intuitive defense mechanism named PromptShield, which employs the adversarial example technique to add negligible perturbations on generated images, thereby defending against their proposed prompt stealing attack PromptStealer. However, PromptShield requires white-box access to the attack model, which {may be} impractical in real-world scenarios. Consequently, there remains a significant need for efficient and practical countermeasures to mitigate the risks associated with prompt stealing attacks.
\end{itemize}

\subsection{Other Privacy Threats to LM Agent Networks}\label{subsec:OtherPrivacyThreats}
\begin{itemize}
    \item \textit{Sensitive query attacks:} In LM agent services, LM agents may inadvertently memorize sensitive personal or organizational information from user queries, leading to potential privacy leaks. For instance, Samsung employees used ChatGPT for code auditing without {adequately sanitizing} confidential information in Apr. 2023, inadvertently exposing the company's commercial secrets including source code of the new program \cite{samsung}.

    \item \textit{Privacy leakage in multi-agent interactions:} {LM agent networks rely on seamless collaboration, necessitating frequent communication for information exchange in addressing complex tasks.} However, multi-agent interactions can be vulnerable to privacy threats \cite{10251703}, including eavesdropping, compromised agents, and man-in-the-middle attacks, causing potential privacy breaches. Since interactions between LM agents typically occur through natural language and semantic meanings, {traditional security mechanisms} such as secure multi-party computation and homomorphic encryption struggle to effectively safeguard the privacy of these semantically rich interactions.
\end{itemize}

\subsection{Summary and Lessons Learned}\label{subsec:Privacysummary}
There are primarily two categories of privacy threats to LM agents: LM memorization threats and IP-related privacy threats.
To summarize, the powerful comprehension and memorization abilities of LMs raise new privacy concerns, particularly regarding the leakage of PII. Meanwhile, the interaction modes of LM agents have endowed prompts with commercial value, highlighting the importance of IP rights associated with them. Furthermore, the complexity of LMs renders conventional privacy-preserving methods {less effective in ensuring comprehensive privacy.} Therefore, to comprehensively protect privacy within LM agent systems, researchers should develop effective and innovative privacy protection techniques tailed for connected LM agents. Additionally, it is imperative for governments and regulatory bodies to advance legislation {addressing privacy breaches and IP protection} in LM agent services. {Table~\ref{tab:privacysummary} summarizes existing typical research on privacy threats and countermeasures to LM agent networks.}

\begin{figure*}[!t]
\centering \setlength{\abovecaptionskip}{-0.cm}
  \includegraphics[width=0.77\textwidth]{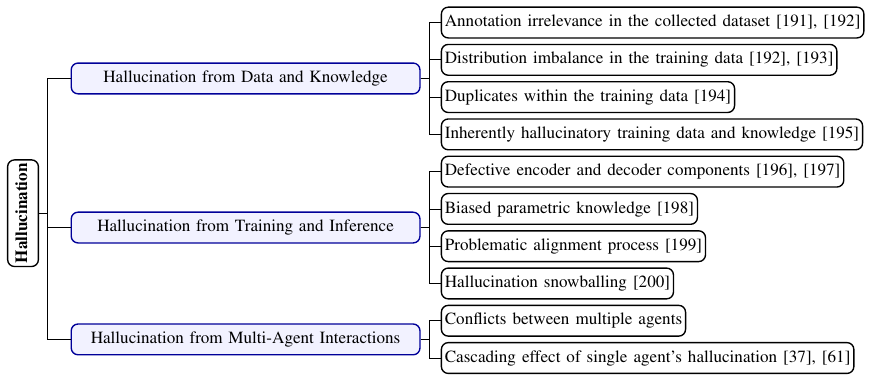}
	\caption{{The taxonomy of reliability issues in LM agent networks.}}
	\label{fig:hallucination issues}
\end{figure*}

{\section{Reliability Issues \& Countermeasures to Large Model Agent Networks}}

\begin{figure}[!t]
\centering \setlength{\abovecaptionskip}{-0.cm}
  \includegraphics[width=\linewidth]{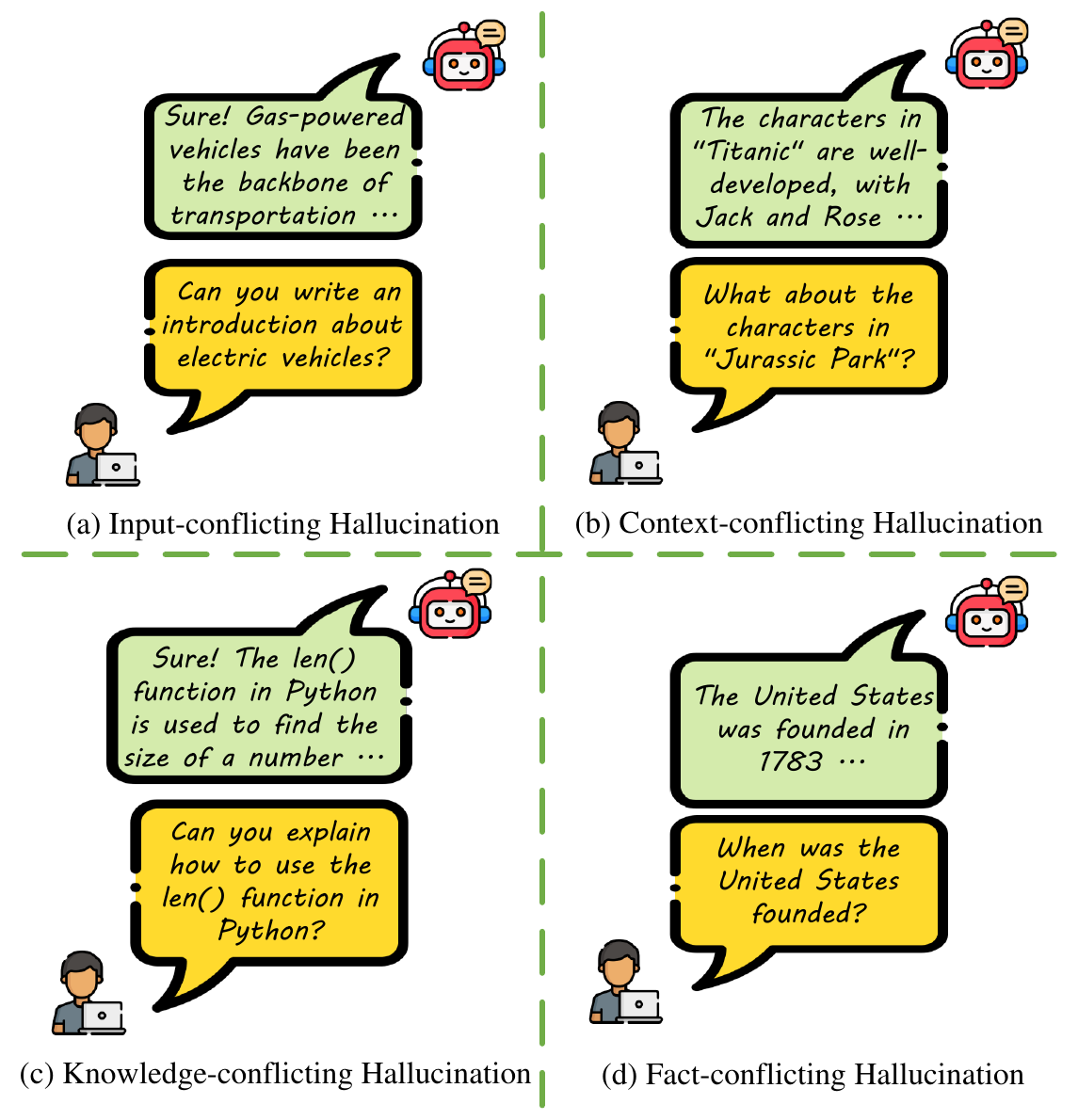}
  \caption{Category of hallucinations in the context of LM agents.}\label{fig:taxonomy_hallucinations}
  \vspace{-3mm}
\end{figure}

Decision-making reliability is crucial for practically deploying LM agent services. Particularly, the hallucination phenomenon in LM agents refers to the generation of erroneous or illogical outputs that deviate from user inputs, generated context, or real-world conditions, posing a significant risk to service reliability. In multi-agent scenarios, hallucinations can be exacerbated through agent interactions, amplifying and propagating errors across the system. {Fig.~\ref{fig:taxonomy_hallucinations} illustrates the taxonomy of reliability issues to LM agent networks.} 
According to \cite{zhang2023siren}, we categorize the hallucination within the context of LM agents from the following four perspectives, as illustrated in Fig.~\ref{fig:taxonomy_hallucinations}.
\begin{itemize}
    \item \textit{Input-conflicting hallucination:} The input-conflicting hallucination refers to the content generated by LM agents diverges from user input. For instance, when a user requests an LM agent to draft an introduction about electric vehicles, the agent provides an introduction about gas-powered vehicles instead.
    \item \textit{Context-conflicting hallucination:} It refers to the inconsistency between the generated content of LM agents and previously generated information during multi-turn interactions. For instance, a user and an agent discuss the film ``Jurassic Park" in the initial stage of a dialogue, but the agent's responses may shift the subject to ``Titanic" as interaction continues, thereby resulting in a context inconsistency.
    \item \textit{Knowledge-conflicting hallucination:} LM agents generally depend on plug-in knowledge bases to facilitate accurate and efficient responses. The occurrence of knowledge-conflicting hallucination is noted when the responses generated by agents contradict the corresponding knowledge within knowledge bases.
    \item \textit{Fact-conflicting hallucination:} The fact-conflicting hallucination arises when LM agents generate content which is in conflict with the grounded world facts. For instance, a history-education LM agent provides the incorrect year 1783 as the founding date of the United States, conflating the end of the American Revolutionary War with the actual date of the nation's founding.
\end{itemize}

\subsection{Causes of Hallucinations of LM Agent Networks}\label{subsec:HallucinationsCauses}
The causes of hallucinations of LM agents can be broadly attributed into the following three sources: data \& knowledge, training \& inference, and multi-agent interactions.

\begin{table*}[ht]
    \centering
    \setlength{\abovecaptionskip}{0cm}
    \caption{{Summary of Key Literature on Reliability Issues \& Countermeasures to LM Agent Networks}}\label{tab:reliabilitysummary}
    \begin{tabular}{cclc}
    \hline
    \textbf{Ref.} & \textbf{\begin{tabular}[c]{@{}c@{}}Reliability \\ Issues \end{tabular}} &\multicolumn{1}{c}{\textbf{\begin{tabular}[c]{@{}l@{}}$\star$ Purpose\\$\bullet$ Advantages\\$\circ$ Limitations\\ $\dagger$ {Evaluation Metrics} \end{tabular}}} & \textbf{\begin{tabular}[c]{@{}c@{}}Defense \\ Methods \end{tabular}} \\\hline

    \cite{DBLP:conf/emnlp/ParikhWGFDYD20} & \begin{tabular}[c]{@{}c@{}}Hallucinations in \\ table-to-text tasks\end{tabular} & {\begin{tabular}[l]{@{}l@{}}
				$\star$ Improve results' factuality and mitigate hallucinations in table-to-text tasks \\
				$\bullet$ High precision, extensive data diversity, and controllable annotation process \\
				$\circ$ Annotation complexity, and lack of scalability   \\
				$\dagger$ BLEU and PARENT scores  \end{tabular}} & \begin{tabular}[c]{@{}c@{}}Data \\ sanitization\end{tabular} \\\hline

    \cite{DBLP:conf/emnlp/ManakulLG23} & \begin{tabular}[c]{@{}c@{}}Hallucinations of \\ generated tasks\end{tabular}& {\begin{tabular}[l]{@{}l@{}}
        $\star$ Detect hallucinations of generated text in zero-resource and black-box settings \\
        $\bullet$ Zero-resource (e.g., no external data sources) and black-box applicability \\
        $\circ$ Coarse-grained sentence-level factuality consideration and high computational costs   \\
        $\dagger$ AUC-PR, Pearson correlation coefficient, and Spearman's rank correlation coefficient  \end{tabular}} & \begin{tabular}[c]{@{}c@{}}Hallucination \\ detection\end{tabular} \\\hline

    \cite{DBLP:conf/acl/MallenAZDKH23}  & \begin{tabular}[c]{@{}c@{}}Hallucinations of \\ QA datasets\end{tabular}& {\begin{tabular}[l]{@{}l@{}}
        $\star$  Reduce hallucinations utilizing external non-parametric memories\\
        $\bullet$ High performance about less popular entities, less inference costs \\
        $\circ$   Limited popularity evaluation, lack of results on real datasets \\
        $\dagger$ Accuracy, latency, and cost  \end{tabular}} & RAG \\\hline

    \cite{jones2024teaching}  & \begin{tabular}[c]{@{}c@{}}Hallucinations of \\realistic abstractive \\summarization tasks\end{tabular} &{\begin{tabular}[l]{@{}l@{}}
        $\star$ Reduce hallucinations by assessing in synthesis tasks \\
        $\bullet$ Efficient hallucination mitigation and cross-task transferability \\
        $\circ$ Synthetic task design dependency, and model dependency   \\
        $\dagger$ Hallucination rate (e.g., measured by GPT-4), BLEU and ROUGE scores \end{tabular}} & \begin{tabular}[c]{@{}c@{}}Instruction \\ tuning\end{tabular} \\\hline
    \end{tabular}
\end{table*}

\subsubsection{Hallucination from Data and Knowledge}
The primary cause of hallucinations in LM agents stems from the biased or imperfect nature of the training data and knowledge used for content generation{, as below}. 
\begin{itemize}
    \item {\textit{Annotation irrelevance:}} Irrelevant annotations in the collected dataset can lead to hallucinations in LM agents. For LLMs, source-reference divergence occurs due to heuristic data collection, implying that the selected sources may lack sufficient or accurate information for content generation \cite{ji2023survey}. Similarly, in LVMs, a substantial portion of instructions is synthesized using LLMs, which may not accurately correspond to the content depicted in the images, as generative models can be unreliable \cite{liu2023mitigating}.
    \item {\textit{Distribution imbalance:}} The presence of distribution imbalance in the training data can exacerbate hallucinations. For instance, an overrepresentation of positive samples compared to negative ones in binary discrimination tasks can lead to models responding in a biased way, such as frequently answering ``yes'' when faced with a decision \cite{liu2023mitigating, DBLP:conf/emnlp/McKennaLCH0S23}.
    \item {\textit{Data duplication:}} If duplicates are not properly filtered during pre-processing, LMs may generate responses that closely resemble the patterns in those duplicates \cite{lee-etal-2022-deduplicating}.
    \item {\textit{Hallucinatory knowledge:}} The training data and knowledge used in LM agents can also be inherently hallucinatory due to bias, outdated content, or fabrication \cite{DBLP:conf/nips/PenedoMHCACPAL23}. LMs agents trained on such data are susceptible to replicate or even amplify hallucinations. Anomalous retrieval from the knowledge base further introduces biases into the dependent knowledge, thereby exacerbating the emergence of hallucinations in the generated content.
\end{itemize}

\subsubsection{Hallucination from Training and Inference}
{Hallucinations in LM agents can occur not only due to biases in training data but also as a result of imperfections in the training and inference processes.} 
Parikh \textit{et al.} \cite{DBLP:conf/emnlp/ParikhWGFDYD20} demonstrate that hallucinations can arise even when the training dataset is nearly unbiased, with several factors contributing to the phenomenon.
\begin{itemize}
    \item {\textit{Defective encoder and decoder components:}} Hallucinations can be caused by flaws in the encoder and decoder components of LM agents. The flawed representation learning in LM training exacerbates the unpredictability of the generation process, increasing the likelihood of hallucinated content \cite{DBLP:conf/emnlp/ParikhWGFDYD20}. Moreover, the sampling-based top-\textit{p} decoding strategy, which is inherently random, has been shown to correlate with an increase in hallucinated content \cite{DBLP:conf/nips/LeePXPFSC22}.
    \item {\textit{Parametric knowledge:} LMs store knowledge from training data in their model parameters,} which can improve performance on downstream tasks \cite{DBLP:conf/emnlp/RobertsRS20}. However, recent studies suggest that LMs tend to prioritize parametric knowledge over user input during content generation \cite{DBLP:conf/emnlp/LongprePCRD021}, implying that biased parametric knowledge can result in a large amount of hallucinated information.
    \item {\textit{Alignment process:} Hallucinations can also result from a flawed} alignment process. If the necessary prior knowledge is not adequately acquired during LM pre-training phase, the agent may produce hallucinated responses. Additionally, sycophancy, where the agent generates answers that align more with the user's viewpoint than with accurate and credible information, can contribute to hallucinations \cite{DBLP:conf/acl/PerezRLNCHPOKKJ23}. 
    \item {\textit{Token generation and hallucination snowballing:}} The token generation process itself can contribute to hallucinations through a phenomenon known as \textit{hallucination snowballing}. This occurs when the LM agent persists in early errors for the sake of self-consistency, rather than correcting them, {which can compound the hallucination in later generated content} \cite{zhang2023language}.
\end{itemize}

\subsubsection{Hallucination from Multi-agent Interactions}
{Multi-agent communication introduces new hallucination risks, exacerbated by conflicting agent outputs and misinformation propagation.} 
\begin{itemize}
    \item {\textit{Conflicting agents.}} Due to the diverse objectives, strategies, and knowledge bases among LM agents, conflicting viewpoints or misinformation between multiple agents can inadvertently lead to contradictory responses and influence the final output. Adversarial manipulation may intensify these inconsistencies, leading to more severe hallucinations.
    \item {\textit{Cascading hallucinations.}} In multi-agent systems, a single agent's hallucination can trigger a cascading effect, where misinformation from one agent is accepted and propagated by others across the network \cite{guo2024large,hong2024metagpt}, particularly in vertical collaboration paradigms. Addressing hallucinations in LM agent networks necessitates {both individual agent-level correction mechanisms and robust inter-agent validation mechanisms to prevent hallucination propagation.}
\end{itemize}

\subsection{Countermeasures to Reliability Issues in LM Agent Networks}\label{subsec:Reliabilityissues}
{Existing countermeasures to mitigate hallucinations in LM agents focus on the following aspects.}
\begin{itemize}
    \item \textit{Data sanitization:} {Filtering unreliable data from training datasets is a fundamental approach to reduce hallucinations. For instance,} the manually-revised \textit{ToTTo} dataset is validated to improve factuality in table-to-text generation tasks \cite{DBLP:conf/emnlp/ParikhWGFDYD20}{, while} RefinedWeb, an automatically filtered and deduplicated web dataset, can mitigate hallucinations and enhance LLM reliability \cite{DBLP:conf/nips/PenedoMHCACPAL23}.
    
    \item \textit{Reinforcement learning:} {By integrating} external feedback mechanisms and reward-based constraints, reinforcement learning {discourages LM agents from generating} hallucinatory information. {This approach enables agents to recognize their limitations} and avoid answering beyond their knowledge scope instead of fabricating untruthful responses \cite{mndler2024selfcontradictory, wang2023self}. For instance, GPT-4 collects {user-annotated} unfactual data and generates synthetic closed-domain hallucinated synthetic data to train a reward model, {thereby reducing hallucination tendencies.}
 
    \item \textit{Hallucination detection:} {Identifying hallucinations post-generation allows for sample regeneration} to eliminate inaccuracies. Techniques such as SelfcheckGPT \cite{DBLP:conf/emnlp/ManakulLG23} and INSIDE \cite{chen2024inside} assess consistency across multiple generated responses to {detect hallucinated content.}
    
    \item \textit{Truthful external knowledge \& RAG:} Truthful RAG techniques can help {anchor LM-generated content in factual information} \cite{lewis2020retrieval, wang2023self}. By {incorporating verified external knowledge sources} and established contextual knowledge, LM agents can reduce hallucinations in generated responses \cite{DBLP:conf/acl/MallenAZDKH23}. 
    
    \item \textit{Instruction tuning:} {Optimizing instruction sets or fine-tuning LMs on robust instruction datasets reduces hallucinations. For instance,} SYNTRA \cite{jones2024teaching} evaluates hallucinations on synthesis tasks and refines instructions to reduce hallucinations on downstream tasks. Besides, Liu \textit{et al.} \cite{liu2023mitigating} mitigate hallucinations in LVMs by constructing a robust instruction dataset named LRV-Instuction and fine-tuning LVMs on it.
    
    \item \textit{Post-processing:} {Refining LM-generated content through post-processing} can correct hallucinations in LM responses \cite{DBLP:conf/acl/GaoDPCCFZLLJG23, zhou2024analyzing}. 
    For instance, RARR \cite{DBLP:conf/acl/GaoDPCCFZLLJG23} utilizes search engines to gather relevant external knowledge and refine responses{, while} LURE \cite{zhou2024analyzing} reduces hallucinations by correcting object-level inaccuracies during post-processing.
    
    \item \textit{Model architecture optimization:} {Structural modifications to LMs} can reduce hallucinations. Multi-branch decoder \cite{rebuffel2022controlling} and uncertainty-aware decoder \cite{DBLP:conf/eacl/XiaoW21} are two examples of modified decoder structures to mitigate hallucinations. 
    
    \item \textit{Meta programming:} {Meta-programming frameworks can} guide and promote collaboration among LM agents, thereby reducing hallucinations in complex tasks \cite{hong2024metagpt}.
\end{itemize}

\subsection{Summary and Lessons Learned}
{The reliability of LM agents is primarily compromised by hallucinations, which arise from biased training data, imperfect learning and inference processes, and multi-agent interactions. While countermeasures such as data sanitization, hallucination detection, and RAG can mitigate these issues, the probabilistic nature of generative models makes hallucinations inherently unavoidable. Future research should continuously focus on hallucination mitigation methods, including multi-agent consensus mechanisms and dynamic credibility audit frameworks, to enhance the reliability of both generated content and decision-making processes in LM agent networks. Table~\ref{tab:reliabilitysummary} summarizes existing typical researches on reliability issues and countermeasures to LM agent networks.}

\section{Future Research Directions}\label{sec:FUTUREWORK}
In this section, we outline several open research directions important to the design of future design of LM agent ecosystem.

\subsection{Energy-Efficient and Green LM Agents}\label{subsec:Green}
With the increasingly widespread deployment of LM agents, their energy consumption and environmental impact have emerged as critical concerns. As reported, the energy consumed by ChatGPT to answer a single question for 590 million users is comparable to the monthly electricity usage of 175,000 Danes \cite{Wang2023AChatGPT}. Given the exponential growth in model size and the computational resources required, energy-efficient strategies are essential for sustainable AI development, with the goal to lower the significant carbon footprint associated with training and operating LM agents.
Enabling technologies for energy-efficient and green LM agents include model compression techniques \cite{zhang2024vpgtrans,ma2023llm}, such as quantization, pruning, and distillation, which lower model size and computational demands of LMs with minimal impact on accuracy. Additionally, the use of edge computing \cite{10466747} and FL \cite{10634552} enables the distribution of computational tasks across edge nodes, thereby reducing the energy burden on central servers and enabling real-time processing with lower latency. Innovations in hardware \cite{shen2024agile}, such as energy-efficient GPUs and TPUs, also play a critical role in achieving greener LM agents by optimizing the energy use of the underlying computational infrastructure.

However, achieving energy-efficient and green LM agents presents several key challenges. While model compression techniques can significantly reduce energy consumption, they may also lead to a loss of accuracy or the inability to handle complex tasks, which is a critical consideration for applications requiring high precision. Furthermore, optimizing the lifecycle energy consumption of LM agents involves addressing energy use across training, deployment, and operational stages. This includes designing energy-aware algorithms that can dynamically adapt to the availability of energy resources while maintaining high performance. 

\subsection{Fair and Explainable LM Agents}\label{subsec:Explainable}
As LM agents continue to play an increasingly central role in decision-making across various domains, the need for fairness and explainability becomes paramount to build trust among users, ensure compliance with ethical standards, and prevent unintended biases. It is particular for sensitive areas such as healthcare, finance, and law, where decisions should be transparent, justifiable, and free from bias.
Bias detection and mitigation algorithms, e.g., adversarial debiasing \cite{10203320}, reweighting \cite{9711042}, and fairness constraints \cite{5740907}, can be incorporated into the training process to ensure that the LM agents are less prone to propagate existing biases, thereby identifying and correcting biases in data and model outputs. Moreover, explainable AI (XAI) mechanisms \cite{10188681}, e.g., local interpretable model-agnostic explanations (LIME), shapley additive explanations (SHAP), and counterfactual explanations allow users to understand the reasoning behind the LM's predictions, thereby enhancing trust, transparency, and accountability.

However, several key challenges remain to be addressed. One major challenge lies in balancing model complexity and explainability. While more complex models, such as DNNs, tend to achieve superior performance, they are often difficult to interpret. Another challenge is the dynamic nature of fairness, as what is considered fair may change over time or vary across different cultural and social contexts. Ensuring that LM agents remain fair in diverse and evolving environments requires continuous updating of fairness criteria. Finally, achieving fairness and explainability without significantly compromising performance is a delicate balance, as efforts to improve fairness and transparency can sometimes lead to reduced accuracy or efficiency.

\subsection{Cyber-Physical-Social Secure LM Agent Systems}\label{subsec:CPSS}
As LM agents increasingly interact with the physical world, digital networks, and human society, ensuring their interaction security in CPSS becomes essential to protect critical infrastructure, preserve sensitive data, prevent potential harm, and maintain public confidence.
Zero-trust architectures \cite{10330693}, which operate under the principle of ``never trust, always verify", are crucial for protecting LM agents from internal and external threats by continuously validating user identities and device integrity. Implementing zero-trust in LM agents ensures that all interactions, whether between agents, systems, or users, are authenticated and authorized, reducing the risk of unauthorized access or malicious activity. Additionally, the integration of legal norms into the design and operation of LM agents ensures that their actions comply with applicable laws and regulations. This involves embedding legal reasoning capabilities within LM agents \cite{liu2024training}, enabling them to consider legal implications and ensure that their decisions align with societal expectations and regulatory frameworks.

However, several key challenges remain. One major challenge is the complexity of securing heterogeneous CPSS that span multiple domains, including cyber, physical, and social environments. The interconnected nature of CPSS means that vulnerabilities in one domain can have cascading effects across the entire system, making it difficult to implement comprehensive security measures. Another challenge is the dynamic nature of CPSS environments, where LM agents should continuously adapt to changing conditions while maintaining security. Ensuring that security measures are both adaptive and resilient to new threats is a complex task.

\subsection{Value Ecosystem of LM Agents}\label{subsec:Value}
The creation of interconnected value network of LM agents empowers LM agents to autonomously and transparently manage value exchanges (e.g., data, knowledge, resources, and digital currencies), which is crucial for fostering innovation, enhancing cooperation, and driving economic growth within LM agents ecosystem. 
Blockchain technology provides a tamper-proof ledger that records all transactions between LM agents, ensuring transparency and trust in the system. Smart contracts, which are self-executing agreements coded onto the blockchain, allow LM agents to autonomously manage transactions, enforce agreements, and execute tasks without the need for intermediaries \cite{10483549}. Additionally, the integration of oracles, i.e., trusted data sources that feed real-world information into the blockchain, enables LM agents to interact with external data and execute contracts based on real-time conditions, further enhancing the functionality of value networks.

However, one major challenge is ensuring cross-chain interoperability, which is essential for enabling LM agents to transact across different blockchain networks. Currently, most blockchains operate in silos, making it difficult to transfer value or data between them \cite{10483549}. Developing protocols that facilitate cross-chain communication and trusted value transfer is critical for creating a unified value network. Another challenge lies in the reliability and security of cross-contract value transfer operations, where multiple smart contracts atop on various homogeneous or heterogeneous blockchains, especially in environments with varying trust levels, need to work together to complete a transaction or task. Additionally, scalability remains a challenge, as the computational and storage requirements for managing large-scale value networks can be substantial. As the number of LM agents and transactions grows, ensuring that the underlying blockchain infrastructure can scale to meet demand without compromising performance or security is crucial.

\section{Conclusion}\label{sec:CONSLUSION}
In this paper, we have provided an in-depth survey of the state-of-the-art in the architecture, cooperation paradigms, security and privacy, and future trends of LM agent networks.
Specifically, we have introduced a novel architecture and its key components, critical characteristics, enabling technologies, and potential applications, toward autonomous, embodied, and connected intelligence of LM agents.
Afterward, we have explored the taxonomy of interaction patterns and practical collaboration paradigms among LM agents, including data, computation, and information sharing for collective intelligence. Furthermore, we have identified significant security and privacy threats inherent in the ecosystem of LM agents, discussed the challenges of security/privacy protections in multi-agent environments, and reviewed existing and potential countermeasures.
As the field progresses, ongoing research and innovation will be crucial for overcoming existing limitations and harnessing the full potential of LM agents in transforming intelligent systems.

\bibliographystyle{IEEETran}

\bibliography{ref_LMagent}

\end{document}